Generative AIBIM: An automatic and intelligent structural design pipeline integrating BIM and generative AI

# Generative AIBIM: An automatic and intelligent structural design pipeline integrating BIM and generative AI

Zhili He[a], Yu-Hsing Wang[a*], Jian Zhang[b]

[a]*Department of Civil and Environmental Engineering,*

*The Hong Kong University of Science and Technology, HKSAR, China*

[b]*School of Civil Engineering, Southeast University, Nanjing, China*

**ABSTRACT:**

AI-based intelligent structural design represents a transformative approach that addresses the inefficiencies inherent in traditional structural design practices. This paper innovates the existing AI-based design frameworks from four aspects and proposes Generative AIBIM: an automatic and intelligent structural design pipeline that integrates Building Information Modeling (BIM) and generative AI. First, the proposed pipeline not only broadens the application scope of BIM, which aligns with BIM's growing relevance in civil engineering, but also marks a significant supplement to previous methods that relied solely on CAD drawings. Second, in Generative AIBIM, a two-stage generation framework incorporating generative AI (TGAI), inspired by the human drawing process, is designed to simplify the complexity of the structural design problem. Third, for the generative AI model in TGAI, this paper pioneers to fuse physical conditions into diffusion models (DMs) to build a novel physics-based conditional diffusion model (PCDM). In contrast to conventional DMs, on the one hand, PCDM directly predicts shear wall drawings to focus on similarity, and on the other hand, PCDM effectively fuses cross-domain information, *i.e.,* design drawings (image data), timesteps, and physical conditions, by integrating well-designed attention modules. Additionally, a new evaluation system including objective and subjective measures (*i.e.*, $Score_{\mathrm{IoU}}$ and FID) is designed to comprehensively evaluate models' performance, complementing the evaluation system in the traditional methods only adopting the objective metric. The quantitative results demonstrate that PCDM significantly surpasses recent state-of-the-art (SOTA) techniques (StructGAN and its variants) across both measures: $Score_{\mathrm{IoU}}$ of PCDM is $30\%$ higher and FID of PCDM is lower than $1/3$ of that of the best competitor. The qualitative experimental results highlight PCDM's superior capabilities in generating high perceptual quality design drawings adhering to essential design criteria. In addition, benefiting from the fusion of physical conditions, PCDM effectively supports diverse and creative designs tailored to building heights and seismic precautionary intensities, showcasing its unique and powerful generation and generalization capabilities. Associated ablation studies further demonstrate the effectiveness of our method.

---

* Corresponding author.

*E-mail addresses:* ceyhwang@ust.hk (Y.-H. Wang).

# 1. Introduction

Structural design plays a pivotal role in the comprehensive lifecycle of building projects, particularly as global urbanization accelerates, leading to a substantial annual increase in building construction [1]. This surge has created a significant demand for efficient and effective structural designs. Currently, structural engineers perform structural designs through a manual process characterized by two key features. First, structural designs heavily rely on extensive structural knowledge and design experience [2]. This human-centric approach makes it nearly impossible to consider all explicit and implicit rules and constraints, leading to the second feature: iterative design refinement [3]. As a result, structural designs are traditionally time-consuming [1], labor-intensive, and inefficient [4], making it challenging to meet current high demands [2]. Consequently, working overtime is prevalent in design institutes. To address these limitations, practitioners require an innovative and intelligent design paradigm.

Recent advancements in artificial intelligence (AI) technologies are revolutionizing numerous industries, such as defect detection [5], anomaly detection [6], the aerospace industry [7], and communication engineering [8]. AI is data-driven [9] and possesses two remarkable attributes. First, AI models are trained on vast amounts of data to learn, memorize, compress, and represent extensive data features. Compared to humans, AI models exhibit limitless memory, as the more data they are trained on, the stronger their performance and representation ability become. For instance, large language models, such as ChatGPT and GPT-4, are trained on immense human knowledge bases, even leading to unforeseen emergent abilities [10]. Second, AI models are automated and efficient, with inference times that are negligible compared to humans. By combining the strengths of AI and structural design, it becomes evident that AI models are inherently suitable for structural design tasks and have great potential to become the new design paradigm. As a result, the integration of AI and structural design has become a prominent research focus in the academic community.

Nevertheless, the architectural diversity of buildings, characterized by a multitude of forms and layouts, currently hinders the practicality of incorporating all building types into AI models, given the nascent stage of exploration in this domain. At present, scholarly efforts are primarily concentrated on leveraging AI for the intelligent design of high-rise residential buildings that employ reinforced concrete shear wall systems [1][2][3][11][12]. This focus is driven by two primary considerations. First, among the myriad of building categories, residential buildings constructed with reinforced concrete shear wall systems emerge as the predominant form of high-rise buildings and occupy a significant proportion in all buildings [1][13]. Moreover, with the global population on a continuous upward trajectory, the demand for such buildings is not only substantial but also expanding [14]. Second, the architectural configuration of these buildings tends to exhibit a higher degree of uniformity, often characterized by rooms with rectangular shapes and structural layouts comprising rectangular walls [11][12]. This uniformity simplifies the application of AI in the structural design process, making it a practical focus for current research endeavors.

These studies of AI-based intelligent structural design can be categorized into two main approaches: regression-based methods and generative model-based methods. Regression-based methods model the structural design task as a regression problem. They train neural networks, such as multi-layer perceptrons (MLPs) [11][15], convolutional neural networks (CNNs) [12], or graph neural networks (GNNs) [16], to regress structural designs in a supervised way using regression loss functions. Essentially, regression-based methods are a direct fitting to the training set, leading to a lack of diversity in the regressed outcomes and diminished generalization capability of the models. Therefore, regression-based methods are not suitable for structural design tasks that require creativity and diversity. Generative model-based methods are a groundbreaking alternative to the traditional regression-based methods. These methods treat the structural design task as a generation problem. They train generative models to learn data distributions of images and then sample from the learned distributions to generate high-quality samples.

StructGAN [1] and its variants [2][3] represent the first generative model-based structural design frameworks. They utilize generative adversarial networks (GANs) [17], one of the most powerful generative models, for the intelligent design of shear walls. Compared with the regression-based methods, StructGAN-based models can generate structural design drawings with higher visual quality and greater diversity. These achievements have attracted significant attention to the StructGAN series of methods, marking a notable and recent advancement in the field. However, there is a continued need for further enhancements and deeper exploration in four key areas, as detailed subsequently.

The first area identified for enhancement involves broadening the application scope. Existing StructGAN-based models are exclusively developed with CAD design drawings [2]. However, Building Information Modeling (BIM) nowadays plays a pivotal role throughout the entire lifecycle of construction projects [18], offering advantages such as improved information management, enhanced efficiency, and increased transparency. In some regions or countries, the utilization of BIM is even mandated for certain projects. For instance, in Hong Kong, capital works projects with a budget exceeding 30 million HKD necessitate to implement BIM [19]. Recognizing this gap, our study aims to integrate BIM with the intelligent design of shear walls. We propose a novel structural design pipeline that merges BIM with generative AI technologies, which we refer to as ***Generative AIBIM*** for ease of reference. This innovative approach seeks to complement existing methodologies that are limited to CAD drawings, with a detailed discussion presented in Section 3.

The second area identified for development concerns the generation framework utilized for producing the structure design drawing. Current StructGAN-based pipelines utilize an end-to-end generation framework incorporating generative AI (EGAI). In this framework, the input is an architectural design drawing, and the framework requires generative AI models (*i.e.*, GANs) to directly produce the corresponding structural design drawing. As illustrated in Fig. 1 (a), this process entails AI models generating shear walls, infill walls, windows, and doors simultaneously while ensuring that the non-structural components match those in the architectural drawing. This task presents considerable challenges, and as depicted in Fig. 1 (a), the output of StructGAN reveals potential for enhancement in both perceptual quality and detail accuracy. Recognizing the inherent complexity of this task, we propose a novel approach by mimicking the human drawing process. This insight has led us to develop a two-stage generation framework incorporating generative AI (TGAI) that simplifies the generative AI models' task to essentially paint a line drawing onto a predefined canvas. Further elaboration on this innovative framework is detailed in Section 4.

The third area for improvement pertains to the choice of generative AI models in the generation framework and the network structures of AI models to generate structure design drawings of better quality. Beyond the limitation of the generation framework, the visual quality of images produced by StructGAN is also constrained by the specific GAN methodology it employs. The adversarial training characteristic of GANs introduces two primary challenges: (1) GAN training is unstable and prone to oscillation, necessitating carefully selected optimization strategies and structural adjustments to achieve stability [20]. For instance, StructGAN incorporates the advanced pix2pixHD model [21] as its foundational architecture; and (2) GANs often struggle to fully capture the diversity of the training data's distribution [22]. This means GANs may focus on certain data distribution modes while neglecting others, leading to the well-documented issue of mode collapse [23]. For example, as illustrated in Fig. 1 (a), designed shear walls fail to adhere to basic design principles, such as maintaining horizontal and vertical boundaries and enclosing shear walls within elevator areas, and the clarity of the generated drawing is insufficient. These observations suggest that critical data modes are overlooked, necessitating significant post-processing by engineers. In response to these limitations, recent advancements have introduced diffusion models (DMs) [24] represented by the denoising diffusion probabilistic model (DDPM) [25], which operate without adversarial training leading to stable training process (results in Subsection 7.5 provide the corresponding experimental evidence) and show remarkable capability in capturing complete image distributions. The advantages make DMs achieve state-of-the-art (SOTA) results not only in image synthesis [26][27] but also in a wide range of computer vision downstream tasks, such as super-resolution [28], deblurring [29][30], inpainting [31], denoising [32], and sharpening of multispectral and hyperspectral images [33][34][35]. The applicability of DMs to shear wall design tasks, however, remains unexplored. In this study, we investigate this potential by introducing a novel physics-based conditional diffusion model (PCDM). In contrast to traditional DMs, PCDM has two main improvements. First, in the optimization process, unlike traditional DMs that predict noise, PCDM directly predicts the shear wall drawings. This change allows PCDM to focus on similarity, thereby enhancing the realism of the generated structural design drawings. The theoretical equivalence between predicting shear wall drawings and noise is discussed in Subsection 5.4, and the associated ablation studies demonstrate the practical effectiveness of predicting shear walls (refer to Subsection 7.3). Second, PCDM effectively integrates multi-source and multi-modal information, namely, design drawings, timestep information, and physical conditions (see Subsection 5.5.1). This advancement provides PCDM with the unique abilities to support diverse structural designs (as detailed in Subsection 7.4.1) and creative structural designs (as described in Subsection 7.4.2) tailored to given physical conditions, including various building heights and earthquake precautionary intensities, demonstrating PCDM's powerful and robust generation and generalization capabilities. It is important to note that such generation capabilities do not exist in conventional DMs and even current state-of-the-art (SOTA) techniques in the field of intelligent structural designs (*i.e.*, StructGAN-based approaches). Furthermore, benefiting from the adopted DM paradigm, PCDM can accurately capture the complete image distributions without neglecting key distribution modes, which brings PCDM two advantages that do not exist in the current SOTA models, as illustrated in Fig. 1. First, PCDM can generate clearer designs with higher perceptual quality. Second, PCDM can correctly learn design experiences and knowledge from human design drawings, making the generated design drawings with accurate details and adhering to essential design criteria, such as horizontal and vertical wall boundaries and incorporating shear walls within elevator areas to form core tubes. Subsections 7.1 and 7.2 provide concrete experimental evidence to support PCDM's enhanced generation capabilities for structural design drawings. Moreover, to facilitate cross-domain data fusion (*i.e.*, combining drawings, timesteps, and physical conditions), we have developed a novel attention block (AB) that includes a self-attention block (SAB) and a parallel cross-attention block (PCAB), alongside an adaptive Instance Normalization (AdaIN) block, all of which are integrated into the neural network of PCDM. The architecture and functionalities of PCDM and the neural network enhancements are detailed in Section 5.

The fourth area identified for enhancement concerns the evaluation metrics utilized. The evaluation system in StructGAN and its derivatives solely has the objective metrics, $Score_{\mathrm{IoU}}$, which assess the predicted outcomes against labels on a pixel-by-pixel basis. This system overlooks subjective metrics that mimic human perception and are capable of quantitatively appraising the visual quality of the generated samples. To address this gap, as detailed in Section 6, our study further introduces a widely adopted subjective metric, the Fréchet Inception Distance (FID), to evaluate whether the generated structural drawings resemble those crafted by engineers. By integrating both objective and subjective metrics, our new evaluation system aims to provide a more holistic assessment of models' performance.

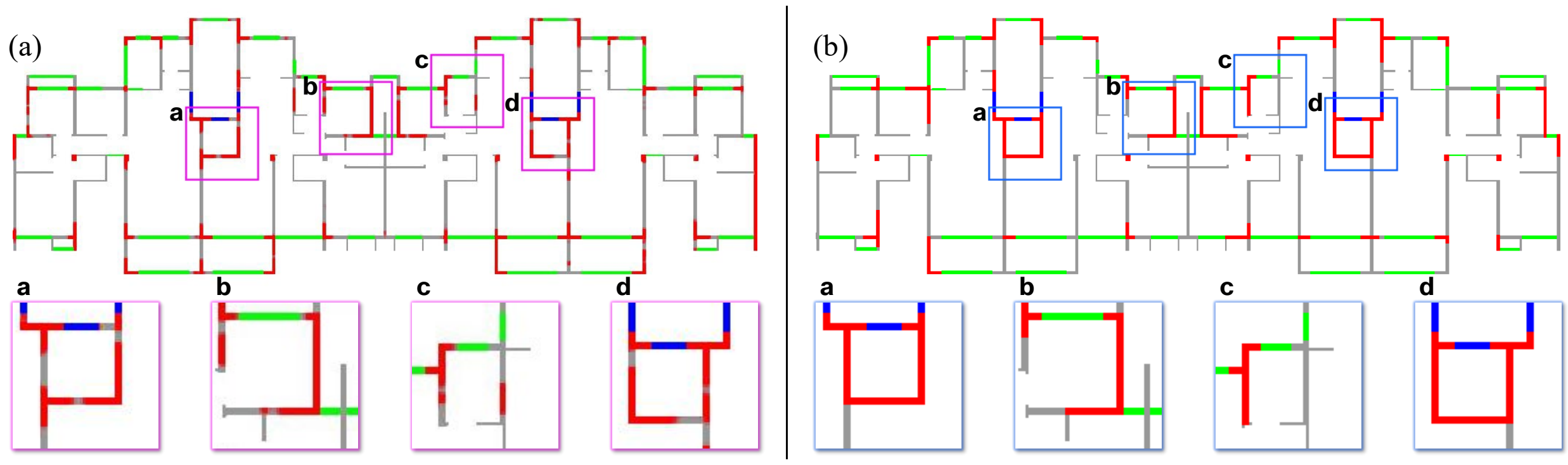

**Fig. 1.** Comparison of (a) structure design generated by StructGAN [1] and (b) structure design generated by the proposed physics-based conditional diffusion model (PCDM) in this study. Note that the red pixels denote shear walls, *i.e.*, the part designed by AI, the grey pixels represent infill walls, green pixels stand for windows, and blue pixels indicate doors. A more detailed explanation is given in Section 4.

To summarize, the principal contributions of this research are outlined as follows:

(1) Application scope: We integrate Building Information Modeling (BIM) with the intelligent design of shear walls and introduce a novel Generative AIBIM pipeline for structural design. This integration pipeline not only broadens the application scope of BIM, which aligns with BIM's growing relevance in civil engineering, but also supplements existing intelligent design methodologies that are limited to CAD drawings.

(2) Generation framework: Drawing inspiration from the human drawing process, we propose a two-stage generation framework incorporating generative AI (TGAI). Compared to the traditional end-to-end generation framework incorporating generative AI (EGAI), this new framework effectively simplifies the shear wall generation problem to facilitate AI models to address.

(3) Generative AI model: This research pioneers the application of DMs to the intelligent design of shear walls and develops an advanced DM, PCDM. In contrast to the SOTA StructGAN-based approaches, on the one hand, benefiting from the fusion of cross-domain information, PCDM supports diverse and creative designs, showcasing its unique and powerful generation and generalization capabilities. On the other hand, PCDM demonstrates more superior capabilities in generating clear design drawings with high-fidelity details. To our knowledge, this represents the first systematic exploration of integrating DMs with civil engineering applications. Our findings suggest that DMs have the potential to replace GANs as a new standard for addressing generative challenges within the civil engineering domain.

(4) Network structure: To support the fusion of data from different domains, we introduce an innovative AB that includes a SAB and a PCAB, along with an AdaIN block, all of which are integrated into the neural network.

(5) Evaluation metric: A new evaluation system, including subjective and objective metrics (namely, FID and $Score_{\mathrm{IoU}}$) is designed to provide a more precise and comprehensive assessment of the models' performance. This system complements the evaluation system in the traditional methods only adopting the objective metric $Score_{\mathrm{IoU}}$.

The organization of the paper is as follows: In Section 2, we review regression-based methods and generative model-based methods. Section 3 provides an in-depth discussion of the proposed structural design pipeline utilizing Generative AIBIM. Section 4 offers detailed descriptions of TGAI. Section 5 elaborates on the design of PCDM and its underlying neural network architecture. Implementation details are thoroughly explained in Section 6. Section 7 presents the experimental results and provides a detailed discussion of the findings. Section 8 concludes the paper and offers recommendations for future research. The paper focuses on conveying the fundamental concepts of DMs in an accessible and straightforward manner, aiming to ensure clarity for the civil engineering audience. Detailed theoretical derivations and proofs are provided in the appendixes for those interested in deeper exploration. The implementation code for Generative AIBIM such as the PCDM code, and the dataset are publicly available at https://github.com/hzlbbfrog/Generative-BIM. Additional resources, including the project page, can be accessed at http://zl-he.com/Generative-BIM.

## 2. Related work

### *2.1. Regression-based methods*

Regression-based methods model the structure design task as a regression problem and then train neural networks to regress structure designs, utilizing regression loss functions—predominantly $L_1$ or $L_2$ loss functions. Pizarro et al. [11][12] were the first to explore regression-based methodologies for structural design. They employed MLPs and CNNs to directly regress pixel-level layouts of shear wall structures in a supervised and end-to-end framework. Thus, their methods are named end-to-end regression-based methods. Despite their innovative approach, the application of end-to-end regression-based methods in image synthesis tasks revealed limitations in terms of representation power, leading to the generation of comparatively coarse designs. The constraints of these methods manifest in three primary ways: (1) the focus on optimizing distortion-based metrics, which often do not align with human perceptual quality [36], resulting in generated images of noticeably lower quality [29]; (2) the tendency of regression-based methods to produce images lacking in high-fidelity details [28], attributed to the conservative nature of regression losses towards high-frequency details [30]; and (3) regression-based methods aim to directly fit the training set, leading to a lack of diversity in the outcomes and a poor generalization capability of the trained AI models.

In response to these challenges, Zhao et al. [4][37] introduced an innovative application of GNNs [16] to refine the regression-based approach. Their method starts with transforming the original architectural layout into a graph representation. Subsequently, they train GNN models to minimize regression loss functions between the GNN-generated design drawings and the actual designs. During the inference phase, these GNN-generated graphs are translated back into shear wall designs through a reverse graph representation process. This kind of approach is named indirect regression-based methods due to its need for multiple stages to complete the structural design. Benefiting from the incorporation of graph constraints, the indirect regression-based methods can produce more accurate design drawings with high fidelity compared to traditional end-to-end regression-based methods. However, this method still encounters the inherent drawbacks of regression-based approaches, namely, limited diversity in generated designs and limited generalization capability of the models. Additionally, the graph representation process introduces a level of complexity that could be perceived as cumbersome.

### *2.2. Generative model-based methods*

Since there is a contradiction between the inherent drawbacks of regression-based approaches and the structure design tasks requiring creativity and diversity, generative model-based methods are further proposed to substitute the regression-based methods. Unlike their predecessors, which directly adapt to datasets, the generative model-based methods model the structure design task as an image generation problem. Then, they train deep generative models to master the intricate data distributions of images, subsequently generating high-quality samples by sampling from these learned data distributions. A wealth of research underscores the capability of deep generative models to produce images that are not only more detailed and diverse but also more convincing and perceptually accurate [28][29][30][38]. Thus, generative model-based methods are more appropriate to the structure design task. Our proposed methodology in this study, therefore, belongs to the kind of methods that leverage the strengths of deep generative models.

So far, researchers have developed many types of deep generative models, and they can be usually categorized into likelihood-based models and models based on adversarial training. The former category encompasses variational autoencoders (VAEs) [39], flow-based generative models [40][41], and autoregressive models [42]. On the other hand, adversarial training has given rise to a well-known subset of generative models: generative adversarial networks (GANs). GANs, in particular, have been designed with a focus on enhancing image generation [43][44], with extensive studies demonstrating their superiority in producing more sophisticated and higher-quality samples compared to likelihood-based methods [38][45]. This has led to their widespread adoption in various computer vision applications, including style transfer [46] and image restoration tasks [47][48]. In the civil engineering domain, GANs have found numerous successful applications, ranging from bolt inspection [49] and crack segmentation [50][51] to the generation of house floor plans [52][53][54].

Drawing inspiration from these achievements of GANs, Liao et al. [1] take the lead to expand the use of GANs to the intelligent design of shear walls, introducing StructGAN [1] along with its variants, StructGAN-PHY [2], and StructGAN-AE [3], and the results demonstrate that StructGAN-based models can generate the shear wall design drawings with higher visual quality and more diversity than the regression-based methods. The progress makes StructGAN-based models the most advanced and representative solutions for intelligent structural design. Despite these successes, there remains a need for further improvement in the application scope, the generation framework, the adopted generative AI model and network structure, and the evaluation metric. Thus, this paper proposes Generative AIBIM to innovate the existing StructGAN-based solutions in the above four aspects.

## 3. Generative AIBIM

### *3.1. Motivation and design concept*

The overall workflow of StructGAN [1] can be succinctly divided into three primary stages: (1) converting CAD drawings into architectural design drawings, (2) transforming architectural design drawings into structural design drawings with the aid of GANs for the intelligent design of shear walls, and (3) evolving structural design drawings into conventional structural models. In the workflow, architectural and structural design drawings serve as the core components of this process. Our research is designed to align with StructGAN and its variants, adhering to the same protocol to facilitate and enhance community collaboration in smart structural design. Consequently, the standards for architectural design and structural design drawings in StructGAN are maintained.

To fill the gap between BIM models and 2D design drawings, we design two independent stages in the Generative AIBIM pipeline, namely, Stage I: from BIM models to architectural design drawings (see Subsection 3.2), and Stage III: from structural design drawings to BIM models (refer Subsection 3.4). Further, the main challenge is how to generate structural design drawings, which will be sent to Stage III from architectural design drawings provided in Stage I. To address this, we design a novel two-stage generation framework incorporating generative AI (TGAI), as summarized in Stage II in Subsection 3.3. Notably, TGAI is the most critical contribution and innovation of this study, and therefore the subsequent sections are all related to this. Besides benefiting from the extraordinary scalability of BIM, BIM models can be integrated with multiple technologies to facilitate entire project lifecycles, and to fully utilize the advantage of BIM, the proposed intelligent structural design pipeline incorporates BIM. In order to highlight the potential advancement brought by Generative AIBIM, which but cannot be achieved by the traditional methods without BIM, such as StructGAN and its variants, our pipeline introduces an additional stage (Stage IV in Subsection 3.5) to review BIM-supported applications throughout the lifecycles of projects.

The following content of the section details our proposed intelligent structural design pipeline, Generative AIBIM.

### 3.2. *Stage I: from BIM models to architectural design drawings*

The objective of this stage is to generate standard architectural design drawings from the provided BIM models. Initially, a BIM model of the project is acquired using BIM design software, with Revit being recommended due to its widespread adoption in BIM project design and management. The primary challenge is converting three-dimensional (3D) BIM models into two-dimensional (2D) architectural design drawings, a task that lacks a direct method. To tackle this, as illustrated in Fig. 2, our solution involves two steps: first, the creation of Dynamo code to automatically export data to Excel from BIM models, and second, the development of a Python script to generate standard architectural design drawings from these Excel files. Although the current methodology may seem indirect, our future research endeavors include developing an easy-to-use Revit plugin that consolidates these functions, thereby simplifying the process for engineers. As noted, our research is dedicated to promoting community collaboration. Therefore, the standards for architectural and structural design drawings established in StructGAN are maintained, ensuring that the methods are interoperable and can be seamlessly integrated within the broader intelligent structural design ecosystem. For instance, in the drawings, windows are represented by green pixels and doors by blue pixels, with detailed rules elucidated in Section 4.

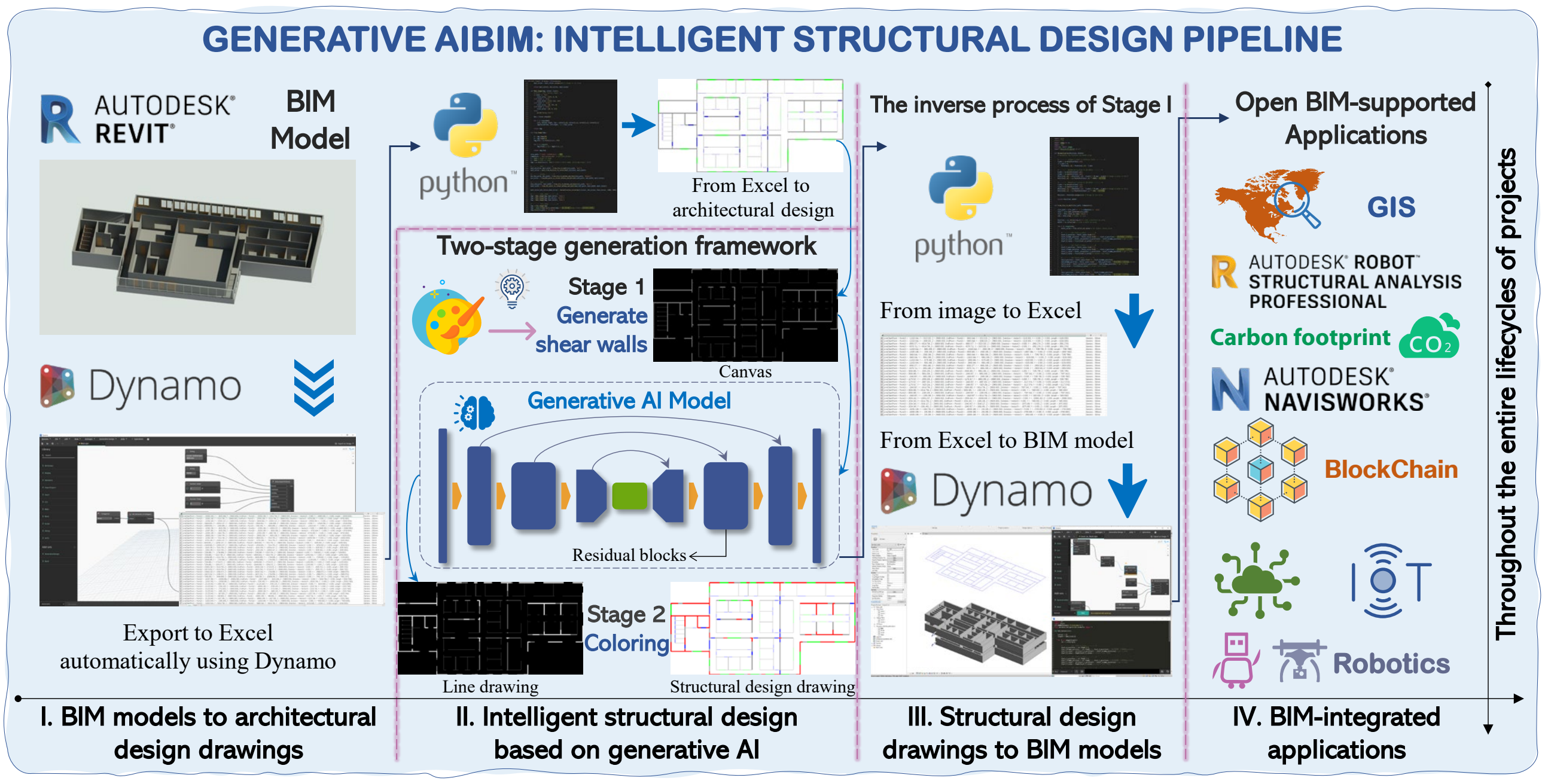

**Fig. 2.** The proposed Generative AIBIM: an automatic and intelligent structural design pipeline integrating BIM and generative AI.

### 3.3. *Stage II: Intelligent structural design based on generative AI*

The objective of the second stage is to execute smart structural design using generative AI. Specifically, we want to generate structural design drawings with shear walls based on the architectural drawings produced in the first stage. Inspired by the process of human drawing, we introduce a two-stage generation framework incorporating generative AI (TGAI), which, as shown in Stage II of Fig. 2, includes two substages: (1) generating shear walls and (2) coloring, to simplify the generation problem. For the generative AI part in TGAI, a novel physics-based conditional diffusion model (PCDM) is proposed. Since TGAI and its incorporated PCDM stand as the core contributions and main innovations, we provide a detailed introduction in Sections 4 and 5, respectively. Notably, TGAI indeed considers compatibility and versatility, thereby enabling the seamless integration of various generative AI methods. This includes not only the GAN-based approaches, such as StructGAN and its variants, but also advanced diffusion-model-based techniques, exemplified by PCDM.

### *3.4. Stage III: from structural design drawings to BIM models*

Although the second stage generates structural design drawings, this output alone is not sufficient. The ultimate requirement for BIM projects and BIM-supported applications is the creation of BIM models, which serve as the foundational elements. Consequently, the objective of the third stage is to reconstruct 3D BIM models from the synthesized 2D images. This stage essentially reverses the process of the first stage, leading to a straightforward solution: First, Python code is used to convert structural design drawings into Excel files. Second, Dynamo code is employed to generate BIM models from these Excel files. Notably, Stages I and III are engineering-oriented, and the core concepts are more important than the specific implementation. Therefore, Subsections 3.2 and 3.4 focus on the fundamental ideas rather than the detailed code implementation. It is worth noting that we provide a tutorial about how to implement the two stages and the corresponding Python code and Dynamo code at Github (see https://github.com/hzlbbfrog/Generative-BIM).

### *3.5. Stage IV: BIM-integrated applications*

BIM has been integrated with various advanced technologies, playing a pivotal role throughout the entire lifecycle of projects, including planning, design, construction, and operation and maintenance [18]. Hence, with BIM models generated in the third stage, we can readily leverage the advantages of integration with BIM. The fourth stage revisits representative BIM-supported applications throughout project lifecycles to show the potential advantages of Generative AIBIM compared to the traditional frameworks without BIM.

During the planning stage, BIM can be integrated with Geographic Information System (GIS) to enable the effective management and analysis of spatial data [18][55].

During the architecture and building design stage, BIM design tools such as Revit, as previously mentioned, are used to complete architectural designs. Additionally, structural analysis software such as Robot Structural Analysis Professional can be employed to guarantee structural safety through static and dynamic analyses. BIM technology is also instrumental in calculating the carbon footprints of buildings [56], supporting the exploration of sustainable and eco-friendly low-carbon designs [57].

During the construction phase, BIM significantly enhances construction management [58], ultimately improving project delivery. For instance, Navisworks, a BIM review and coordination software, visualizes and unifies design and construction data, helping to identify and resolve clash and interference issues. Moreover, the integration of blockchain technology with BIM can elevate construction process management, including supply chain and contract management [59].

In the operation and maintenance stage, the fusion of Internet of Things (IoT) sensors [60] with BIM enables real-time monitoring and supervision, facilitating project management [61]. Chen et al. [62] developed a BIM intelligent operation and maintenance platform by combining BIM with IoT, big data, and cloud computing, which supports multiple tasks, such as equipment management, space management, and security management. Additionally, some researchers have integrated robotics, such as unmanned ground vehicles (UGVs) [63] and unmanned aerial vehicles (UAVs) [64], with BIM to enhance the efficiency of facility inspections [65].

## 4. Two-stage generation framework incorporating generative AI (TGAI)

As introduced in Section 3, architectural and structural design drawings serve as the fundamental components of the structural design problem. In line with StructGAN, we first establish standards for these design drawings. Architectural design drawings consist of three components: non-structural infill walls, windows, and outdoor gates, depicted by grey pixels (RGB tuple=(152,152,152)), green pixels (RGB tuple=(0,255,0)), and blue pixels (RGB tuple=(0,0,255)), respectively (refer to Fig. 3 (a)). The remaining area, represented by white pixels (RGB

tuple=(255,255,255)), is empty. In structural design drawings, an additional component appears—shear walls, represented by red pixels (RGB tuple=(255,0,0)). It is important to emphasize that the set of shear walls in structural design drawings is a subset of the set of infill walls in architectural design drawings. Hence, the essence of this structural design problem is to effectively transforming a portion of infill walls (grey pixels) into shear walls (red pixels).

For generating structural design drawings, StructGAN and its variants aim to address this challenge in an end-to-end manner. Within their EGAI framework, architectural design drawings serve as the initial inputs, and the AI models (particularly GANs) are trained to directly synthesize structural design drawings (see Fig. 3 (a)). This process requires the AI models to successfully accomplish three tasks simultaneously: (1) select appropriate pixels from the grey pixels and transform their color to red; (2) ensure that the RGB tuples of unselected grey pixels remain unchanged; and (3) maintain the RGB tuples of green, blue, and white pixels unaltered. These tasks are challenging for GANs, which inherently have limited representation and generation capabilities, resulting in potentially unsatisfactory generation outcomes (as depicted in Fig. 1).

This paper therefore introduces a novel perspective to improve the generated outcomes of structural design drawings: generating shear walls is akin to AI models drawing red shear walls on a semi-finished painting (*i.e.*, an architectural design drawing) without elements any other elements of the painting. This analogy suggests a latent relationship between the structural generation task and the drawing procedure. Prompted by this insight, a question emerges: can we draw inspiration from the human drawing process to achieve the objective? With this concept in mind, the authors embark on further exploration.

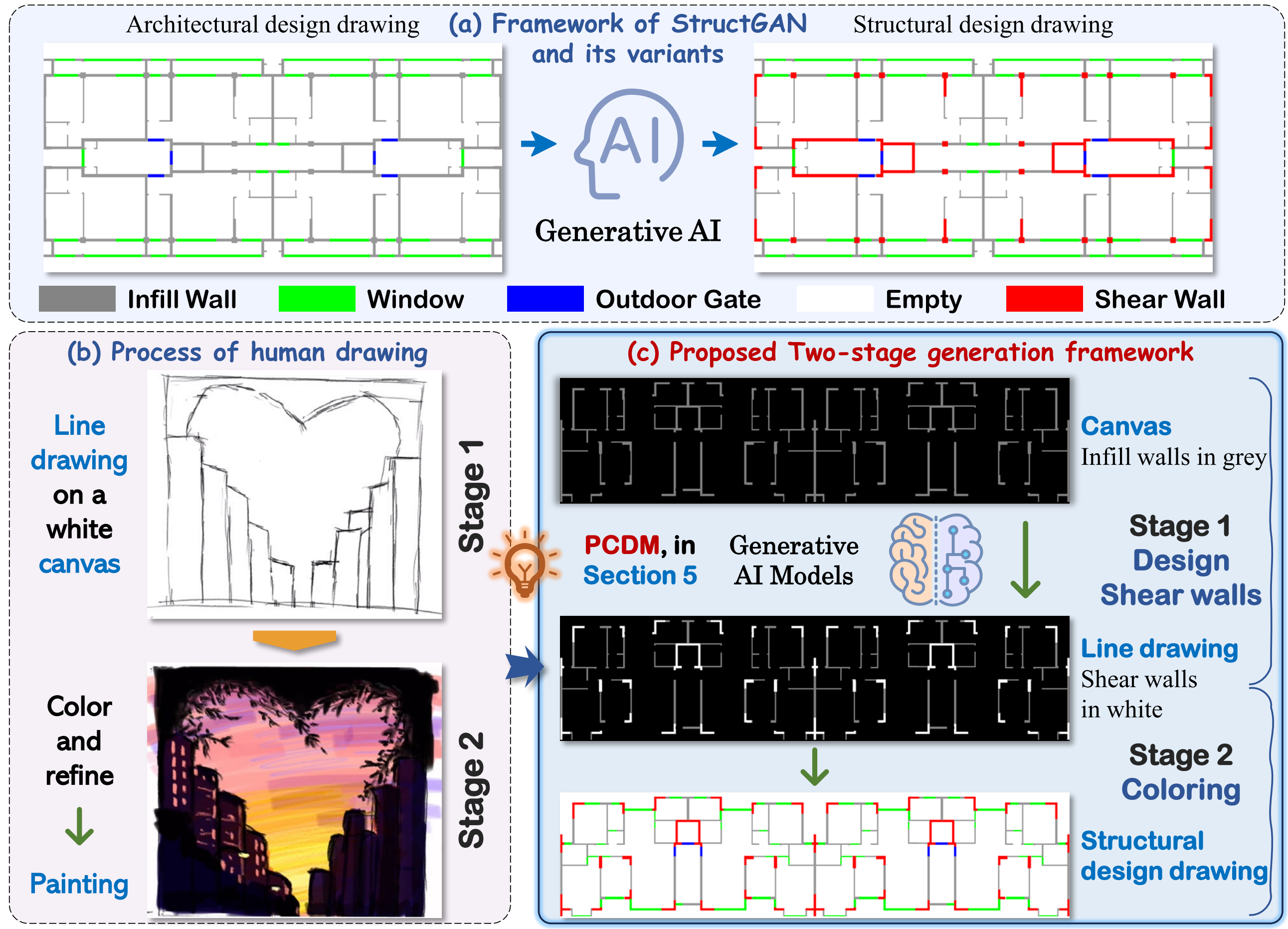

**Fig. 3.** Illustrations of the generation framework: (a) framework adopted in StructGAN [1] and its variants, (b) the process of human drawing, and (c) our proposed two-stage generation framework (TGAI).

Firstly, let us revisit the most common drawing process, which is also the simplest for humans to execute, as depicted in Fig. 3 (b). The drawing process involves two stages: (1) draw essential lines on the pre-prepared canvas and (2) color and refine the line drawing to obtain the final painting. Since this process is easily implemented by humans, it should be reproducible by AI. Inspired by this idea, as illustrated in Fig. 3 (c), we therefore design a novel two-stage generation framework incorporating generative AI (TGAI) that mimics human painting. Initially, a canvas is prepared, retaining only infill walls (represented by grey pixels in Fig. 3 (c)) and removing other components to simplify the task. This simplification is justified because only infill walls undergo changes during the generation process. Furthermore, original three-channel RGB images are compressed into single-channel images (*i.e.*, grey) to reduce data dimensionality and computational costs.

The first stage of TGAI is the design stage to generate shear walls, where the AI model is trained to design shear walls on the given canvas, specifically, it selects appropriate pixels on the canvas and changes their color from grey to white, similar to humans creating a line drawing (see Fig. 3 (b) and (c)). In the second stage, the final structural design drawing is produced through a coloring process. Initially, the white pixels are colored red to represent the AI-generated shear walls. Then, since gates and windows remain unchanged, these elements are simply colored according to their designated lines, referencing the initial architectural drawing. This stage can be executed with straightforward code and does not require additional training.

The new two-stage generation framework significantly simplifies the complexity of generating structural design drawings. Additionally, in the generative AI part of this framework, we have designed an advanced diffusion model (DM) for the structural design problem, termed the physics-based conditional diffusion model (PCDM), which has powerful generation and representation capabilities. The details of PCDM are expounded in Section 5, and the experimental results presented in Section 7 corroborate the effectiveness of the proposed generation framework.

## 5. The proposed physics-based conditional diffusion model (PCDM)

To facilitate the following discussion and avoid confusion, relevant notations and assumptions that are widely accepted and adopted in machine learning research are introduced first.

**Notations.** Random vectors, random matrices, or tensors in the neural networks are denoted by non-tilt boldface upper case letters, such as $\mathbf{X}$ and $\mathbf{M}$. Vectors or specific values of random vectors are denoted by non-tilt boldface lower case letters, such as $\mathbf{x}$ and $\mathbf{y}$. Scalars are represented by tilt non-bold lower-case letters, such as $x$ and $y$. Additionally, vectors are all column vectors. For instance, $\mathbf{x} = [x_1, x_2, \dots, x_N]^\top$ is adopted to represent the vector $\mathbf{x}$ formed by $N$ scalars.

**Assumption 1 (Independence Assumption).** All the random variables in random matrices are mutually independent. According to this assumption, the random matrix $\mathbf{X}_t \in \mathbb{R}^{1\times H\times W}$ can be equivalently transformed into a random vector* $\mathbf{X}_t \in \mathbb{R}^D$, where $D = 1 \times H \times W$. Because random vectors are easier to analyze and understand than random matrices, the research objects in this study are all random vectors transformed from random matrices, unless otherwise stated.

### 5.1. *Preliminary and motivation of PCDM*

As outlined in Section 1, DMs are now among the most prominent deep learning techniques, surpassing GANs in image synthesis capabilities. Consequently, DMs have been chosen as the foundational generative model for this research. Before delving into the specifics of the proposed PCDM, it is essential to review DMs and conditional

* The notation $\mathbf{X}_t$ is kept here for notational simplicity.

diffusion models (CDMs) to provide the necessary background knowledge.

Conventional DMs aim to learn and model the underlying data distribution $q_0(\mathbf{x}_0)$, from image data $\mathbf{x}_0$ by performing variational inference on a Markovian process. They can be approximated as a Markovian hierarchical variational autoencoder [66]. In this paper, $\mathbf{x}_0 \in \mathbb{R}^D$ represents the line drawing of structural design (refer to Fig. 3 (c)), where $D = 1 \times H \times W$, with 1 indicating the number of channels being 1, and $H$ and $W$ representing the height and width of the drawing, respectively. To learn the data distribution $q_0$, firstly, DMs progressively add Gaussian noise to the original clean image by utilizing a fixed $T$-step Markov chain to obtain a list comprising of noisy samples $[\mathbf{x}_1, \mathbf{x}_2, \dots, \mathbf{x}_T]$. The transition probability between adjacent timesteps is denoted as $q(\mathbf{x}_t|\mathbf{x}_{t-1})$. This adding-noise process is commonly referred to as the forward diffusion process [67] or the Gaussian diffusion process [27]. Subsequently, DMs learn a transition distribution $p_{\boldsymbol{\theta}}(\mathbf{x}_{t-1}|\mathbf{x}_t)$ with parameters $\boldsymbol{\theta}$ from data, which is a parametric approximation [27] to a real conditional distribution $q(\mathbf{x}_{t-1}|\mathbf{x}_t)$. According to this distribution, DMs can sample new clean samples from Gaussian noise via a reverse Markov chain so as to achieve the goal of data generation. This process is usually named the reverse denoising process [26] or the reverse diffusion process [67]. In this study, the distribution $p_{\boldsymbol{\theta}}(\mathbf{x}_{t-1}|\mathbf{x}_t)$ is named the denoising model because of its denoising function, and the process of modeling $p_{\boldsymbol{\theta}}(\mathbf{x}_{t-1}|\mathbf{x}_t)$ is named the optimization process [68] due to its optimization nature.

In the conventional DM paradigm, the sampling in the reverse denoising process is random and cannot be controlled. However, the generation problem researched in this paper requires that the generated structural drawings completely follow architectural drawings, for example, positions and thicknesses of walls remain unchanged before and after generation (refer to Stage 1 in Fig. 3 (c) for a more intuitive understanding). Obviously, this is intractable under the traditional DM framework. To solve this kind of controllable generation problem, conventional DMs are extended to conditional diffusion models (CDMs) [27][28][66]. CDMs maintain the forward diffusion process of DMs, however, in the optimization process, the modeling objective is changed to a new transition distribution $p_{\boldsymbol{\theta}}(\mathbf{x}_{t-1}|\mathbf{x}_t, \mathbf{y})$ conditioned on $\mathbf{y}$, so that we can explicitly control generation results through conditional information $\mathbf{y}$. Then, in the reverse denoising process, the transition kernel $p_{\boldsymbol{\theta}}(\mathbf{x}_{t-1}|\mathbf{x}_t)$ is also substituted with $p_{\boldsymbol{\theta}}(\mathbf{x}_{t-1}|\mathbf{x}_t, \mathbf{y})$. In this study, $\mathbf{y} \in \mathbb{R}^D$ means the canvas of architectural design (see Fig. 3 (c)).

Furthermore, it is obvious that the ratios of shear walls vary for different structure heights and different seismic precautionary intensities. For instance, buildings constructed in zones with an 8-degree seismic precautionary intensity require more shear walls than those built in zones with a 7-degree seismic precautionary intensity. Here, the 8-degree and 7-degree seismic precautionary intensities mean that the design basic acceleration values of ground motion in the corresponding zones are $0.20$~$0.30g$ and $0.10$~$0.15g$, respectively, as per the Chinese General Code for Seismic Precaution of Buildings and Municipal Engineering [69] and the Chinese Code for Seismic Design of Buildings [70]. Here, $g$ represents the gravitational acceleration. To guide the generation of the AI model to consider the above physical information, and prevent it from designing shear walls arbitrarily, a physical condition, denoted as $d$, integrating heights and seismic precautionary intensities, is embedded in the conditional transition kernel of CDM. A detailed introduction about the physical condition is presented in Subsection 5.5.1. Therefore, the modeling objective in the optimization process is changed into a new probability distribution $p_{\boldsymbol{\theta}}(\mathbf{x}_{t-1}|\mathbf{x}_t, \mathbf{y}, d)$. To simplify the expression, this type of CDM is named PCDM, *i.e.*, a physics-based conditional diffusion model. The forward diffusion process and the reverse denoising process are also retained in PCDM. The details of PCDM are delineated in the following subsections.

### *5.2. Forward diffusion process*

In the forward process, we inject a small amount of Gaussian noise $\boldsymbol{\epsilon} \in \mathbb{R}^D$ step-by-step through a $T$-step Markov chain, as shown in Fig. 4. Following [67], the transition probability is defined as

$$q(\mathbf{x}_t|\mathbf{x}_{t-1}) \triangleq \mathcal{N}\big(\mathbf{x}_t; \sqrt{1-\beta_t}\mathbf{x}_{t-1}, \beta_t \mathbf{I}_D\big). \tag{1}$$

It is noteworthy that the random variables in the Gaussian noise $\boldsymbol{\epsilon}$ are independently sampled from the standard normal distribution, implying that the mean value is 0, and the covariance matrix is an identity matrix, namely, $\boldsymbol{\epsilon} \sim \mathcal{N}(0, \mathbf{I}_D)$, where $\mathbf{I}_D$ denotes the identity matrix in $\{0,1\}^{D \times D}$. The term $\beta_t$ is the noise schedule to control the amount of noise for different timesteps, and $\beta_t \in (0,1)$ for $\forall t \in \{1,2, \dots, T\}$. In the original DDPM, $\beta_t$ is designed as a linear noise schedule. However, DDPM+ [71] has revealed that the linear schedule may cause the end of the forward process to be too noisy, which means that it cannot make a significant contribution to sample quality. Consequently, DDPM+ proposes a cosine schedule to alleviate this issue. In this paper, we also substitute the linear schedule with the cosine schedule. In the cosine schedule, two functions over timesteps, $g(t)$ and $\bar{\alpha}_t$, are initially defined:

$$g(t) \triangleq \cos^2 \left( \frac{t/T + s}{1 + s} \times \frac{\pi}{2} \right), \tag{2}$$

$$\bar{\alpha}_t \triangleq \frac{g(t)}{g(0)}, \tag{3}$$

where $s$ is a small offset. Then, $\beta_t$ can be calculated based on $\bar{\alpha}_t$:

$$\beta_t = 1 - \frac{\bar{\alpha}_t}{\bar{\alpha}_{t-1}}. \tag{4}$$

As introduced in Subsection 5.1, PCDM necessitates learning a denoising model $p_{\boldsymbol{\theta}}(\mathbf{x}_{t-1} | \mathbf{x}_t, \mathbf{y}, d)$ based on the given dataset $(\mathbf{x}_0, \mathbf{y}, d)$. To achieve this, we firstly need to obtain the noisy sample $\mathbf{x}_t$ from the clean sample $\mathbf{x}_0$. If we rely solely on Eq. $(1)$, we have to iteratively sample $t$ times to acquire $\mathbf{x}_t$, which is both time-consuming and computationally demanding because $t$ can be very large. However, owing to the favorable properties of the Gaussian distribution, we can obtain the closed-form expression of $q(\mathbf{x}_t | \mathbf{x}_0)$, allowing us to directly obtain $\mathbf{x}_t$ through a single sampling for arbitrary $t$:

$$\mathbf{x}_t \sim q(\mathbf{x}_t | \mathbf{x}_0) = \mathcal{N}(\mathbf{x}_t; \sqrt{\bar{\alpha}_t} \mathbf{x}_0, (1 - \bar{\alpha}_t) \mathbf{I}_D). \tag{5}$$

The derivation process for the above equation is detailed in Appendix A.1. As for how to sample data, the reparameterization trick [38] can be effortlessly employed to achieve this:

$$\mathbf{x}_t = \sqrt{\bar{\alpha}_t} \mathbf{x}_0 + \sqrt{1 - \bar{\alpha}_t} \boldsymbol{\epsilon}_0^t. \tag{6}$$

where $\boldsymbol{\epsilon}_0^t \sim \mathcal{N}(0, \mathbf{I}_D)$ represents the added noise from $\mathbf{x}_0$ to $\mathbf{x}_t$.

### 5.3. *Reverse denoising process*

Literally, the reverse denoising process is the reverse process of the forward diffusion process. Specifically, the objective of the reverse denoising process is to regenerate clean samples $\hat{\mathbf{x}}_0$ from $\mathbf{x}_T$ and to make $\hat{\mathbf{x}}_0$ follow the real data distribution $q_0$ as much as possible. Here, the symbols with hats denote estimated data. Consequently, the reverse denoising process can be regarded as an inference process or a sampling process. A clear illustration of this process is visually represented in Fig. 4.

Clearly, to execute the reverse process, the prerequisite is to determine the probability distribution of $\mathbf{x}_T$ and the value of the unknown $T$. Revisiting Eq. $(6)$, we can find that as $t$ increases, $g(t)$ gradually approaches 0, implying that $\bar{\alpha}_t$ is nearing 0. As a result, $\mathbf{x}_0$ progressively loses its distinguishable features and converges towards a standard normal distribution when $t$ becomes sufficiently large. This means when $T$ is a large number, approximately, $\mathbf{x}_T \sim \mathcal{N}(0, \mathbf{I}_D)$. Following DDPM [27], $T$ is set to $1000$, and at this point, the difference between the distribution of $\mathbf{x}_T$ and Gaussian noise $\mathcal{N}(0, \mathbf{I}_D)$ becomes negligible. To show the difference, we denote the sampled data from $\mathcal{N}(0, \mathbf{I}_D)$ as $\hat{\mathbf{x}}_T$.

After obtaining $\hat{\mathbf{x}}_T$, clearly, if we can further ascertain $q(\mathbf{x}_{t-1} | \mathbf{x}_t)$, we can easily achieve this reconstruction goal through a reverse Markov chain. Thus, another critical question concerns how to compute $q(\mathbf{x}_{t-1} | \mathbf{x}_t)$.

Frustratingly, it is intractable to obtain $q(\mathbf{x}_{t-1}|\mathbf{x}_t)$ [67]. However, DDPM reveals that if this distribution is conditioned on $\mathbf{x}_0$, it becomes tractable:

$$q(\mathbf{x}_{t-1}|\mathbf{x}_t, \mathbf{x}_0) = \mathcal{N}\left(\mathbf{x}_{t-1}; \tilde{\boldsymbol{\mu}}_t(\mathbf{x}_t, \mathbf{x}_0), \widetilde{\boldsymbol{\Sigma}}_t\right), \tag{7}$$

where

$$\tilde{\boldsymbol{\mu}}_t(\mathbf{x}_t, \mathbf{x}_0) = \frac{\sqrt{1-\beta_t}(1-\bar{\alpha}_{t-1})}{1-\bar{\alpha}_t}\mathbf{x}_t + \frac{\sqrt{\bar{\alpha}_{t-1}}\beta_t}{1-\bar{\alpha}_t}\mathbf{x}_0, \tag{8}$$

and

$$\widetilde{\boldsymbol{\Sigma}}_t = \tilde{\beta}_t \mathbf{I}_D, \tilde{\beta}_t = \frac{1-\bar{\alpha}_{t-1}}{1-\bar{\alpha}_t}\beta_t. \tag{9}$$

The key mathematical intuition behind $q(\mathbf{x}_{t-1}|\mathbf{x}_t, \mathbf{x}_0)$ is that, firstly, $q(\mathbf{x}_t|\mathbf{x}_{t-1}) = q(\mathbf{x}_t|\mathbf{x}_{t-1}, \mathbf{x}_0)$ due to the Markov property (*i.e.*, the memoryless property). Secondly, based on Bayes' rule, $q(\mathbf{x}_t|\mathbf{x}_{t-1}, \mathbf{x}_0)$ can be reformulated as

$$q(\mathbf{x}_t|\mathbf{x}_{t-1}, \mathbf{x}_0) = \frac{q(\mathbf{x}_t|\mathbf{x}_0)q(\mathbf{x}_{t-1}|\mathbf{x}_t, \mathbf{x}_0)}{q(\mathbf{x}_{t-1}|\mathbf{x}_0)}. \tag{10}$$

According to Eq. (5), $q(\mathbf{x}_t|\mathbf{x}_0)$ and $q(\mathbf{x}_{t-1}|\mathbf{x}_0)$ can both be parameterized, and Eq. (1) gives the definition of $q(\mathbf{x}_t|\mathbf{x}_{t-1})$, that is, the definition of $q(\mathbf{x}_t|\mathbf{x}_{t-1}, \mathbf{x}_0)$. Consequently, $q(\mathbf{x}_{t-1}|\mathbf{x}_t, \mathbf{x}_0)$ can be parameterized after a series of mathematical deductions. The detailed derivations of Eqs. (7), (8), and (9) are given in Appendix A.2.

Although the closed-form solution of $q(\mathbf{x}_{t-1}|\mathbf{x}_t, \mathbf{x}_0)$ has been obtained, we still cannot sample new data starting from $\hat{\mathbf{x}}_T$ based on it, as $\mathbf{x}_0$ remains unknown during the sampling stage. At this point, AI comes into play. The direct idea is that if we can model a transition distribution $p_{\boldsymbol{\theta}}(\mathbf{x}_{t-1}|\mathbf{x}_t, \mathbf{y}, d)$ utilizing a neural network to approximate $q(\mathbf{x}_{t-1}|\mathbf{x}_t, \mathbf{x}_0)$, then we can achieve the aforementioned goal through a new reverse Markov chain. Since $\mathbf{y}$ and $d$ are pre-given, and $\hat{\mathbf{x}}_T$ can be sampled from $\mathcal{N}(0, \mathbf{I}_D)$, and if we further assume $p_{\boldsymbol{\theta}}(\mathbf{x}_{t-1}|\mathbf{x}_t, \mathbf{y}, d)$ has been established, $\hat{\mathbf{x}}_{T-1}$ can be easily obtained by sampling from $p_{\boldsymbol{\theta}}(\hat{\mathbf{x}}_{T-1}|\hat{\mathbf{x}}_T, \mathbf{y}, d)$. By repeating this step, the clean sample $\hat{\mathbf{x}}_0$ can be ultimately obtained. This is the overall view of the reverse denoising process, summarized in Algorithm 1.

**Algorithm 1** Reverse denoising process.

**Require:** canvas $\mathbf{y}$, physical condition $d$, and $p_{\boldsymbol{\theta}}(\mathbf{x}_{t-1}|\mathbf{x}_t, \mathbf{y}, d)$
1: Sample $\hat{\mathbf{x}}_T \sim \mathcal{N}(0, \mathbf{I}_D)$
2: **for** $t = T, T-1, \ldots, 1$ **do**
3: Sample $\hat{\mathbf{x}}_{t-1} \sim p_{\boldsymbol{\theta}}(\hat{\mathbf{x}}_{t-1}|\hat{\mathbf{x}}_t, \mathbf{y}, d)$
4: **end for**
5: **return** $\hat{\mathbf{x}}_0$

Given that sampling is random, if we execute the reverse denoising process multiple times, $\hat{\mathbf{x}}_0$ obtained each time will differ even if $p_{\boldsymbol{\theta}}$, $\mathbf{y}$, and $d$ are all fixed. This suggests that for a well-trained PCDM, we can obtain an "infinite" number of design drawings for the same input conditions $(\mathbf{y}, d)$, and at the same time, the generated designs all meet the conditions. Some examples are presented in Subsection 7.4.1 to demonstrate the diversity of results generated by PCDM. It is important to note that the diverse design is impossible for StructGAN because the prediction is fixed if the model is fixed, and the conventional DMs cannot achieve it either since they do not fuse the physical conditions.

As regards how to train a neural network to represent $p_{\boldsymbol{\theta}}(\mathbf{x}_{t-1}|\mathbf{x}_t, \mathbf{y}, d)$, the subsequent subsection provides a detailed introduction.

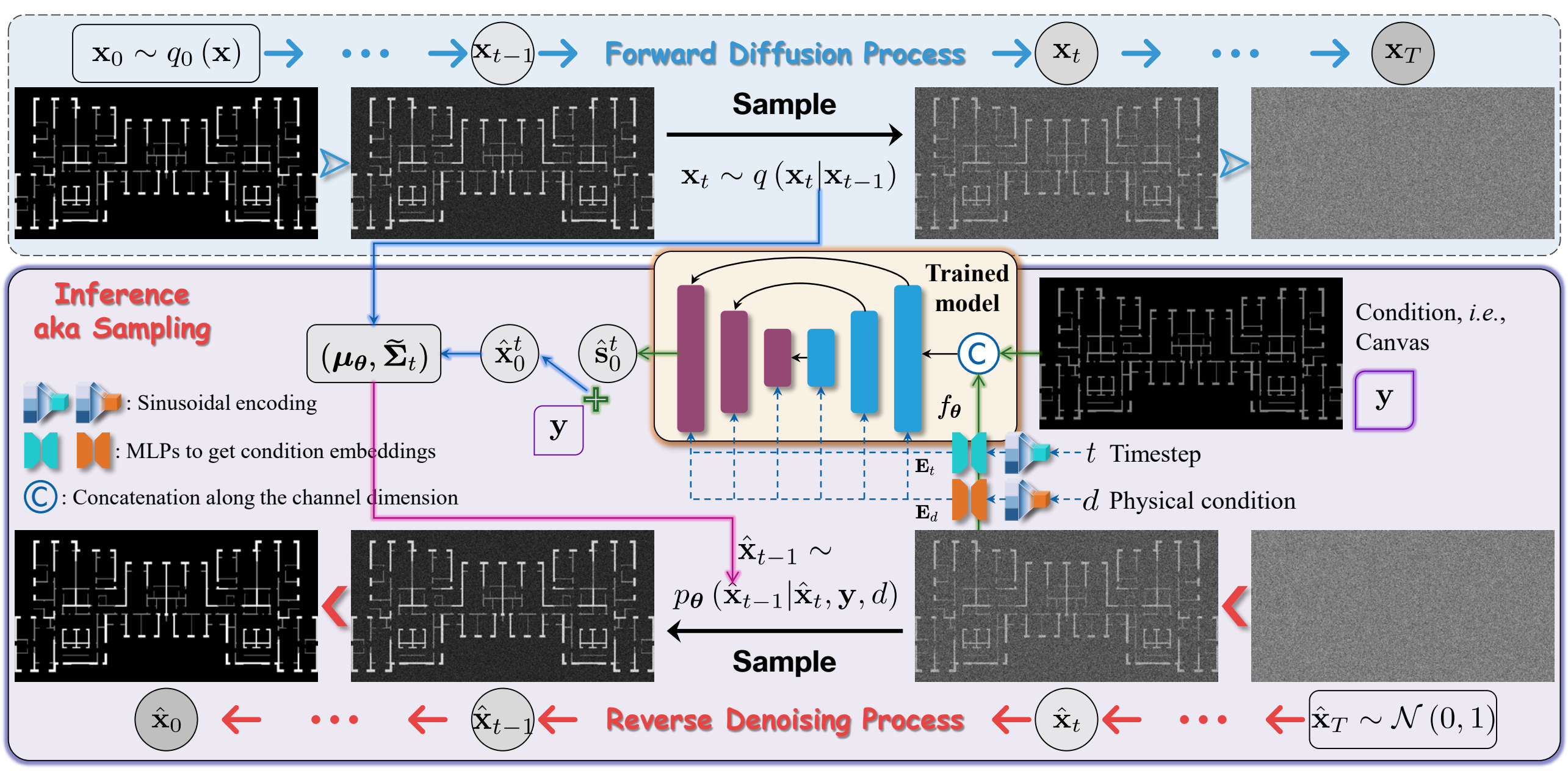

**Fig. 4.** Schematic diagram of the forward diffusion process and the reverse denoising process.

### 5.4. *Optimization process: Parameterization of* $p_{\boldsymbol{\theta}}(\boldsymbol{x}_{t-1}|\boldsymbol{x}_t, \boldsymbol{y}, d)$ *and training algorithm*

In the optimization process, we want to effectively model $p_{\boldsymbol{\theta}}(\mathbf{x}_{t-1}|\mathbf{x}_t, \mathbf{y}, d)$ to make it close to $q(\mathbf{x}_{t-1}|\mathbf{x}_t, \mathbf{x}_0)$ as far as possible.

Since $q(\mathbf{x}_{t-1}|\mathbf{x}_t, \mathbf{x}_0)$ is a Gaussian distribution, the most straightforward idea to model $p_{\boldsymbol{\theta}}(\mathbf{x}_{t-1}|\mathbf{x}_t, \mathbf{y}, d)$ is modeling it to be a Gaussian distribution as well:

$$p_{\boldsymbol{\theta}}(\mathbf{x}_{t-1}|\mathbf{x}_t, \mathbf{y}, d) \triangleq \mathcal{N}\big(\mathbf{x}_{t-1}; \boldsymbol{\mu}_{\boldsymbol{\theta}}(\mathbf{x}_t, t, \mathbf{y}, d), \boldsymbol{\Sigma}_{\boldsymbol{\theta}}(\mathbf{x}_t, t, \mathbf{y}, d)\big), \tag{11}$$

where $\mu_{\boldsymbol{\theta}}$ denotes the mean vector of $p_{\boldsymbol{\theta}}$, and $\boldsymbol{\Sigma}_{\boldsymbol{\theta}}$ represents the covariance matrix of $p_{\boldsymbol{\theta}}$. Because $\mathbf{x}_t$ is a vector and does not include timestep information, $t$ is explicitly embedded into $\boldsymbol{\mu}_{\boldsymbol{\theta}}$ and $\boldsymbol{\Sigma}_{\boldsymbol{\theta}}$ to introduce time information. Intuitively, we can minimize the distance between $\boldsymbol{\mu}_{\boldsymbol{\theta}}$ and $\tilde{\boldsymbol{\mu}}_t$ and the distance between $\boldsymbol{\Sigma}_{\boldsymbol{\theta}}$ to $\widetilde{\boldsymbol{\Sigma}}_t$ to make $p_{\boldsymbol{\theta}}(\mathbf{x}_{t-1}|\mathbf{x}_t, \mathbf{y}, d)$ approach $q(\mathbf{x}_{t-1}|\mathbf{x}_t, \mathbf{x}_0)$. DDPM finds that if we let $\boldsymbol{\Sigma}_{\boldsymbol{\theta}}$ be equal to $\widetilde{\boldsymbol{\Sigma}}_t$, namely, $\boldsymbol{\Sigma}_{\boldsymbol{\theta}}$ is not learnable, the sample quality will be almost unaffected, but the question is simplified into "how to model $\boldsymbol{\mu}_{\boldsymbol{\theta}}$ to make it close to $\tilde{\boldsymbol{\mu}}_t$". Therefore, $\boldsymbol{\Sigma}_{\boldsymbol{\theta}}$ is set to $\widetilde{\boldsymbol{\Sigma}}_t$ here, following DDPM.

A straightforward way to model $\boldsymbol{\mu}_{\boldsymbol{\theta}}$ is to directly build a neural network to represent it, $\boldsymbol{\mu}_{\boldsymbol{\theta}} = NN_{\boldsymbol{\theta}}(\mathbf{x}_t, t, \mathbf{y}, d)$, and to train the network to minimize the distance between $\boldsymbol{\mu}_{\boldsymbol{\theta}}$ and $\tilde{\boldsymbol{\mu}}_t$. The distribution $p_{\boldsymbol{\theta}}$ can be well modeled. The classical mean squared error (MSE) loss is employed to measure the similarity:

$$\mathcal{L}_{\text{MSE}} = \mathbb{E}_{t\sim\text{Uniform}[1,T], \mathbf{x}_0\sim q_0, \boldsymbol{\epsilon}\sim\mathcal{N}(0,\mathbf{I}_D)} \|\tilde{\boldsymbol{\mu}}_t - \boldsymbol{\mu}_{\boldsymbol{\theta}}\|_2^2, \tag{12}$$

where $\text{Uniform}[1, T]$ means the uniform distribution. Since $\tilde{\boldsymbol{\mu}}_t$ and $\boldsymbol{\mu}_{\boldsymbol{\theta}}$ are both defined over $t$, $\mathbf{x}_0$, and $\boldsymbol{\epsilon}$ (see Eqs. (6) and (8)), $\mathcal{L}$ is an expectation value over $t$, $\mathbf{x}_0$, and $\boldsymbol{\epsilon}$ to eliminate randomness.

Further, since $\mathbf{x}_t$ is known (it is a condition of $p_{\boldsymbol{\theta}}$) and $\mathbf{x}_t$ is also included in $\tilde{\boldsymbol{\mu}}_t$ (see Eq. (8)), we can consider retaining $\mathbf{x}_t$ and estimating the unknown part in $\tilde{\boldsymbol{\mu}}_t$ rather than estimating the whole $\tilde{\boldsymbol{\mu}}_t$ to reduce the prediction difficulty. Following this idea, we have two choices. The first option, as shown in Fig. 5 (a), is to construct a neural network $\hat{\mathbf{x}}_0^t = \boldsymbol{x}_{\boldsymbol{\theta}}(\mathbf{x}_t, t, \mathbf{y}, d)$ to approximate the clean line drawing $\mathbf{x}_0$, where $\hat{\mathbf{x}}_0^t$ represents the estimated line drawing when the time index is $t$, and the new loss function is

$$\mathcal{L}_{\text{MSE}}^{\mathbf{x}_0} = \mathbb{E}_{t,\mathbf{x}_0,\boldsymbol{\epsilon}} \|\mathbf{x}_0 - \hat{\mathbf{x}}_0^t\|_2^2. \tag{13}$$

Next, $\boldsymbol{\mu}_{\boldsymbol{\theta}}$ can be modeled according to $\hat{\mathbf{x}}_0^t$ (refer to Eq. (8)):

$$\boldsymbol{\mu}_{\boldsymbol{\theta}}=\frac{\sqrt{1-\beta_t}(1-\bar{\alpha}_{t-1})}{1-\bar{\alpha}_t}\mathbf{x}_t+\frac{\sqrt{\bar{\alpha}_{t-1}}\beta_t}{1-\bar{\alpha}_t}\hat{\mathbf{x}}_0^t. \tag{14}$$

Then, plug Eq. (14) into Eq. (11) to model the whole transition distribution:

$$p_{\boldsymbol{\theta}}(\mathbf{x}_{t-1}|\mathbf{x}_t,\mathbf{y},d)=\mathcal{N}\left(\mathbf{x}_{t-1};\frac{\sqrt{1-\beta_t}(1-\bar{\alpha}_{t-1})}{1-\bar{\alpha}_t}\mathbf{x}_t+\frac{\sqrt{\bar{\alpha}_{t-1}}\beta_t}{1-\bar{\alpha}_t}\hat{\mathbf{x}}_0^t,\tilde{\beta}_t\mathbf{I}_D\right). \tag{15}$$

According to this transition, we can run the reverse denoising process (see Subsection 5.3) to generate new samples.

As shown in Fig. 5 (b), the second choice is predicting noise. We first plug Eq. (6) into Eq. (8) to eliminate $\mathbf{x}_0$:

$$\tilde{\boldsymbol{\mu}}_t=\frac{1}{\sqrt{1-\beta_t}}\left(\mathbf{x}_t-\frac{\beta_t}{\sqrt{1-\bar{\alpha}_t}}\boldsymbol{\epsilon}_0^t\right). \tag{16}$$

The derivation of the above formula is detailed in Appendix A.3. Imitating the first choice, we can build a neural network $\hat{\boldsymbol{\epsilon}}_0^t=\boldsymbol{\epsilon}_{\boldsymbol{\theta}}(\mathbf{x}_t,t,\mathbf{y},d)$ to approximate $\boldsymbol{\epsilon}_0^t$, where $\hat{\boldsymbol{\epsilon}}_0^t$ denotes the estimated value to $\boldsymbol{\epsilon}_0^t$, and the corresponding loss function is

$$\mathcal{L}_{\text{MSE}}^{\boldsymbol{\epsilon}}=\mathbb{E}_{t,\mathbf{x}_0,\boldsymbol{\epsilon}}\|\boldsymbol{\epsilon}_0^t-\hat{\boldsymbol{\epsilon}}_0^t\|_2^2. \tag{17}$$

Referring to Eq. (16), $\boldsymbol{\mu}_{\boldsymbol{\theta}}$ can be further modeled by $\hat{\boldsymbol{\epsilon}}_0^t$:

$$\boldsymbol{\mu}_{\boldsymbol{\theta}}=\frac{1}{\sqrt{1-\beta_t}}\left(\mathbf{x}_t-\frac{\beta_t}{\sqrt{1-\bar{\alpha}_t}}\hat{\boldsymbol{\epsilon}}_0^t\right), \tag{18}$$

Then, we can get $p_{\boldsymbol{\theta}}$:

$$p_{\boldsymbol{\theta}}(\mathbf{x}_{t-1}|\mathbf{x}_t,\mathbf{y},d)=\mathcal{N}\left(\mathbf{x}_{t-1};\frac{1}{\sqrt{1-\beta_t}}\left(\mathbf{x}_t-\frac{\beta_t}{\sqrt{1-\bar{\alpha}_t}}\hat{\boldsymbol{\epsilon}}_0^t\right),\tilde{\beta}_t\mathbf{I}_D\right). \tag{19}$$

Obviously, the two options are theoretically equivalent because the optimization objectives are both to make $\boldsymbol{\mu}_{\boldsymbol{\theta}}$ close to $\tilde{\boldsymbol{\mu}}_t$, namely, $p_{\boldsymbol{\theta}}(\mathbf{x}_{t-1}|\mathbf{x}_t,\mathbf{y},d)$ close to $q(\mathbf{x}_{t-1}|\mathbf{x}_t,\mathbf{x}_0)$. However, it is worth noting that in practice, the emphasis of these two choices is different. The first option predicts the clean line drawing $\mathbf{x}_0$, which is definite, therefore, the option focuses on ***similarity***. The second choice estimates noise, and noise is random, thus, the option pays more attention to ***variability***. Therefore, the second choice is applied in almost all DM-based methods for synthesis of general images, such as DDPM, DDPM+, DDIM [44], and Stable Diffusion [26]. This is because these methods want to generate diverse images across various image categories, which is aligned with the emphasis of the second choice. Nonetheless, the research scope of this study is image generation for the specific category, namely, structural design drawings. Our goal is that DMs can concentrate on ***similarity*** to generate realistic structural design drawings. Therefore, this paper adopts the first prediction choice. The ablation studies in Subsection 7.3 provide experimental evidence for our analysis.

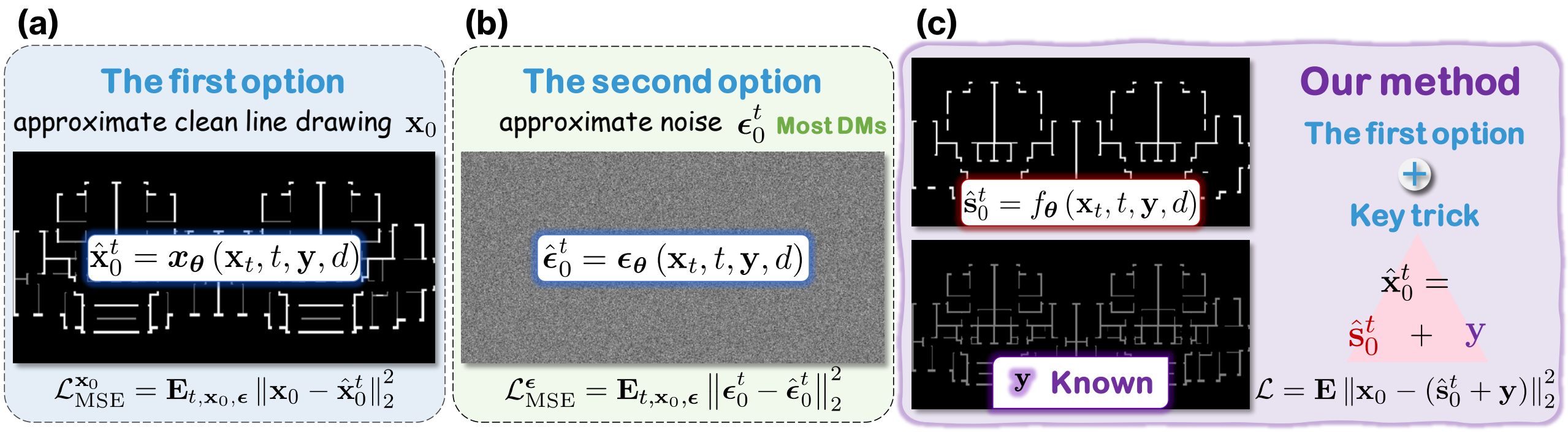

**Fig. 5.** Different prediction options: (a) the first option approximating a clean image, (b) the second option approximating noise, and (c) our proposed technique based on the first option.

Then, the problem is further turned into "how to obtain a more accurate estimation of $\mathbf{x}_0$". This paper designs a clever technique as shown in Fig. 5 (c). First, we notice that $\mathbf{x}_0$ can be divided into two parts:

$$\mathbf{x}_0 = \mathbf{s}_0 + \mathbf{y}, \tag{20}$$

where $\mathbf{s}_0$ only contains the shear walls in $\mathbf{x}_0$, and the other pixels in $\mathbf{x}_0$ are set to the background. For simplicity, $\mathbf{s}_0$ is named the shear wall drawing. $\mathbf{y}$ denotes the infill wall image (or canvas, refer to Section 4) and is known when training the network. It is worth noting that for $\mathbf{x}_0$ and $\mathbf{y}$, the value of the background is $-1$, the value of the infill wall pixels is $0$, and the value of the pixels representing shear walls is $1$. Thus, $\mathbf{x}_0 \subset \{-1{,}0{,}1\}$, and $\mathbf{y} \subset \{-1{,}0\}$. For $\mathbf{s}_0$, we set the value of shear wall pixels to $1$ and other pixel values set to $0$. Based on the above rules, Eq. $(20)$ can be obtained.

Further, imitating the operation in Eq. $(14)$, because $\mathbf{y}$ is known, we can predict $\mathbf{s}_0$ rather than estimate the entire $\mathbf{x}_0$ to reduce the prediction difficulty. Based on this idea, the neural network can be constructed as:

$$\hat{\mathbf{s}}_0^t = f_{\boldsymbol{\theta}}(\mathbf{x}_t, t, \mathbf{y}, d), \tag{21}$$

where $\hat{\mathbf{s}}_0^t$ is the estimation of $\mathbf{s}_0$ when the time index is $t$. The new MSE loss function is still Eq. $(13)$, but the only difference is that $\hat{\mathbf{x}}_0^t = \boldsymbol{x}_{\boldsymbol{\theta}}(\mathbf{x}_t, t, \mathbf{y}, d)$ is substituted with $\hat{\mathbf{x}}_0^t = \hat{\mathbf{s}}_0^t + \mathbf{y}$:

$$\mathcal{L}_{\text{MSE}}^{\mathbf{x}_0} = \mathbb{E}_{t,\mathbf{x}_0,\boldsymbol{\epsilon}} \|\mathbf{x}_0 - \hat{\mathbf{x}}_0^t\|_2^2 = \mathbb{E}_{t,\mathbf{x}_0,\boldsymbol{\epsilon}} \|\mathbf{x}_0 - (\hat{\mathbf{s}}_0^t + \mathbf{y})\|_2^2. \tag{22}$$

The above algorithm is presented in Fig. 5 (c).

It is important to note that the final prediction of this algorithm is still the line drawing $\mathbf{x}_0$ and is equivalent to the vanilla first option. However, in contrast to the vanilla first option, the advancement of this trick can be understood from two perspectives: (1) In the first option, the generation goal is $\mathbf{x}_0$, and the model needs to produce shear walls and infill walls simultaneously. The proposed trick changes the goal to the shear wall drawing $\mathbf{s}_0$, then the AI model only needs to consider the generation of one type of wall. The concentration and simplification of prediction targets is beneficial for generating high-quality shear wall drawings. (2) The first option directly predicts $\hat{\mathbf{x}}_0^t$ but neglects the known $\mathbf{y}$. This trick fully incorporates the known information $\mathbf{y}$.

In addition to leveraging $\mathcal{L}_{\text{MSE}}^{\mathbf{x}_0}$, to further enhance the similarity in the generated results, the VGG loss function $\mathcal{L}_{\text{VGG}}$ is also adopted to train PCDM together with $\mathcal{L}_{\text{MSE}}^{\mathbf{x}_0}$. The VGG loss has been extensively used in deep generative models to improve perceptual similarity [21] and is defined as

$$\mathcal{L}_{\text{VGG}} = \mathbb{E}_{t,\mathbf{x}_0,\boldsymbol{\epsilon},\ell} w_\ell \|\mathcal{VGG}_\ell(\mathbf{x}_0) - \mathcal{VGG}_\ell(\hat{\mathbf{x}}_0^t)\|_1, \tag{23}$$

where $\mathcal{VGG}_\ell(\cdot)$ means the $\ell$-th layer feature decoded by VGGNet, and $w_\ell$ denotes the weight of the $\ell$-th layer. For simplicity, $w_\ell$ is set to 1 for all $\ell$. Then, the total loss function can be formulated as follows:

$$\mathcal{L} = \mathcal{L}_{\text{MSE}}^{\mathbf{x}_0} + 0.1 \times \mathcal{L}_{\text{VGG}}. \tag{24}$$

The above algorithm is summarized in Algorithm 2 for clearer explanation. Besides, the upper portion of Fig. 6 also introduce the training process in the form of graphics, which is also worth referring to for a more vivid understanding of the entire optimization process.

**Algorithm 2** Training of PCDM.

**Require:** denoiser network $f_{\boldsymbol{\theta}}$, noise schedule $\{\bar{\alpha}_t\}_{t=1}^T$ and $\{\beta_t\}_{t=1}^T$
1: Initialize model parameters $\boldsymbol{\theta}$
2: **while** not converged **do**
3: **Input** line drawing $\mathbf{x}_0$, canvas $\mathbf{y}$, physical condition $d$
4: $t$~Uniform$\{1, .., T\}$
5: $\boldsymbol{\epsilon}_0^t$~$\mathcal{N}(0, \mathbf{I}_D)$
6: $\mathbf{x}_t \leftarrow \sqrt{\bar{\alpha}_t}\mathbf{x}_0 + \sqrt{1-\bar{\alpha}_t}\boldsymbol{\epsilon}_0^t$ ▷ Run forward diffusion process to get the noisy sample
7: $\hat{\mathbf{s}}_0^t \leftarrow f_{\boldsymbol{\theta}}(\mathbf{x}_t, t, \mathbf{y}, d)$ ▷ Predict the shear wall drawing
8: $\hat{\mathbf{x}}_0^t = \hat{\mathbf{s}}_0^t + \mathbf{y}$ ▷ Get the estimated line drawing
8: Perform gradient descent step on
$\nabla_{\boldsymbol{\theta}}(\mathcal{L}_{\text{MSE}}^{\mathbf{x}_0} + 0.1 \times \mathcal{L}_{\text{VGG}})$

9: **end while**
10: **return** $f_{\boldsymbol{\theta}}$

In this subsection, we affirm the correctness of the parameterization of $p_{\boldsymbol{\theta}}(\mathbf{x}_{t-1}|\mathbf{x}_t, \mathbf{y}, d)$ from an intuitive perspective. For those seeking more detailed validation, rigorous proofs are provided in Appendix B. However, if readers find this material challenging, skipping Appendix B will not impede the understanding of this research or the principles of diffusion models (DMs).

### *5.5. Optimization process: Design of the neural network*

The subsection primarily focuses on how to build a denoising network $f_{\boldsymbol{\theta}}$ to obtain $\hat{\mathbf{s}}_0^t$ given the data tuple $(\mathbf{x}_t, t, \mathbf{y}, d)$. Specifically, three key tasks are researched and addressed in this subsection. First, Subsection 5.1 analyzes the significance of fusing the physical condition $d$, and this subsection investigates how to effectively model $d$. Second, the conditions, $t$, $d$ and $\mathbf{y}$, are totally different in terms of data dimensions and the nature of information they provide, indicating that they cannot be embedded into $f_{\boldsymbol{\theta}}$ in a consistent way. Therefore, this subsection studies how to effectively incorporate the varied conditions into $f_{\boldsymbol{\theta}}$. Last, this subsection provides a detailed design for fusing cross-domain data.

#### *5.5.1. Physical condition and its superiority*

**Physical condition in PCDM.** As discussed in Subsection 5.1, the ratios of shear walls vary significantly for different seismic precautionary intensities and structure heights. The dataset provided in StructGAN[†], which is named Original-dataset in this paper, considers the buildings in 7-degree and 8-degree seismic precautionary zones since most regions in China belong to these two categories, and the design drawings of the buildings constitutes the Group7 sub-dataset (for the buildings in 7-degree seismic precautionary zones) and the Group8 sub-dataset (for the buildings in 8-degree seismic precautionary zones) of Original-dataset, respectively. Additionally, the height of structures is also a significant factor in the design of shear walls for 7-degree seismic precautionary zones [1]. Thus, StructGAN further divides the Group7 sub-dataset into two sub-datasets: Group7-H1 for buildings with heights less than or equal to 50 $m$ and Group7-H2 for those buildings whose heights $> 50\ m$, in accordance with the structural height regulations in the Chinese Technical Specification for Concrete Structures of Tall Buildings [72]. As a result, there are 3 sub-datasets in Original-dataset: Group7-H1 (7-degree and heights $\leq 50\ m$), Group7-H2 (7-degree and heights $> 50\ m$), and Group8 (8-degree). PCDM is trained on the same dataset (slight modifications are conducted, referring to Subsection 6.1) and assigns distinct physical conditions to each sub-datasets to consider their differences. Specifically, first, 7-degree and 8-degree zones mean the structures in the zones usually are conducted seismic design according to the requirements of the 7-degree and 8-degree seismic precautionary intensities (A few exceptions can refer to the Chinese General Code [69]). Further, the requirements are reflected in the design basic acceleration values of ground motion: the values for 7-degree and 8-degree intensities are 0.10~$0.15g$ and 0.20~$0.30g$, respectively. Given that $g = 10\ m/s^2$, the design basic acceleration values are changed to 1.0~1.5 $m/s^2$ and 2.0~3.0 $m/s^2$. Therefore, we simply assign 1.0 to Group7-H1, 1.5 to Group7-H2, and 2.5 to Group8 as physical conditions. Clearly, if physical conditions are carefully fine-tuned to more finely classify buildings, for example, in addition to the acceleration of ground motion, we can consider more factors, which are usually taken into account by engineers in the design stage, such as design characteristic period of ground motion, construction sites, the importance of architectures, or natural periods of structures (refer to [69]), the design performance of the AI models has the potential to be closer to that of humans. Here, we leave this investigation to future work.

[†] https://github.com/wenjie-liao/StructGAN_v1

**Superiority of PCDM: Diverse and creative designs.** To avoid the impact of data variability among sub-datasets, StructGAN separately trains three models based on the three sub-datasets. However, the training strategy has two significant drawbacks. First, the amount of data is insufficient for each model, which may result in overfitting and hinder the performance of the models. Second, it neglects the beneficial impact of data similarity among sub-datasets. Therefore, StructGAN-AE [3] introduces a pre-trained strategy to alleviate these issues, which includes two stages. Stage 1: pre-training stage. Models are trained based on mixed data consisting of the three sub-datasets. Stage 2: individual training stage. The pre-trained models are individually re-trained on each of the three sub-datasets. Nevertheless, this solution is still not elegant as on the one hand, it requires at least four training sessions (once for pre-training and three times for individual training) to enable models to "***see***" all the data, and on the other hand, it needs to save three models (one model per sub-dataset). A crucial question thus arises: ***is there a method where a single model can just "see" and analyze all the data by once training?*** PCDM can achieve this. Because PCDM assigns different physical conditions to the images from different sub-datasets, it can be trained on the entire dataset (*i.e.*, a mixed dataset comprising all the three sub-datasets) in the training stage. In the inference stage, PCDM can infer correct structural designs for different sub-datasets by inputting physical conditions.

Additionally, the information fusion provides PCDM with two unique functions. First, a single trained PCDM can generate multiple designs for the same input drawings and the corresponding physical conditions simultaneously. The function is named diverse design and presented in Subsection 7.4.1. Second, for the same input drawing, PCDM can generate different designs if the input physical condition of the drawing changes and, at the same time can ensure that the results meet the changed physical conditions. For example, if a design drawing originally belongs to Group7-H1, PCDM can generate the correct structural designs when the input condition is 1.0. If the physical condition changes to 2.5, PCDM can still generate a structural design drawing that satisfies the new condition. This new function brings significant flexibility and creativity to design and is termed creative design, as later demonstrated in Subsection 7.4.2. It is important to note that these two functions cannot be achieved by StructGAN and its variants due to the inherent limitations of GAN. Conventional DMs also cannot achieve these functions as they do not incorporate physical conditions.

#### *5.5.2. Condition embedding*

Conditions in CDMs can usually be divided into two categories: global conditions and local conditions. Global conditions, such as text descriptions and temporal data, offer global information, while local conditions, like sketches, depth maps, and masks, provide refined guidance considering local details. Some studies [73][74] suggest that global conditions are suitable to be projected into vector embeddings and then plug the latent embeddings into convolution blocks or attention blocks to provide high-density control. On the other hand, local conditions are suitable to be concatenated with noisy samples and are sent into neural networks, which can ensure spatial sizes remain unchanged to preserve complete and detailed information.

As previously analyzed, PCDM considers three conditions: the timestep $t$, the canvas $\mathbf{y}$, and the physical condition $d$, among which $t$ and $d$ belong to global conditions, and $\mathbf{y}$ is a local condition. Therefore, as shown in Fig. 6, for $t$ and $d$, we first map them into latent embeddings and then plug the embeddings into residual blocks (*i.e.*, Res-Block or RB in Fig. 6) and attention blocks (*i.e.*, AB in Fig. 6). Residual blocks and attention blocks are expounded in the next subsection. The mapping of global conditions involves two steps, as shown in the middle parts of Fig. 4 and the top parts of Fig. 6. The first step is sinusoidal encoding, which is borrowed from Transformer [75]. Taking $d$ as an example, the sinusoidal approach encodes $d$ into a $D_d^{\mathrm{En}}$-dimensional vector $\mathbf{En}_d \in \mathbb{R}^{D_d^{\mathrm{En}}}$:

$$\mathbf{En}_d[i] = \begin{cases} \cos\left(d \times P^{-\frac{i-1}{\frac{D_d^{\mathrm{En}}}{2}}}\right) & 1 \le i \le \frac{D_d^{\mathrm{En}}}{2} \\ \sin\left(d \times P^{-\frac{\left(i-\frac{D_d^{\mathrm{En}}}{2}\right)-1}{\frac{D_d^{\mathrm{En}}}{2}}}\right) & \frac{D_d^{\mathrm{En}}}{2}+1 \le i \le D_d^{\mathrm{En}} \end{cases}, \tag{25}$$

where $D_d^{\mathrm{En}}$ is even and set to 32 in all the experiments. $P$ is a period constant set to 10000. The coding of time, $\mathbf{En}_t \in \mathbb{R}^{D_t^{\mathrm{En}}}$, can be obtained by the same calculations. Since the variation of $t$ is greater than that of $d$, we use a longer encoding to represent $t$, which is denoted as $D_t^{\mathrm{En}}$, and $D_t^{\mathrm{En}}$ is set to $3D_d^{\mathrm{En}}$. Subsequently, the second step is to embed the two vectors into latent space through two MLPs. Specifically, $\mathbf{En}_t$ is mapped into $\mathbf{E}_t \in \mathbb{R}^{D_t}$, and $\mathbf{En}_d$ is projected into $\mathbf{E}_d \in \mathbb{R}^{D_d}$. For simplicity, $D_t$ is set to $D_t^{\mathrm{En}}$ and $D_d = D_d^{\mathrm{En}}$. Here, $\mathbf{E}_t$ and $\mathbf{E}_d$ represent the latent embeddings of $t$ and $d$, respectively. For the local condition $\mathbf{y}$, the plugging operations are more straightforward. Initially, $\mathbf{y}$ is concatenated with the noisy sample $\mathbf{x}_t$ to obtain the composite embedding $(\mathbf{x}_t \Cup \mathbf{y})$, where $\Cup$ denotes a concatenation operation along the channel dimension. Subsequently, the embedding is directly sent to the denoising network.

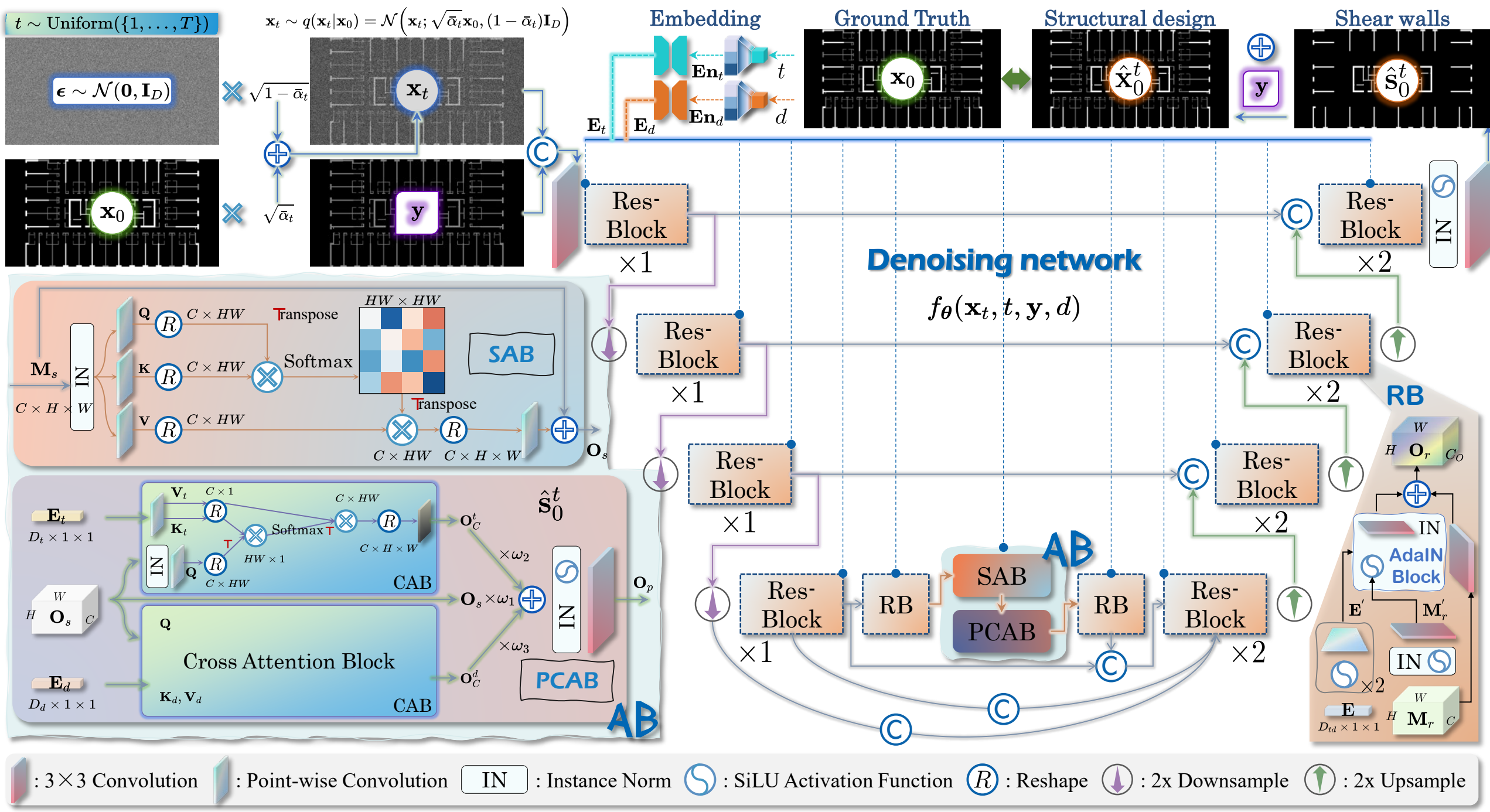

**Fig. 6.** Illustration of the entire optimization process.

### 5.5.3. *Denoising network*

U-shape networks represented by U-Net [76] have achieved significant success in DMs [24][26][27][28][44][71] due to their simplicity and effectiveness in feature fusion [77] and have almost become the de-facto standard backbone framework for DMs [78]. In light of these successful applications, the denoising network $f_{\boldsymbol{\theta}}(\mathbf{x}_t, t, \mathbf{y}, d)$ is also designed as a U-structure, as shown in Fig. 6. The core components are the residual blocks (Res-Block or RB) and the attention block (AB) of the network, and details of the two modules are delineated below.

**Res-Block (or RB).** Res-Blocks aim to extract and fuse features from $(\mathbf{x}_t \Cup \mathbf{y})$ and $(t, d)$. Following the convention of DMs [71], the input of a Res-Block consists of 2 parts: the first part comes from previous blocks and is denoted as $\mathbf{M}_r \in \mathbb{R}^{C \times H \times W}$, and the second part is the embedding of global conditions to strengthen the guidance effect of global conditions and is denoted as $\mathbf{E}$. In PCDM, $\mathbf{E}$ is the concatenation of $\mathbf{E}_t$ and $\mathbf{E}_d$:

$$\mathbf{E} = \mathbf{E}_t \Cup \mathbf{E}_d, \mathbf{E} \in \mathbb{R}^{D_{td}}, \tag{26}$$

where $D_{td} = D_t + D_d$. Subsequently, $\mathbf{E}$ is expanded to the matrix $\mathbf{E} \in \mathbb{R}^{D_{td}\times 1\times 1}$ to align with $\mathbf{M}_r$ at the dimensional level to facilitate subsequent calculations. Then, $\mathbf{E}$ goes through two layers (each layer includes a SiLU function [79] + a point-wise convolution) to change the channel into $2 \times C_O$, where $C_O$ denotes the output channel. The intermediate tensor is denoted as $\mathbf{E}' \in \mathbb{R}^{(2\times C_O)\times 1\times 1}$. Similarly, after an Instance Normalization (IN) layer, a SiLU function, and a 3 × 3 convolution, an intermediate tensor of $\mathbf{M}_r$ is obtained, denoted as $\mathbf{M}_r' \in \mathbb{R}^{C_O\times H\times W}$. In Classifier Guidance DM [25], an Adaptive Group Normalization (AdaGN) is proposed to boost the fusion ability of multi-source data. Inspired by the work, an Adaptive Instance Normalization (AdaIN) is designed and is integrated into each Res-Block to fuse $\mathbf{E}'$ and $\mathbf{M}_r'$. Following ResNet [80], a skip connection is introduced to merge intermediate data and the input feature in order to solve the degradation problem [80] in deep networks. Ultimately, the output feature map of a Res-Block is obtained, denoted as $\mathbf{O}_r \in \mathbb{R}^{C_O\times H\times W}$.

**Attention Block (or AB).** The inherent locality of convolutional operations enables convolutional neural networks (CNNs) to be good at modeling local information but concurrently poor at capturing global dependencies [81]. The self-attention mechanism, however, can bring intrinsic global perception without the complex hierarchical stacking structures like those in CNNs, and it has been successfully applied to computer vision [82][83][84]. Further, researchers have discovered that CNN+self-attention hybrid models can benefit from both the local perception capabilities of CNN and the capability of modeling long-range context features of self-attention. This insight has led to the development of many influential models, such as TransUNet [85] for medical image segmentation and BGCrack [81] for crack identification. Inspired by these applications, AB is designed based on self-attention and is embedded into the initial denoising network composed of CNNs. As for the embedded position, as illustrated in Fig. 6, an AB is inserted into the high-level features, guided by two considerations. On the one hand, high-level features possess larger receptive fields, which bring more redundant long-term context information [81][86]. Therefore, it is more effective to embed AB within high-level features. On the other hand, since high-level features have a smaller spatial resolution, such an arrangement can also help reduce computation overhead [81]. The structure of AB is delineated next.

AB consists of two sub-modules: a self-attention block (SAB) and a parallel cross-attention block (PCAB) appended to SAB. First, SAB performs self-attention on the output of the previous RB, denoted as $\mathbf{M}_s \in \mathbb{R}^{C\times H\times W}$, to model the internal long-range relationship within the feature map:

$$\mathbf{O}_s = \mathrm{SAB}(\mathbf{M}_s), \tag{27}$$

where $\mathbf{O}_s$ represents the output feature map. Subsequently, PCAB takes $\mathbf{O}_s$ along with the condition embeddings, $\mathbf{E}_t$ and $\mathbf{E}_d$, as the inputs. It calculates the cross self-attention between $\mathbf{O}_s$ and $\mathbf{E}_t$, and between $\mathbf{O}_s$ and $\mathbf{E}_d$ using two cross-attention blocks (CABs), respectively, to learn the latent global dependencies between the feature map and conditions:

$$\mathbf{O}_C^t = \mathrm{CAB}(\mathbf{O}_s, \mathbf{E}_t), \tag{28a}$$

$$\mathbf{O}_C^d = \mathrm{CAB}(\mathbf{O}_s, \mathbf{E}_d), \tag{28b}$$

where $\mathbf{O}_C^t$ and $\mathbf{O}_C^d$ are the outputs. Further, a parallel fusion strategy, borrowing from the successful experience of PAM [5] and BGCrack, is conducted to fuse multiple features:

$$\mathbf{M}_{Fuse} = w_1 \times \mathbf{O}_s + w_2 \times \mathbf{O}_C^t + w_3 \times \mathbf{O}_C^d, \tag{29}$$

where $\mathbf{M}_{Fuse}$ denotes the fused feature map, and $w_1$, $w_2$, and $w_3$ are 3 learnable parameters to adjust the weights. Subsequently, a straightforward convolution block with the structure of "GN→SiLU→3 × 3 *Conv*" is applied to $\mathbf{M}_{Fuse}$ to obtain the output of PCAB, $\mathbf{O}_p \in \mathbb{R}^{C\times H\times W}$. The calculation process for AB is expounded in Appendix C.

# 6. Details of implementation

## 6.1. Dataset

The dataset utilized in this paper, referred to as Modified-dataset, is derived from Original-dataset in StructGAN. The process of establishing and detailing the training set of Modified-dataset is described as follows:

1. Image collection: All images in the training set of Modified-dataset are sourced from Original-dataset, specifically 63 images from the Group7-H1 sub-dataset, 55 images from the Group7-H2 sub-dataset, and 57 images from the Group8 sub-dataset.
2. Resolution adjustments: In line with StructGAN, the spatial resolution of the images in Modified-dataset is adjusted to 512×1024.
3. Color correction: It was noted that the colors of some images in Original-dataset did not accurately reflect the descriptions in the text (refer to Section 4 or the StructGAN paper). To address this, the RGB values of all pixels in all images were standardized to ensure consistency with the RGB tuples.
4. Data augmentation: Modified-dataset employs the same data augmentation strategy as Original-dataset. Each image is flipped vertically and horizontally and rotated 180 degrees. Post-augmentation, the image counts are 252 for Group7-H1, 220 for Group7-H2, and 228 for Group8, respectively.
5. Image combination: As discussed in Subsection 5.5.1, PCDM is trained on a mixed dataset comprising all three sub-datasets. To facilitate the integration of physical conditions into the image data and to enable PCDM to easily distinguish images from different sub-datasets, three identifiers are added to the names of the images, such as “7degree” for images in Group7-H1 and “8degree” for images in Group8. So far, the three sub-datasets of Modified-dataset have been obtained. Subsequently, the three sub-datasets are merged to create a new mixed dataset, which forms the final training set of Modified-dataset, now containing 700 images. Modified-dataset is referred to the mixed dataset, unless otherwise stated.

The images in the test set of Modified-dataset also come from Original-dataset, and the establishment of the test set follows the four steps of the above process except for the data augmentation step: image collection, resolution adjustments, color correction, and image combination.

Modified-dataset is open-sourced and publicly available at https://github.com/hzlbbfrog/Generative-BIM.

## 6.2. Training and inference environment

The hardware used for training and testing is: an AMD Ryzen 7 5800X3D CPU, a Nvidia GeForce RTX 4090 GPU (24 GB), and 128 GB of RAM. The main software environment is: Windows 10 operating system, CUDA 11.6.2, CUDNN 8.6.0, and Python 3.8.15. The deep learning framework employs version 1.12.1 of PyTorch, developed by Meta.

## 6.3. Training policy

The Adam optimizer is utilized for training the networks, and the hyper-parameters are set to the default values $\beta = (0.9, 0.999)$, $\varepsilon = 10^{-8}$ and $weight\ decay = 0$. The batch size is set to 1, and the learning rate is fixed at $1 \times 10^{-4}$. The networks are all trained for 70 epochs before inference. For simplicity, no additional data augmentation and learning rate schedulers are adopted in the training stage. It is worth noting that all the models are implemented in this unified training pipeline to ensure fairness in comparisons, unless specified otherwise.

## 6.4. Evaluation metrics

To validate and evaluate the performance of the proposed models, two types of metrics are employed: an objective evaluation metric and a subjective evaluation metric. StructGAN and its subsequent work only utilize the objective metric, namely, $Score_{\mathrm{IoU}}$, which is calculated by comparing predictions with labels pixel by pixel, while neglecting subjective metrics. In fact, subjective metrics such as Inception Score (IS) [87] and Fréchet Inception Distance (FID) [88][89] are even more crucial as they can quantitatively evaluate the visual quality of generated samples by simulating human perception. Their values match consistently well with human judgment [90] and indicate whether the generation samples "***look***" realistic or not. Therefore, subjective metrics are introduced to determine if the generated drawings ***look like*** the structural design drawings designed by engineers. Combining both metrics can lead to a more accurate and thorough evaluation of the models' performance.

#### *6.4.1. Objective metric*

Following StructGAN and its variants, $Score_{\mathrm{IoU}}$ is adopted to evaluate the generation results, which can be defined by the following formula:

$$Score_{\mathrm{IoU}} = \eta_{\mathrm{SWratio}} \times (\eta_{\mathrm{SIoU}} \cdot \mathrm{SIoU} + \eta_{\mathrm{WIoU}} \cdot \mathrm{WIoU}). \tag{30}$$

In the above equation,

$$\eta_{\mathrm{SWratio}} = \left(1 - \frac{\left|\mathrm{SWratio}_{pre} - \mathrm{SWratio}_{label}\right|}{\mathrm{SWratio}_{pre}}\right) \times 100\%, \tag{31}$$

where $\mathrm{SWratio}_{pre}$ and $\mathrm{SWratio}_{label}$ represents the ratios of shear walls in a prediction and in a label, respectively, and they can be calculated by the following equations:

$$\mathrm{SWratio}_{pre} = \frac{A_{pre}^{\mathrm{SW}}}{A_{pre}^{\mathrm{W}}}, \tag{32a}$$

$$\mathrm{SWratio}_{label} = \frac{A_{label}^{\mathrm{SW}}}{A_{label}^{\mathrm{W}}} \tag{32b}$$

where $A_{pre}^{\mathrm{SW}}$ and $A_{label}^{\mathrm{SW}}$ denote the areas of shear walls in a predicted design drawing and in a label, respectively. Similarly, $A_{pre}^{\mathrm{W}}$ and $A_{label}^{\mathrm{W}}$ represent the total areas of walls, comprising both the area of shear walls and the area of infill walls, in a generated image and in a label, respectively.

Additionally, in Eq. (30), $\eta_{\mathrm{SIoU}}$ and $\eta_{\mathrm{WIoU}}$ represent the weight coefficients of SIoU and WIoU, with both coefficients set to $0.5$. SIoU refers to the intersection over union (IoU) specific to shear walls, while WIoU represents the weighted IoU. Furthermore, $\eta_{\mathrm{SW_{ratio}}}$ denotes the correction coefficient for the overall quantity of shear walls. The three variables are defined as follows:

$$\mathrm{SIoU} = \frac{A_{\mathrm{SWinter}}}{A_{\mathrm{SWunion}}}, \tag{33}$$

where $A_{\mathrm{SWinter}}$ is the intersection area of the shear walls between predictions and labels, and $A_{\mathrm{SWunion}}$ is the corresponding union area.

$$\mathrm{WIoU} = \sum_{i=0}^{4} \frac{w_i p_{ii}}{\sum_{j=0}^{4} p_{ij} + \sum_{j=0}^{4} p_{ji} - p_{ii}}, \tag{34}$$

where $i \in \{0,1,\dots,4\}$ denotes the category number, and $w_i$ represents the weight parameter of the class. Specifically, $i = 0$ denotes black background, $i = 1$ denotes shear walls, $i = 2$ denotes infill walls, $i = 3$ denotes windows, and $i = 4$ denotes outdoor gates. The corresponding weights are $\{w_0 = 0, w_1 = 0.4, w_2 = 0.4, w_3 = 0.1, w_4 = 0.1\}$. $p_{ij}$ represents the number of pixels that actually belong to category $i$ but are predicted to category $j$.

Based on the above definitions, we can see that $Score_{\mathrm{IoU}}$ is an objective metric that evaluates generation results through pixel-level comparison. It is clear that the larger the values of $\eta_{\mathrm{SWratio}}$, SIoU, and WIoU, namely, the larger the value of $Score_{\mathrm{IoU}}$, the closer the prediction aligns with the label.

#### *6.4.2. Subjective metric*

IS and FID are the two most commonly used metrics to evaluate the perceptual quality of the generated images. Considering FID is more robust to noise than IS [90], FID is adopted in this study.

Specifically, FID first embeds two image sets (*i.e.*, real images $\mathbf{r}$ and generated images $\mathbf{g}$) into high-level feature space using a pre-trained Inception-V3 model [91]. Subsequently, FID treats the feature space as a continuous multivariate Gaussian distribution and estimates the mean vector and covariance matrix of the representations in the feature space for both the real data and the generated data, respectively, denoted as $\mathcal{N}(\boldsymbol{\mu}_{\mathbf{r}}, \boldsymbol{\Sigma}_{\mathbf{r}})$ and $\mathcal{N}(\boldsymbol{\mu}_{\mathbf{g}}, \boldsymbol{\Sigma}_{\mathbf{g}})$. Finally, the value of FID is the Fréchet distance between these two Gaussian distributions, which provides a quantitative measure of generation quality:

$$\mathrm{FID}(\mathbf{r}, \mathbf{g}) = \|\boldsymbol{\mu}_{\mathbf{r}} - \boldsymbol{\mu}_{\mathbf{g}}\|_2^2 + \mathrm{Tr}\left(\boldsymbol{\Sigma}_{\mathbf{r}} + \boldsymbol{\Sigma}_{\mathbf{g}} - 2\sqrt{\boldsymbol{\Sigma}_{\mathbf{r}}\boldsymbol{\Sigma}_{\mathbf{g}}}\right), \tag{35}$$

where $\mathrm{Tr}$ denotes the trace of matrix.

In practice, this study utilizes pytorch-fid[‡], the official PyTorch implementation of FID, to compute the FIDs of the generated results. It is important to note that FID is inversely correlated with the perceptual quality of the generated samples; thus, a lower FID value indicates better model performance.

## 7. Experiments

Extensive experiments have been conducted to validate the effectiveness of the proposed method. Specifically, Subsections 7.1 and 7.2 provide detailed quantitative and qualitative comparisons with the recent SOTA models (StructGAN and its variants), respectively. The results of ablation studies are presented in Subsection 7.3. Subsection 7.4 analyzes the powerful generation and generalization capabilities of PCDM. Subsection 7.5 discusses the stability of the proposed generative AI approach.

### *7.1. Quantitative evaluation*

Table 1 quantitatively compares the results obtained from PCDM, StructGAN, and StructGAN-AE. It is important to note that StructGAN-SA-PT and StructGAN-AA-PT in Table 1 are the top 2 methods proposed in StructGAN-AE. Except for FID, the results of StructGAN and its variants are taken from the paper of StructGAN-AE [3]. Based on the open-sourced code of StructGAN, the StructGAN model is re-trained on the three sub-datasets of Modified-dataset, respectively, to obtain FID with all the hyperparameters set to the default values in the original code. Since StructGAN-AE does not open-source its code, there is no FID of StructGAN-AE in Table 1. PCDM is tested on the three sub-datasets, respectively, to obtain the values of the metrics for different sub-datasets. The best $Score_{\mathrm{IoU}}$ and FID for each sub-dataset have been highlighted in red and bold in Table 1, and it is evident that PCDM significantly outperforms all the SOTA competitors across both the objective metric ($Score_{\mathrm{IoU}}$) and the subjective metric (FID). For instance, in Group7-H2, $Score_{\mathrm{IoU}}$ of PCDM $(= 0.74)$ is $0.17$ (or $\approx 30\%$) higher than the best score of the GAN-based methods $(= 0.57)$, and in Group8, FID of PCDM $(= 11.39)$ is lower than one-third of FID of StructGAN $(= \frac{43.47}{3} = 14.49)$. The quantitative results demonstrate that on the one hand, the generations of PCDM are closer to the manual labels at the pixel level (based on the objective metric), and on the other hand, the superiority of PCDM on the subjective metric showcases the generated designs of PCDM look more realistic and more like the engineers' designs, which is also validated by the qualitative comparison in Subsection 7.2.

---

[‡] https://github.com/mseitzer/pytorch-fid

**Table 1** Quantitative evaluation of the results obtained from different models.

| Test set | Metrics | StructGAN [1] | StructGAN -SA-PT [3] | StructGAN -AA-PT [3] | **PCDM** (Our work) |
|---|---|---|---|---|---|
| **Group7-H1** | SIoU ↑ | 0.41 | 0.58 | 0.60 | 0.68 |
| | WIoU ↑ | 0.59 | 0.64 | 0.67 | 0.64 |
| $\text{SWratio}_{label} = 0.41$ | $\text{SWratio}_{pre}$ ← | 0.50 | 0.53 | 0.48 | 0.42 |
| | $\eta_{\text{SWratio}}$ ↑ | 80% | 77% | 81% | 98.18% |
| | $Score_{\text{IoU}}$ ↑ | 0.40 | 0.48 | 0.52 | **0.65** |
| | FID ↓ | 70.32 | / | / | **23.57** |
| **Group7-H2** | SIoU ↑ | 0.58 | 0.64 | 0.68 | 0.85 |
| | WIoU ↑ | 0.63 | 0.63 | 0.64 | 0.66 |
| $\text{SWratio}_{label} = 0.58$ | $\text{SWratio}_{pre}$ ← | 0.66 | 0.63 | 0.72 | 0.56 |
| | $\eta_{\text{SWratio}}$ ↑ | 86% | 90% | 81% | 97.62% |
| | $Score_{\text{IoU}}$ ↑ | 0.52 | 0.57 | 0.54 | **0.74** |
| | FID ↓ | 46.14 | / | / | **21.19** |
| **Group8** | SIoU ↑ | 0.74 | 0.80 | 0.78 | 1.10 |
| | WIoU ↑ | 0.72 | 0.74 | 0.71 | 0.70 |
| $\text{SWratio}_{label} = 0.66$ | $\text{SWratio}_{pre}$ ← | 0.76 | 0.72 | 0.73 | 0.74 |
| | $\eta_{\text{SWratio}}$ ↑ | 87% | 91% | 90% | 90.24% |
| | $Score_{\text{IoU}}$ ↑ | 0.65 | 0.70 | 0.67 | **0.81** |
| | FID ↓ | 43.47 | / | / | **11.39** |

Notes: ↑ means for some metric, the higher the value, the better. ↓ means for some metric, the lower the value, the better. ← means for some metric, the closer the value is to the true value, the better.

### *7.2. Qualitative comparison with StructGAN*

This section presents the qualitative comparison results between PCDM and StructGAN, which are depicted in Fig. 7 below. It is important to note that since StructGAN does not open-source the trained models, we retrain StructGAN based on the open-sourced code[§] without changing any configuration.

Based on the visualization results, clearly, on the one hand, the generated results of PCDM are clearer and have higher perceptual quality than those of StructGAN. On the other hand, the designs of PCDM meet have more accurate details meeting essential design criteria, but those of StructGAN do not, which is reflected in three aspects: (1) the wall boundaries are all horizontal and vertical in the designs of PCDM but the wall boundaries in those of StructGAN present various angles, such as the inset a in Group7-H1, the inset b and the inset c in Group7-H2, and the inset b in Group8; (2) the designed shear walls are as continuous as possible in the results of PCDM, but the designed walls of StructGAN are always intermittent, such as the inset b and the inset d in Group7-H2, and the inset b and the inset c in Group8; and (3) for elevator areas, PCDM's results meet the requirements for closure to form core tubes, but StructGAN fails to cover the crucial data mode, such as the inset c in Group7-H1, the inset c in Group7-H2, and the inset d in Group8. In summary, these two advantages make the generated designs of PCDM more realistic and closer to human engineers' designs, which is also validated by the quantitative evaluation in Subsection 7.1. Additionally, the drawbacks in the structural design drawings generated by StructGAN such as blurred images, unclear wall boundaries and intermittent walls, are impossible to exist in practical engineering projects. Therefore, the generations of StructGAN brings engineers much more post processing workload than those of PCDM.

[§] https://github.com/wenjie-liao/StructGAN_v1

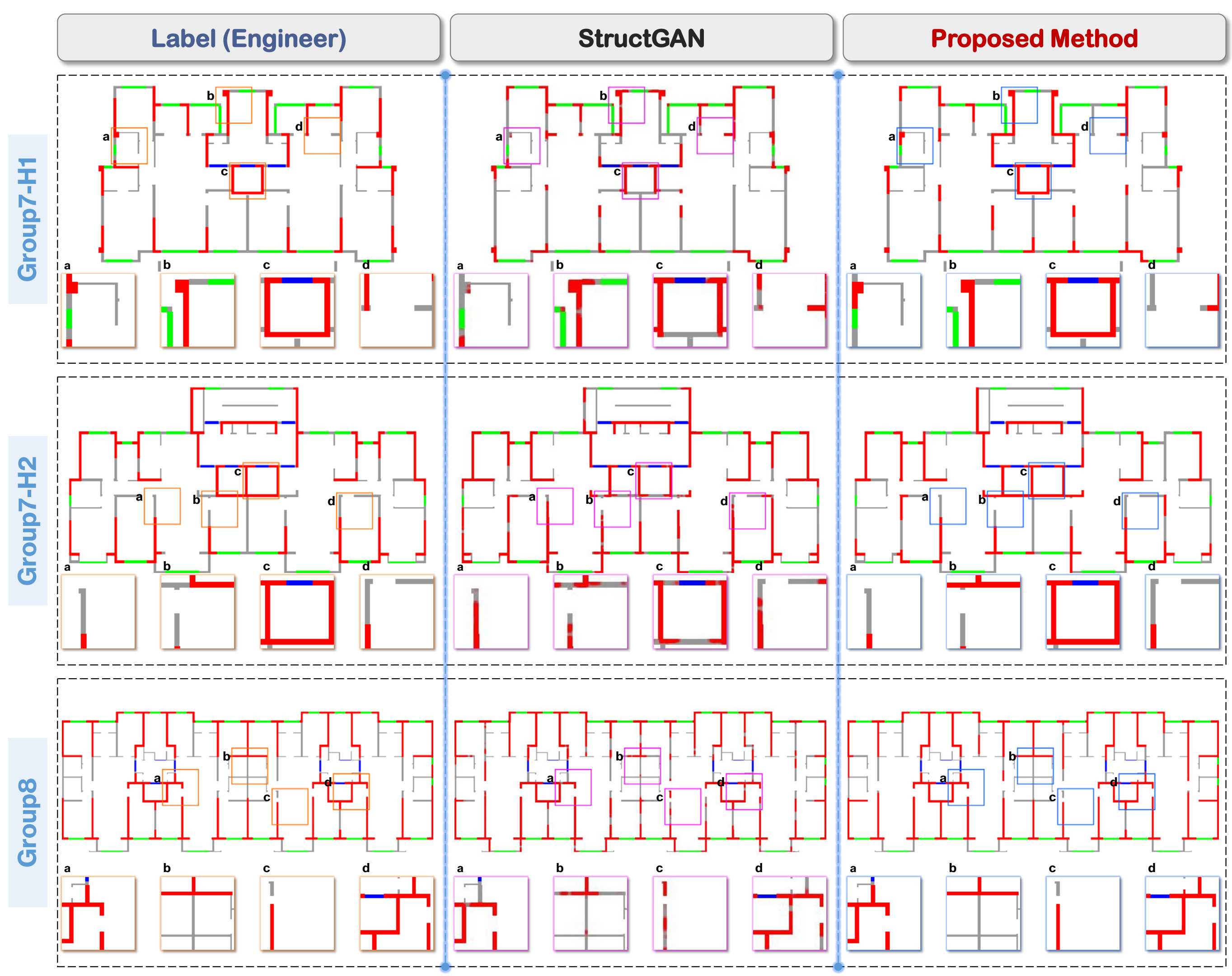

**Fig. 7.** Comparison of visualization results of labelling and the ones obtained from StructGAN [1] and our proposed PCDM.

### *7.3. Ablation study*

Ablation experiments are conducted to examine the validity of the designs, as delineated below.

#### *7.3.1. Predicting clean line drawing or noise*

As mentioned in Subsection 5.4, this study predicts the line drawing $\mathbf{x}_0$ by first estimating the shear wall drawing $\mathbf{s}_0$ rather than noise $\boldsymbol{\epsilon}_0^t$ for focusing on perceptual similarity. Subsection 5.4 analyzes the equivalence between predicting $\mathbf{x}_0$ and $\boldsymbol{\epsilon}_0^t$ from a theoretical perspective. This subsection demonstrates the analysis from an experimental standpoint. As shown in Fig. 8, the generation results are similar to the labels when the model predicts $\mathbf{x}_0$. For example, the background, shear walls, and infill walls in the two pairs of images have the same colors. However, when the model predicts $\boldsymbol{\epsilon}_0^t$, there is a significant difference between the results and the labels, and the difference is variable and has no discernible pattern. Therefore, the experimental results provide solid evidence to confirm the theoretical analysis in Subsection 5.4 that estimating $\mathbf{x}_0$ focusing on similarity, and more attention is paid to variability when estimating $\boldsymbol{\epsilon}_0^t$.

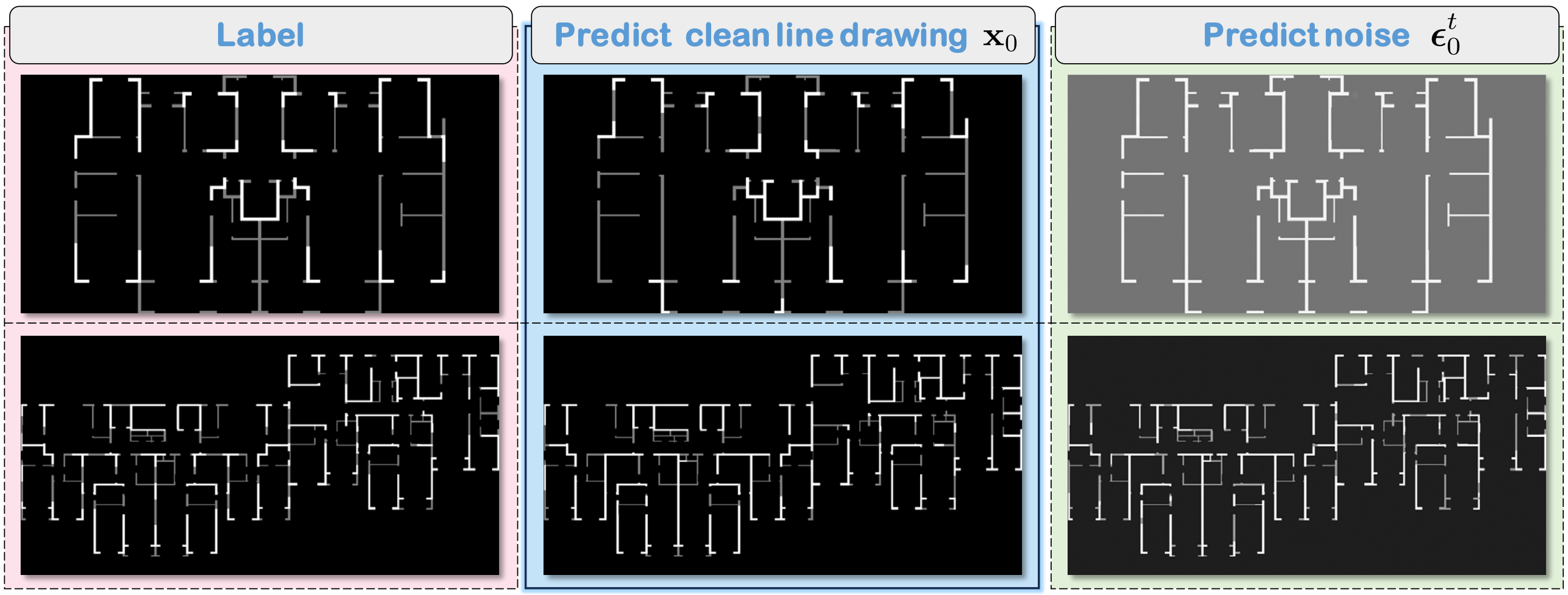

**Fig. 8.** Results generated by different prediction choices.

### *7.4. Analysis of generation and generalization abilities of PCDM*

In this subsection, we analyze the generation and generalization abilities of PCDM from two perspectives. The first aspect is to investigate its diverse design ability, as demonstrated in Subsection 7.4.1, and in Subsection 7.4.2, we explore whether PCDM has the capability for creative structural design. It is important to note that the powerful design abilities do not exist in the current SOTA techniques and the conventional DMs.

#### *7.4.1. Diverse design: Generation of different design drawings without changing the model and input information*

In this subsection, we first select one image from each of the three groups and then repeatedly execute the sampling process to check if the model can generate various designs while ensuring the model and conditions remain unchanged for the same input image. Some generated designs are presented in Fig. 9 below. It is clear that the generations are similar but not the same each time. For example, the inset a in Group7-H1 and the inset c in Group8 show the differences between generated results. Additionally, $\mathrm{SWratios}_{pre}$ of sampling five times for the selected images from the three groups are listed in Table 2. We can find that the shear wall ratios are similar for the generations of the same input drawing. These illustrate that PCDM can generate different but similar designs while keeping the input drawings and conditions unchanged.

Further, as demonstrated in the inset c in Group7-H1, the inset c in Group7-H2, and the inset a and the inset d in Group8, we can see that the different designs all have high perceptual quality and satisfy the fundamental design criteria. For example, the images are clean enough, the wall boundaries are horizontal and vertical, and the shear walls are incorporated within elevator areas to form core tubes.

These findings demonstrate that PCDM possesses powerful generation ability, allowing for diverse structural design by controlling input information unchanged.

**Table 2** $\mathbf{SWratio}_{pre}$ of the generated results by multiple samplings.

| Image groups and physical condition | First | Second | Third | Fourth | Fifth |
|---|---|---|---|---|---|
| Group7-H1 (1.0) | 0.46 | 0.45 | 0.43 | 0.46 | 0.45 |
| Group7-H2 (1.5) | 0.61 | 0.58 | 0.62 | 0.59 | 0.59 |
| Group8 (2.5) | 0.73 | 0.74 | 0.73 | 0.72 | 0.72 |

#### *7.4.2. Creative design: Generation of different results to meet the requirement of different physical conditions*

In this experiment, we aim to explore for the same input drawing, whether the generated designs can be correspondingly altered when we control the input physical conditions change, while still meeting the modified physical conditions.

The quantitative evaluation results are presented in Table 3. We can find that as the input physical conditions increase, the $\mathrm{SWratio}_{pre}$ increases accordingly. Furthermore, the $\mathrm{SWratio}_{pre}$ is consistently close to the corresponding $\mathrm{SWratio}_{label}$. For example, for the images in Group8 (with a default physical condition of 2.5), if the physical condition is changed to $1.5$ (1.5 is the condition of Group7-H2), the $\mathrm{SWratio}_{pre}$ changes to 0.57, which is close to 0.58 (the $\mathrm{SWratio}_{label}$ of Group7-H2). Similarly, if the physical condition is changed to the condition of Group7-H1, which is $1.0$, the new $\mathrm{SWratio}_{pre}$ is 0.40 and is close to 0.41 (the $\mathrm{SWratio}_{label}$ of Group7-H1). Furthermore, as depicted in Fig. 10, we take one architectural design drawing in Group7-H2 and one architectural layout in Group8 as examples to investigate the generation results from a qualitative standpoint. It is clear that PCDM generates more shear walls as the physical condition increases, which is consistent with actual situations. Additionally, the generated results still have high visual quality and align with human design criteria, such as the boundaries of the shear walls are all horizontal and vertical, and the shear walls are all enclosed in the elevator areas to form core tubes, as exemplified by all the 6 subfigures labeled with the identifier "b". These quantitative results and visualization results collectively demonstrate that PCDM has accurately learned the implicit relationship between shear wall design and physical information, and thus can generate reasonable and high-quality designs that meet various physical conditions.

It is worth noting that during the training stage, we only train PCDM using the ground-truth physical conditions. For instance, the architectural drawings in Group7-H1 sub-dataset are solely trained with a condition of $1.0$. This means that the experiment conducted in this subsection is a new task for the trained model. Nevertheless, PCDM performs this task extremely well, which demonstrates its powerful generalization and generation capabilities. We can utilize PCDM conduct creative structural design by controlling input information.

**Table 3** Quantitative evaluation of the results obtained from different models.

| Original set and physical condition | $\mathrm{SWratio}_{label}$ for the original set | $\mathrm{SWratio}_{pre}$ of different input physical conditions | | |
|---|---|---|---|---|
| | | 1.0 (Group7-H1) | 1.5 (Group7-H2) | 2.5 (Group8) |
| Group7-H1 (1.0) | 0.41 | **0.42** | 0.56 | 0.71 |
| Group7-H2 (1.5) | 0.58 | 0.41 | **0.56** | 0.73 |
| Group8 (2.5) | 0.66 | 0.40 | 0.57 | **0.74** |

Note: Bold values represent the input physical condition is consistent with the original dataset.

### 7.5. *Validation of training stability and sampling stability*

This subsection analyses the training stability and sampling stability of PCDM.

Specifically, in order to evaluate the training stability, we first train PCDM for 70 epochs (refer to Subsection 6.3). Then, we select the 10th, 20th, 30th, 40th, and 50th epochs and run once the reverse denoising process on the test set for the five models, respectively. Here, the reason for selecting five models is to evaluate the stability during the training process. Next, we calculate FIDs for the five generation results, respectively. So far, we have obtained a set of data points from once training. In order to make the statistical results universal, we subsequently repeat the above processes 15 times and draw a boxplot to represent the $15 \times 5$ data points, as shown in Fig. 11 (a). Based on the results, we can find that in the early stages of training (namely, the 10th epoch), there are a few outside points (a small circle represents an outlier), and as the training progresses, there are no more outliers. This means as the training progresses, there is a decreasing trend in data volatility. In a boxplot, the box part contains data from the first quartile to the third quartile (namely, the middle $50\%$ of the data). As depicted in Fig. 11 (a), overall, the five boxes are narrow, which shows the data distribution is concentrated. This demonstrates the training process of PCDM is stable. Further, we can infer the training of DMs is stable, which provides experimental evidence for the analysis about the stable training feature within DMs in Section 1.

To assess the sampling stability, we select the 10th, 20th, 30th, 40th, and 50th epochs during a training session and run 15 reverse denoising processes for each trained model. Then, we compute FIDs for the $15 \times 5$ sampling results to obtain $15 \times 5$ data points, which are represented on a violin plot, as shown in Fig. 11 (b). In a violin plot, the black thick bar contains data from the lower quartile to the upper quartile (namely, the middle 50% of the data), and the black fine bar represents the 95% confidence interval. Based on Fig. 11 (b), it is clear that for the 5 sets of data, the difference of FID for the middle 50% of the data is less than 2 and the difference of FID for the data in the 95% confidence interval is less than 4, which means for the same model, the volatility of the generation results is low, namely, the sampling process of PCDM is stable.

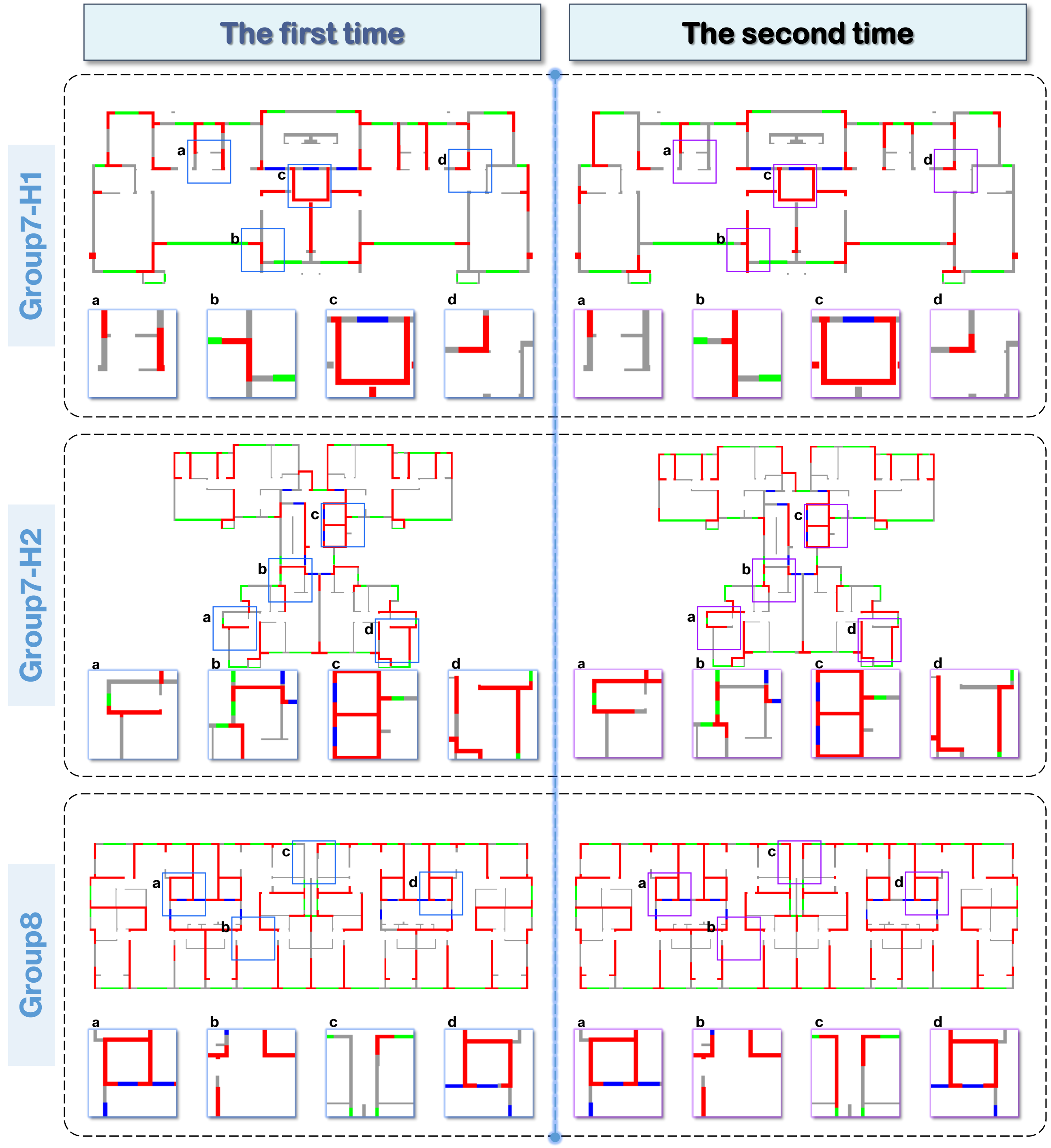

**Fig. 9.** Generation of different designs while the model and input information remain unchanged.

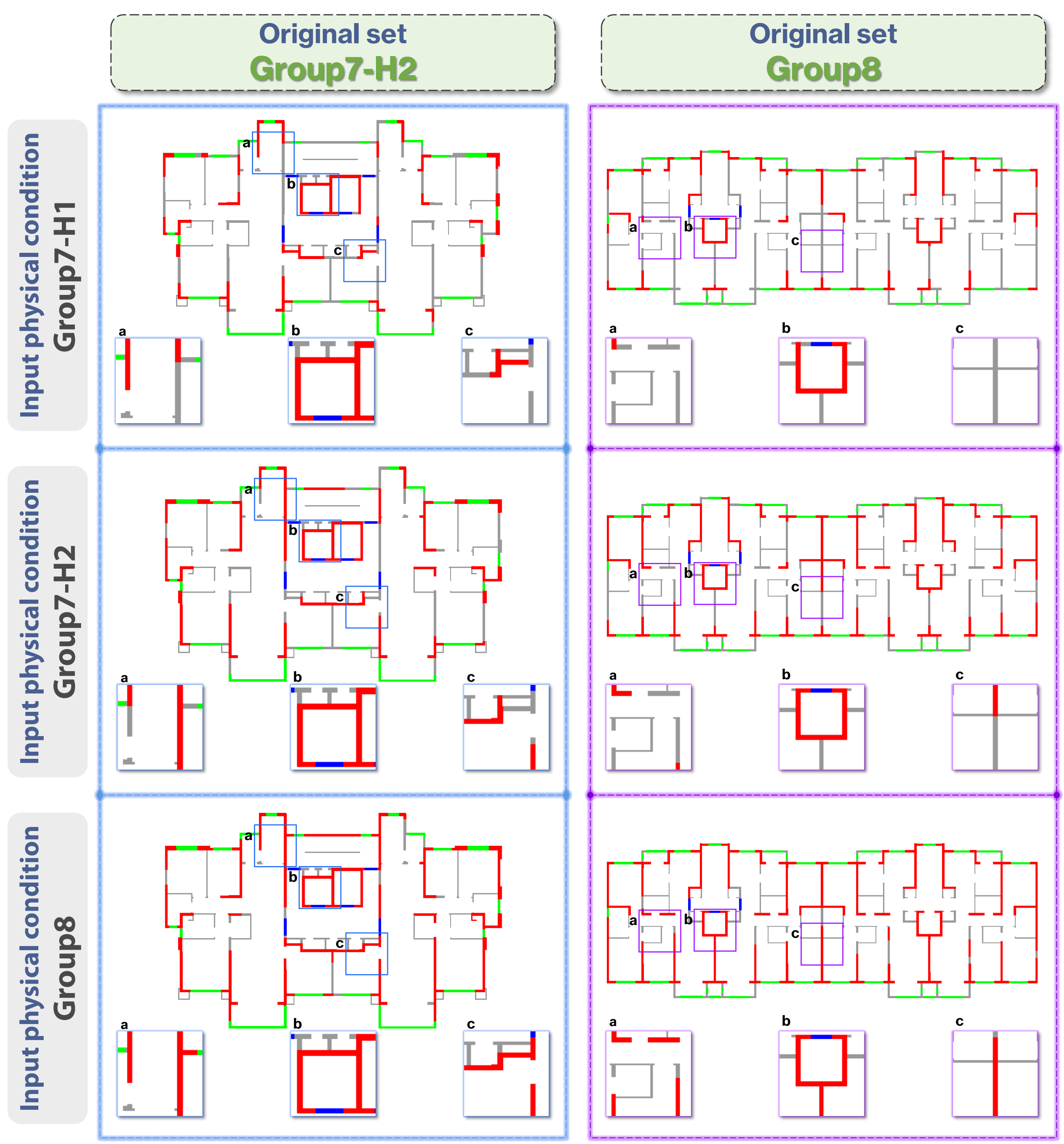

**Fig. 10.** Generation of different results that meet the requirements for different conditions.

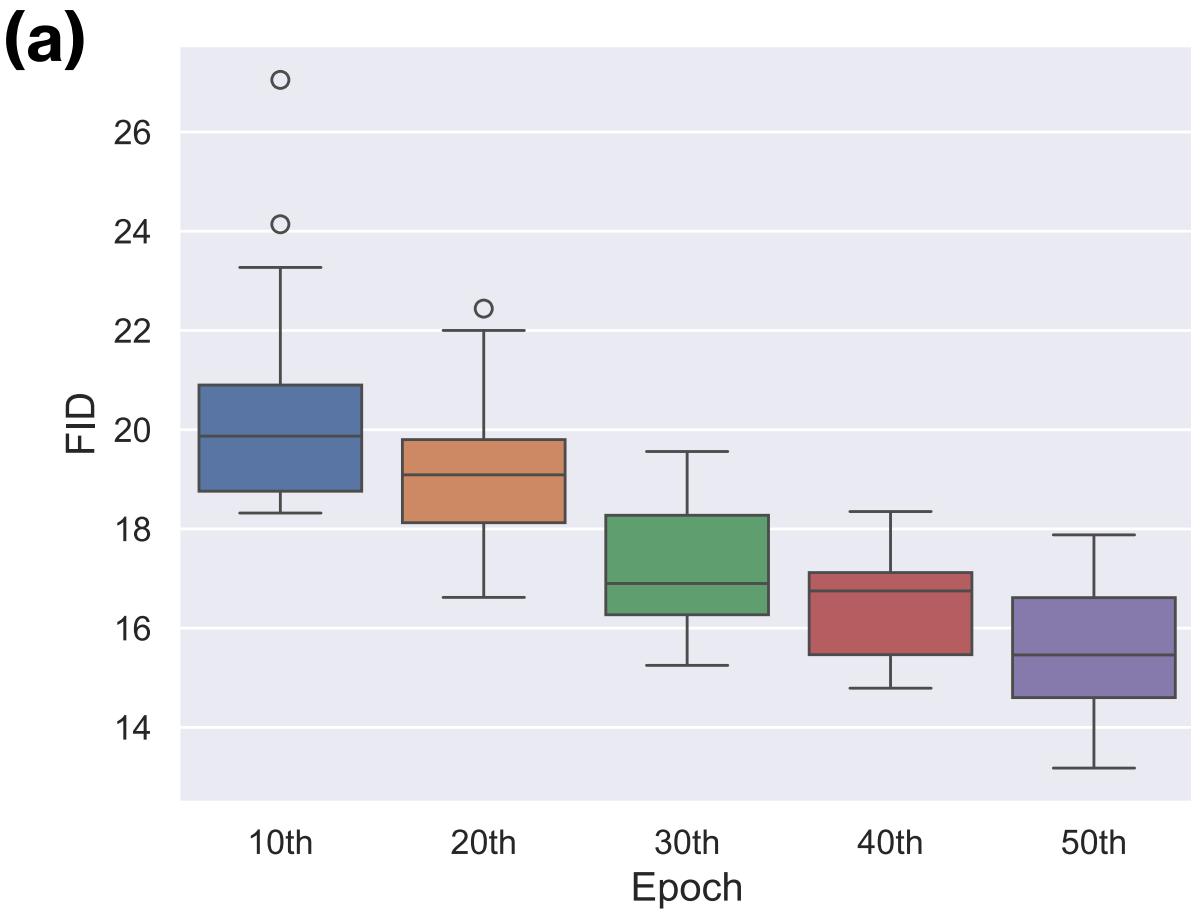

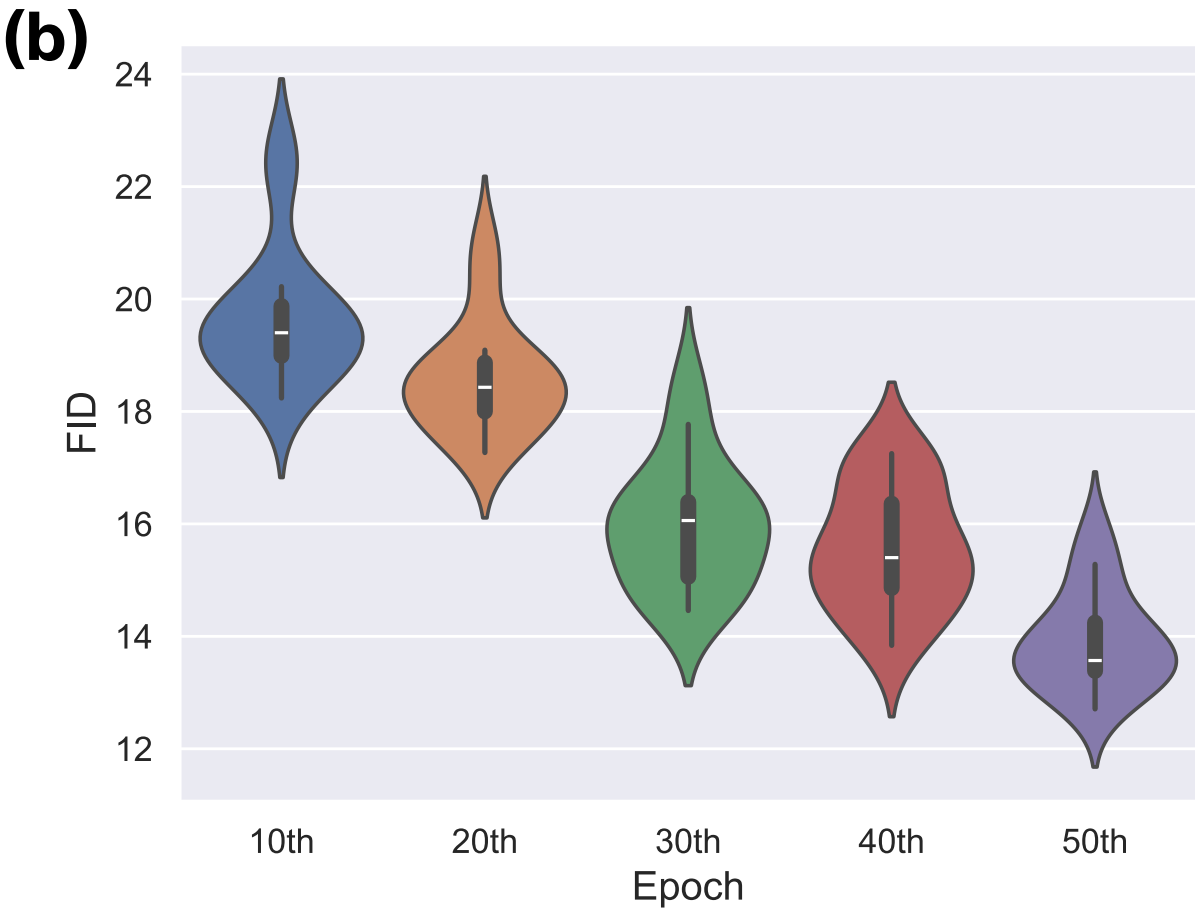

**Fig. 11.** Statistical results for validating stability: (a) evaluation results of 15 training sessions and (b) evaluation results of 15 sampling sessions. Note that in a boxplot, the black horizontal line inside a box represents the median, while the two black horizontal lines above and below the box indicate the upper adjacent value and the lower adjacent value, respectively.

## 8. Conclusions

This study introduces Generative AIBIM: an automated and intelligent structural design pipeline that integrates Building Information Modeling (BIM) with generative AI technologies. This pipeline is designed to tackle four key challenges currently facing AI-based intelligent structural design: expanding application scope, enhancing the generation framework for producing structural design drawings, utilizing advanced generative AI models for improved design quality (especially for shear walls) under various physical conditions, and employing more precise and comprehensive evaluation metrics for model performance assessment. The implementation codes and the dataset used in this study are publicly available at https://github.com/hzlbbfrog/Generative-BIM. The significant contributions and findings of this study are summarized below.

This integration of BIM and generative AI, on the one hand, broadens the application scope of BIM to shear wall intelligent design, which is in line with the ongoing developments in BIM technology and its applications within civil engineering. On the other hand, this pipeline conducts structural design based on BIM marking a significant advancement and supplement to the existing research frameworks that relied solely on CAD drawings. The core contribution of Generative AIBIM is the developed two-stage generation framework incorporating generative AI (TGAI), which is inspired by the human painting process and can significantly simplify the complexity of the structural design problem. TGAI consists of a design stage to generate shear walls and a coloring stage, and the pivotal component is the design stage, where this study proposes a novel generative AI model, the physics-based conditional diffusion model (PCDM). PCDM predicts shear wall drawings directly, as opposed to the noise predicted by traditional diffusion models (DMs). This method enables the model to concentrate on similarities, thus improving the realism of the generated structural design drawings. Furthermore, PCDM effectively fuses multi-source and multi-modal information, including design drawings (image data), timesteps, and physical conditions. The fusion enables PCDM to conduct diverse and creative structural designs accommodating different building heights and earthquake precautionary intensities, and such unique abilities do not exist in the conventional DMs and the GAN-based models adopted in previous intelligent structural design studies. In order to facilitate and enhance cross-domain data fusion, an attention block, including a self-attention block (SAB) and a parallel cross-attention block (PCAB), as well as an adaptive Instance Normalization (AdaIN) block, is designed and embedded in the neural network of PCDM. Last, considering StructGAN and its variants only utilize the objective metric, and in order to evaluate the performance of models more comprehensively, the Fréchet Inception Distance (FID), a widely used subjective metric, is further introduced to quantitatively measure the perceptual quality of the generated design drawings.

The comprehensive experimental results demonstrate the superior generation and representation capabilities of PCDM compared to the current SOTA models (namely, StructGAN and its variants). From the perspective of quantitative evaluation, PCDM significantly surpasses all the competitors across both the objective metric ($Score_{\mathrm{IoU}}$) and the subjective metric (FID). For example, in Group7-H2, $Score_{\mathrm{IoU}}$ of PCDM is $30\%$ higher than the best score of other models, and in Group8, FID of PCDM is lower than one-third of FID of StructGAN. From the qualitative comparison results, designs generated by PCDM are more realistic and more like the engineers' designs compared to those of StructGAN. The advancement is reflected in two aspects: (1) the generations of PCDM are clearer and have higher visual quality; and (2) the generated results of PCDM have more accurate details aligning with engineer design criteria. Associated ablation studies provide solid evidence of the effectiveness of our proposed approaches. Besides, extensive experiments are conducted, and the statistical results demonstrate that the training process and the sampling process of PCDM are both stable. This study introduces DMs to the civil engineering community, and the findings in this paper suggest that DMs have the huge potential to replace GANs and emerge as the new paradigm and benchmark in addressing the generative problems within the civil engineering domain. We also provide a thorough and intuitive exposition of DMs to facilitate comprehension by the civil engineering community and to catalyze further innovations and advancements in the field.

Looking ahead, there are still several promising directions for further research. First, the generation framework. Inspired by the indirect regression-based methods, Graph Neural Networks (GNNs) can be integrated in the current two-stage generation framework using PCDM to produce more refined design drawings, leveraging the synergistic potential of graph knowledge and generative AI. Second, the generative AI model. While this study provides a foundational exploration of DMs, there is ample scope for enhancing the performance of PCDM through further fine-tuning configurations, the exploration of more advanced DM-based approaches, or incorporating more appropriate physical rules. Finally, the generated objects. The application of the proposed Generative AIBIM is not limited to the design of shear wall structures but can be extended to other types of building structures, such as steel or reinforced concrete frame structures. This versatility underscores the broad applicability of our proposed pipeline in meeting diverse design needs across the civil engineering domain.

## Declaration of competing interest

The authors declare that they have no known competing financial interests or personal relationships that could have appeared to influence the work reported in this paper.

## Acknowledgements

The research presented was financially supported by the Innovation Technology Fund, Midstream Research Programme for Universities [project no. MRP/003/21X], and the Hong Kong Research Grants Council [project no. 16205021]. The first author would like to sincerely thank Miss Shuqi Wei for drawing the two subgraphs in Fig. 3 (b). The authors would like to thank Dr. Wenjie Liao (the first author of StuctGAN) of Tsinghua University for his valuable suggestions. The authors want to thank Dr. Cheng Ning Loong, Dr. Wai Yi Chau, Mr. Hao Xu, Mr. Tin Long Leung, and Mr. Jimmy Wu of HKUST for their valuable feedback. Finally, contributions by the anonymous reviewers are also highly appreciated.

## Appendix A Detailed derivation of formulas

### A.1 Derivation of Eq. (5)

Firstly, in Eq $(4)$, we let $\alpha_t = \frac{\bar{\alpha}_t}{\bar{\alpha}_{t-1}}$, then

$$\beta_t = 1 - \alpha_t. \tag{A1}$$

Plug $(\mathrm{A1})$ into Eq. $(1)$:

$$q(\mathbf{x}_t|\mathbf{x}_{t-1}) \triangleq \mathcal{N}(\mathbf{x}_t; \sqrt{\alpha_t}\mathbf{x}_{t-1}, (1-\alpha_t)\mathbf{I}_D). \tag{A2}$$

Using the reparameterization trick, $\mathbf{x}_t$ can be expressed as

$$\mathbf{x}_t = \sqrt{\alpha_t}\mathbf{x}_{t-1} + \sqrt{1-\alpha_t}\boldsymbol{\epsilon}_{t-1}^{t}. \tag{A3}$$

Let $t$ be $t-1$:

$$\mathbf{x}_{t-1} = \sqrt{\alpha_{t-1}}\mathbf{x}_{t-2} + \sqrt{1-\alpha_{t-1}}\boldsymbol{\epsilon}_{t-2}^{t-1}, \tag{A4}$$

where $\boldsymbol{\epsilon}_{t-1}^{t} \sim \mathcal{N}(0, \mathbf{I}_D)$ represents the added Gaussian noise from $\mathbf{x}_{t-1}$ to $\mathbf{x}_t$. Similarly, $\boldsymbol{\epsilon}_{t-2}^{t-1} \sim \mathcal{N}(0, \mathbf{I}_D)$. Plug $(\mathrm{A4})$ into $(\mathrm{A3})$:

$$\mathbf{x}_t = \sqrt{\alpha_t}\sqrt{\alpha_{t-1}}\mathbf{x}_{t-2} + \sqrt{\alpha_t}\sqrt{1-\alpha_{t-1}}\boldsymbol{\epsilon}_{t-2}^{t-1} + \sqrt{1-\alpha_t}\boldsymbol{\epsilon}_{t-1}^{t}. \tag{A5}$$

Based on the property of the Gaussian distribution, we know that the combination of two Gaussian distributions $\sqrt{\alpha_t}\sqrt{1-\alpha_{t-1}}\boldsymbol{\epsilon}_{t-2} + \sqrt{1-\alpha_t}\boldsymbol{\epsilon}_{t-1}$ is also a Gaussian distribution:

$$\sqrt{\alpha_t}\sqrt{1-\alpha_{t-1}}\boldsymbol{\epsilon}_{t-2}^{t-1} + \sqrt{1-\alpha_t}\boldsymbol{\epsilon}_{t-1}^{t} \sim \mathcal{N}(0, (1-\alpha_t\alpha_{t-1})\mathbf{I}_D). \tag{A6}$$

Then, apply the reparameterization trick again:

$$\sqrt{\alpha_t}\sqrt{1-\alpha_{t-1}}\boldsymbol{\epsilon}_{t-2}^{t-1} + \sqrt{1-\alpha_t}\boldsymbol{\epsilon}_{t-1}^{t} = \sqrt{1-\alpha_t\alpha_{t-1}}\boldsymbol{\epsilon}_{t-2}^{t}, \tag{A7}$$

where $\boldsymbol{\epsilon}_{t-2}^{t} \sim \mathcal{N}(0, \mathbf{I}_D)$. Combine $(\mathrm{A7})$ and $(\mathrm{A5})$:

$$\mathbf{x}_t = \sqrt{\alpha_t\alpha_{t-1}}\mathbf{x}_{t-2} + \sqrt{1-\alpha_t\alpha_{t-1}}\boldsymbol{\epsilon}_{t-2}^{t}. \tag{A8}$$

By analyzing $(\mathrm{A8})$ and $(\mathrm{A3})$, we can easily infer

$$\mathbf{x}_t = \sqrt{\alpha_t\alpha_{t-1}\cdots\alpha_1}\mathbf{x}_0 + \sqrt{1-\alpha_t\alpha_{t-1}\cdots\alpha_1}\boldsymbol{\epsilon}_0^{t} = \sqrt{\prod_{i=1}^{t}\alpha_i}\,\mathbf{x}_0 + \sqrt{1-\prod_{i=1}^{t}\alpha_i}\,\boldsymbol{\epsilon}_0^{t}, \tag{A9}$$

where $\boldsymbol{\epsilon}_0^{t} \sim \mathcal{N}(0, \mathbf{I}_D)$.

Based on the definition of $\bar{\alpha}_t$ (see Eq $(3)$), we can expand $t \in \{1, \dots, T\}$ to $t = 0$:

$$\bar{\alpha}_0 = \frac{g(0)}{g(0)} = 1. \tag{A10}$$

Because of $\alpha_t = \frac{\bar{\alpha}_t}{\bar{\alpha}_{t-1}}$, for $t \in \{1, \dots, T\}$, we can obtain

$$\bar{\alpha}_t = \prod_{i=1}^{t}\alpha_i\,. \tag{A11}$$

Embed Eq. $(\mathrm{A11})$ into Eq. $(\mathrm{A9})$ to simplify Eq. $(\mathrm{A9})$:

$$\mathbf{x}_t = \sqrt{\bar{\alpha}_t}\mathbf{x}_0 + \sqrt{1-\bar{\alpha}_t}\boldsymbol{\epsilon}_0^{t}, \tag{A12}$$

Then convert Eq. $(\mathrm{A12})$ into the form of a probability distribution:

$$q(\mathbf{x}_t|\mathbf{x}_0) = \mathcal{N}(\mathbf{x}_t; \sqrt{\bar{\alpha}_t}\mathbf{x}_0, (1-\bar{\alpha}_t)\mathbf{I}_D). \tag{A13}$$

### A.2 Derivation of Eqs. (7), (8), and (9)

Before introducing the derivation, a key theorem is proved first.

**Theorem 1.** For any probability density functions (pdfs), if they have the definition like

$$q(x) \propto \exp(-Ax^2 + Bx + C)\,, A > 0,$$

the pdfs are Gaussian distributions.

*Proof.* First of all, we can add a coefficient $D$ to replace the $\propto$:

$$q(x) = D \times \exp(-Ax^2 + Bx + C)\,, A > 0. \tag{A14}$$

Then, we transform the above formula through some math operations:

$$\begin{aligned} q(x) &= D \times \exp(-Ax^2 + Bx + C) \\ &= D \times \exp\left[-A\left(x^2 - \frac{B}{A}x - \frac{C}{A}\right)\right] \\ &= D \times \exp\left[-A\left(\left(x - \frac{B}{2A}\right)^2 - \frac{B^2}{4A^2} - \frac{C}{A}\right)\right] \\ &= D \times \exp\left[-A\left(x - \frac{B}{2A}\right)^2 + \frac{B^2 + 4AC}{4A}\right] \\ &= D \times \exp\left(\frac{B^2 + 4AC}{4A}\right)\exp\left[-\left(\sqrt{A}\left(x - \frac{B}{2A}\right)\right)^2\right]. \end{aligned} \tag{A15}$$

Further, according to the property of pdfs, we know that $\int_{\mathbb{R}} q(x)\,dx = 1$, so

$$\begin{aligned} \int_{\mathbb{R}} q(x)\,\mathrm{d}x &= \int_{\mathbb{R}} D \times \exp\left(\frac{B^2 + 4AC}{4A}\right)\exp\left[-\left(\sqrt{A}\left(x - \frac{B}{2A}\right)\right)^2\right]\mathrm{d}x \\ &= D \times \exp\left(\frac{B^2 + 4AC}{4A}\right)\int_{\mathbb{R}} \exp\left[-\left(\sqrt{A}\left(x - \frac{B}{2A}\right)\right)^2\right]\mathrm{d}x. \end{aligned} \tag{A16}$$

Then, we let $h = \sqrt{A}(x - \frac{B}{2A})$. Therefore, we have $x = \frac{\sqrt{A}}{A}h + \frac{B}{2A}$. In other words, $\mathrm{d}x = \frac{\sqrt{A}}{A}\mathrm{d}h$. Next, substitute $x$ in the above equation with $h$ to simplify the formula:

$$\int_{\mathbb{R}} q(x)\,\mathrm{d}x = D \times \exp\left(\frac{B^2 + 4AC}{4A}\right)\frac{\sqrt{A}}{A}\int_{\mathbb{R}} \exp(-h^2)\,\mathrm{d}h. \tag{A17}$$

Next, we need to calculate the integral $I = \int_{\mathbb{R}} \exp(-h^2)\,\mathrm{d}h$:

$$\begin{aligned} I \times I &= \left(\int_{\mathbb{R}} \exp(-x^2)\,\mathrm{d}x\right) \times \left(\int_{\mathbb{R}} \exp(-y^2)\,\mathrm{d}y\right) \\ &= \int_{\mathbb{R}}\int_{\mathbb{R}} \exp[-(x^2 + y^2)]\,\mathrm{d}x\,\mathrm{d}y \\ &= \int_0^{2\pi}\left(\int_0^{+\infty} r\exp[-(r^2)]\,\mathrm{d}r\right)\mathrm{d}\theta \\ &= \int_0^{2\pi}\left(\int_0^{+\infty} \frac{1}{2}\exp[-u]\,du\right)\mathrm{d}\theta \\ &= \int_0^{2\pi}\frac{1}{2}\mathrm{d}\theta \\ &= \pi. \end{aligned} \tag{A18}$$

Therefore, $I = \int_{\mathbb{R}} \exp(-h^2)\,\mathrm{d}h = \sqrt{\pi}$. Substitute $\int_{\mathbb{R}} \exp(-h^2)\,\mathrm{d}h$ in Eq. (A17) with $\sqrt{\pi}$:

$$\int_{\mathbb{R}} q(x)\,\mathrm{d}x = D \times \exp\left(\frac{B^2 + 4AC}{4A}\right)\frac{\sqrt{A}}{A}\sqrt{\pi} = 1. \tag{A19}$$

So,

$$D \times \exp\left(\frac{B^2 + 4AC}{4A}\right) = \frac{\sqrt{A}}{\sqrt{\pi}}. \tag{A20}$$

Combining Eqs. (A15) and (A20), we have

$$q(x) = \frac{\sqrt{A}}{\sqrt{\pi}} \exp\left[-A\left(x - \frac{B}{2A}\right)^2\right]. \tag{A21}$$

Then, we let

$$A = \frac{1}{2\sigma^2}, \sigma > 0 \tag{A22}$$

and

$$\frac{B}{2A} = \mu, \tag{A23}$$

Plug Eqs. (A22) and (A23) into Eq. (A21):

$$\begin{aligned} q(x) &= \frac{\sqrt{\frac{1}{2\sigma^2}}}{\sqrt{\pi}} \exp\left[-\frac{1}{2\sigma^2}(x-\mu)^2\right] \\ &= \frac{1}{\sqrt{2\pi}\sigma} \exp\left[-\frac{1}{2\sigma^2}(x-\mu)^2\right]. \end{aligned} \tag{A24}$$

Obviously, Eq. (A21) is just the expression of the Gaussian distribution, where $\mu$ is the mean, $\sigma^2$ is the variance, and $\sigma$ is named the standard deviation. Thus, we have proved Theorem 1. □

**Remark 1.** There are two points we should be aware of: (1) In Theorem 1, we only consider the situation in which $x$ is a scalar. However, it can be easily extended to the situation where $\mathbf{x}$ is a vector. Because the proof is similar to the proof of Theorem 1, for simplicity, we omit the proof but hold the conclusion. (2) Note that there are no $C$ and $D$ that are terms independent of $x$ in Eq. (A21), and the function of $C$ and $D$ is to make the integral of $q(x)$ equal to 1 (see Eqs. (A19) and (A20)), which means we can neglect the terms not involving $x$, and it does not affect the results. This is further clarified in the following derivation.

Then, let us start the derivation of Eqs. (7), (8), and (9). Based on Eq. (10), $q(\mathbf{x}_{t-1}|\mathbf{x}_t, \mathbf{x}_0)$ can be written as

$$\begin{aligned} q(\mathbf{x}_{t-1}|\mathbf{x}_t, \mathbf{x}_0) &= \frac{q(\mathbf{x}_t|\mathbf{x}_{t-1}, \mathbf{x}_0) q(\mathbf{x}_{t-1}|\mathbf{x}_0)}{q(\mathbf{x}_t|\mathbf{x}_0)} \\ &= \frac{q(\mathbf{x}_t|\mathbf{x}_{t-1}) q(\mathbf{x}_{t-1}|\mathbf{x}_0)}{q(\mathbf{x}_t|\mathbf{x}_0)}. \quad \text{(Markov Property)} \end{aligned} \tag{A25}$$

According to Eq. (1) and the property of the multivariate Gaussian distribution, we have

$$\begin{aligned} q(\mathbf{x}_t|\mathbf{x}_{t-1}) &= \frac{1}{(2\pi)^{D/2}|\beta_t \mathbf{I}_D|^{1/2}} \exp\left(-\frac{1}{2}\left(\mathbf{x}_t - \sqrt{1-\beta_t}\mathbf{x}_{t-1}\right)^\top (\beta_t \mathbf{I}_D)^{-1} \left(\mathbf{x}_t - \sqrt{1-\beta_t}\mathbf{x}_{t-1}\right)\right) \\ &\propto \exp\left(-\frac{1}{2}\frac{(\mathbf{x}_t - \sqrt{1-\beta_t}\mathbf{x}_{t-1})^\top (\mathbf{x}_t - \sqrt{1-\beta_t}\mathbf{x}_{t-1})}{\beta_t}\right). \end{aligned} \tag{A26}$$

Correspondingly, based on Eq. (5), we can get

$$q(\mathbf{x}_t|\mathbf{x}_0) \propto \exp\left(-\frac{1}{2}\frac{(\mathbf{x}_t - \sqrt{\bar{\alpha}_t}\mathbf{x}_0)^\top (\mathbf{x}_t - \sqrt{\bar{\alpha}_t}\mathbf{x}_0)}{1-\bar{\alpha}_t}\right), \tag{A27}$$

and

$$q(\mathbf{x}_{t-1}|\mathbf{x}_0) \propto \exp\left(-\frac{1}{2}\frac{(\mathbf{x}_{t-1} - \sqrt{\bar{\alpha}_{t-1}}\mathbf{x}_0)^\top (\mathbf{x}_{t-1} - \sqrt{\bar{\alpha}_{t-1}}\mathbf{x}_0)}{1-\bar{\alpha}_{t-1}}\right). \tag{A28}$$

Plug Eqs. (A26), (A27), and (A28) into Eq. (A25):

$$q(\mathbf{x}_{t-1}|\mathbf{x}_t,\mathbf{x}_0) \propto \exp\left[-\frac{1}{2}\left(\frac{(\mathbf{x}_t-\sqrt{1-\beta_t}\mathbf{x}_{t-1})^\top(\mathbf{x}_t-\sqrt{1-\beta_t}\mathbf{x}_{t-1})}{\beta_t} + \frac{(\mathbf{x}_{t-1}-\sqrt{\bar{\alpha}_{t-1}}\mathbf{x}_0)^\top(\mathbf{x}_{t-1}-\sqrt{\bar{\alpha}_{t-1}}\mathbf{x}_0)}{1-\bar{\alpha}_{t-1}} - \frac{(\mathbf{x}_t-\sqrt{\bar{\alpha}_t}\mathbf{x}_0)^\top(\mathbf{x}_t-\sqrt{\bar{\alpha}_t}\mathbf{x}_0)}{1-\bar{\alpha}_t}\right)\right], \tag{A29}$$

Because the random vector here is $\mathbf{X}_{t-1}$, and we want to get the distribution of $\mathbf{X}_{t-1}$, namely, $\mathbf{x}_{t-1}$ is the core object. Next, remove the innermost brackets related to $\mathbf{x}_{t-1}$:

$$q(\mathbf{x}_{t-1}|\mathbf{x}_t,\mathbf{x}_0) \propto \exp\left[-\frac{1}{2}\left(\frac{\mathbf{x}_t^\top\mathbf{x}_t-\mathbf{x}_{t-1}^\top(2\sqrt{1-\beta_t}\mathbf{x}_t)+(1-\beta_t)\mathbf{x}_{t-1}^\top\mathbf{x}_{t-1}}{\beta_t} + \frac{\mathbf{x}_{t-1}^\top\mathbf{x}_{t-1}-\mathbf{x}_{t-1}^\top(2\sqrt{\bar{\alpha}_{t-1}}\mathbf{x}_0)+\bar{\alpha}_{t-1}\mathbf{x}_0^\top\mathbf{x}_0}{1-\bar{\alpha}_{t-1}} - \frac{(\mathbf{x}_t-\sqrt{\bar{\alpha}_t}\mathbf{x}_0)^\top(\mathbf{x}_t-\sqrt{\bar{\alpha}_t}\mathbf{x}_0)}{1-\bar{\alpha}_t}\right)\right]. \tag{A30}$$

Then, collect like terms based on $\mathbf{x}_{t-1}$:

$$\begin{aligned} &q(\mathbf{x}_{t-1}|\mathbf{x}_t,\mathbf{x}_0) \\ &\propto \exp\left[-\frac{1}{2}\left(\left(\frac{1-\beta_t}{\beta_t}+\frac{1}{1-\bar{\alpha}_{t-1}}\right)\mathbf{x}_{t-1}^\top\mathbf{x}_{t-1}-\mathbf{x}_{t-1}^\top\left(\frac{2\sqrt{1-\beta_t}\mathbf{x}_t}{\beta_t}+\frac{2\sqrt{\bar{\alpha}_{t-1}}\mathbf{x}_0}{1-\bar{\alpha}_{t-1}}\right)+C_0(\mathbf{x}_t,\mathbf{x}_0)\right)\right] \\ &= \exp\left[-\frac{1}{2}\left(\mathbf{x}_{t-1}^\top\left(\frac{1-\beta_t}{\beta_t}+\frac{1}{1-\bar{\alpha}_{t-1}}\right)\mathbf{x}_{t-1}-\mathbf{x}_{t-1}^\top\left(\frac{2\sqrt{1-\beta_t}\mathbf{x}_t}{\beta_t}+\frac{2\sqrt{\bar{\alpha}_{t-1}}\mathbf{x}_0}{1-\bar{\alpha}_{t-1}}\right)\right)\right]. \end{aligned} \tag{A31}$$

where $C_0(\mathbf{x}_t,\mathbf{x}_0)$ does not involve $\mathbf{x}_{t-1}$, and its function is to let the integral of the pdf be 1. Recall Remark 1, it is clear that we can represent $q(\mathbf{x}_{t-1}|\mathbf{x}_t,\mathbf{x}_0)$ without knowing the specific expression of $C_0(\mathbf{x}_t,\mathbf{x}_0)$, so omit this term.

Because the exponent term of the pdf is a quadratic polynomial of $\mathbf{x}_{t-1}$, the unique distribution satisfying this is a Gaussian distribution based on Theorem 1. Thus, $q(\mathbf{x}_{t-1}|\mathbf{x}_t,\mathbf{x}_0)$ can be modeled as follows:

$$q(\mathbf{x}_{t-1}|\mathbf{x}_t,\mathbf{x}_0)=\mathcal{N}\left(\mathbf{x}_{t-1};\tilde{\boldsymbol{\mu}}_t(\mathbf{x}_t,\mathbf{x}_0),\widetilde{\boldsymbol{\Sigma}}_t(\mathbf{x}_t,\mathbf{x}_0)\right), \tag{A32}$$

where $\tilde{\boldsymbol{\mu}}_t(\mathbf{x}_t,\mathbf{x}_0)$ is the mean vector, and $\widetilde{\boldsymbol{\Sigma}}_t(\mathbf{x}_t,\mathbf{x}_0)$ is the covariance matrix. Based on the property of the multivariate Gaussian distribution, Eq. (A32) can be written as

$$\begin{aligned} q(\mathbf{x}_{t-1}|\mathbf{x}_t,\mathbf{x}_0) &\propto \exp\left(-\frac{1}{2}(\mathbf{x}_{t-1}-\tilde{\boldsymbol{\mu}}_t(\mathbf{x}_t,\mathbf{x}_0))^\top\left(\widetilde{\boldsymbol{\Sigma}}_t(\mathbf{x}_t,\mathbf{x}_0)\right)^{-1}(\mathbf{x}_{t-1}-\tilde{\boldsymbol{\mu}}_t(\mathbf{x}_t,\mathbf{x}_0))\right) \\ &= \exp\left(-\frac{1}{2}\left(\mathbf{x}_{t-1}^\top\widetilde{\boldsymbol{\Sigma}}_t^{-1}-\tilde{\boldsymbol{\mu}}_t^\top\widetilde{\boldsymbol{\Sigma}}_t^{-1}\right)(\mathbf{x}_{t-1}-\tilde{\boldsymbol{\mu}}_t)\right) \\ &= \exp\left(-\frac{1}{2}\left(\mathbf{x}_{t-1}^\top\widetilde{\boldsymbol{\Sigma}}_t^{-1}\mathbf{x}_{t-1}-\mathbf{x}_{t-1}^\top\widetilde{\boldsymbol{\Sigma}}_t^{-1}\tilde{\boldsymbol{\mu}}_t-\tilde{\boldsymbol{\mu}}_t^\top\widetilde{\boldsymbol{\Sigma}}_t^{-1}\mathbf{x}_{t-1}+\tilde{\boldsymbol{\mu}}_t^\top\widetilde{\boldsymbol{\Sigma}}_t^{-1}\tilde{\boldsymbol{\mu}}_t\right)\right). \end{aligned} \tag{A33}$$

Because $\tilde{\boldsymbol{\mu}}_t^\top\widetilde{\boldsymbol{\Sigma}}_t^{-1}\mathbf{x}_{t-1}$ is a scalar, $\tilde{\boldsymbol{\mu}}_t^\top\widetilde{\boldsymbol{\Sigma}}_t^{-1}\mathbf{x}_{t-1}=\left(\tilde{\boldsymbol{\mu}}_t^\top\widetilde{\boldsymbol{\Sigma}}_t^{-1}\mathbf{x}_{t-1}\right)^\top=\mathbf{x}_{t-1}^\top\left(\widetilde{\boldsymbol{\Sigma}}_t^{-1}\right)^\top\tilde{\boldsymbol{\mu}}_t$. Further, $\widetilde{\boldsymbol{\Sigma}}_t$ is symmetric, so $\widetilde{\boldsymbol{\Sigma}}_t^{-1}$ is symmetric. Then we have $\widetilde{\boldsymbol{\Sigma}}_t^{-1}=\left(\widetilde{\boldsymbol{\Sigma}}_t^{-1}\right)^\top$, namely, $\tilde{\boldsymbol{\mu}}_t^\top\widetilde{\boldsymbol{\Sigma}}_t^{-1}\mathbf{x}_{t-1}=\mathbf{x}_{t-1}^\top\widetilde{\boldsymbol{\Sigma}}_t^{-1}\tilde{\boldsymbol{\mu}}_t$. Plug this equation into Eq. (A33):

$$q(\mathbf{x}_{t-1}|\mathbf{x}_t,\mathbf{x}_0) \propto \exp\left(-\frac{1}{2}\left(\mathbf{x}_{t-1}^\top\widetilde{\boldsymbol{\Sigma}}_t^{-1}\mathbf{x}_{t-1}-2\mathbf{x}_{t-1}^\top\widetilde{\boldsymbol{\Sigma}}_t^{-1}\tilde{\boldsymbol{\mu}}_t+C_1(\mathbf{x}_t,\mathbf{x}_0)\right)\right)$$

$$\propto \exp\left(-\frac{1}{2}\left(\mathbf{x}_{t-1}^{\top}\widetilde{\mathbf{\Sigma}}_t^{-1}\mathbf{x}_{t-1}-\mathbf{x}_{t-1}^{\top}\left(2\widetilde{\mathbf{\Sigma}}_t^{-1}\tilde{\boldsymbol{\mu}}_t\right)\right)\right), \tag{A34}$$

where $C_1(\mathbf{x}_t,\mathbf{x}_0)$ denotes the terms independent of $\mathbf{x}_{t-1}$. Recalling Remark 1, we also omit $C_1(\mathbf{x}_t,\mathbf{x}_0)$ here just like in Eq. (A31). It is clear that $C_0(\mathbf{x}_t,\mathbf{x}_0)=C_1(\mathbf{x}_t,\mathbf{x}_0)$. Then, things become very easy. Combining Eqs. (A31) and (A34), we can establish 2 equations containing $\tilde{\boldsymbol{\mu}}_t$ and $\widetilde{\mathbf{\Sigma}}_t$ by the method of undetermined coefficients:

$$\begin{cases} \widetilde{\mathbf{\Sigma}}_t^{-1}=\dfrac{1-\beta_t}{\beta_t}+\dfrac{1}{1-\bar{\alpha}_{t-1}} \\ 2\widetilde{\mathbf{\Sigma}}_t^{-1}\tilde{\boldsymbol{\mu}}_t=\dfrac{2\sqrt{1-\beta_t}\mathbf{x}_t}{\beta_t}+\dfrac{2\sqrt{\bar{\alpha}_{t-1}}\mathbf{x}_0}{1-\bar{\alpha}_{t-1}} \end{cases}. \tag{A35}$$

Thus, $\tilde{\boldsymbol{\mu}}_t$ and $\widetilde{\mathbf{\Sigma}}_t$ can be derived from the above two equations.

Let us firstly derive $\widetilde{\mathbf{\Sigma}}_t$ by the first equation in Eq. (A35). Because $\widetilde{\mathbf{\Sigma}}_t^{-1}\in\mathbb{R}^{D\times D}$ and the right-hand side of the first equation is a scalar, the first equation can be transformed into a stricter form:

$$\widetilde{\mathbf{\Sigma}}_t^{-1}=\left(\frac{1-\beta_t}{\beta_t}+\frac{1}{1-\bar{\alpha}_{t-1}}\right)\mathbf{I}_D. \tag{A36}$$

Then, we can obtain $\widetilde{\mathbf{\Sigma}}_t$:

$$\begin{aligned}\widetilde{\mathbf{\Sigma}}_t&=1\Big/\left(\frac{1-\beta_t}{\beta_t}+\frac{1}{1-\bar{\alpha}_{t-1}}\right)\mathbf{I}_D\\ &=1\Big/\frac{(1-\beta_t)(1-\bar{\alpha}_{t-1})+\beta_t}{\beta_t(1-\bar{\alpha}_{t-1})}\mathbf{I}_D\\ &=\frac{\beta_t(1-\bar{\alpha}_{t-1})}{1-\bar{\alpha}_{t-1}-\beta_t+\beta_t\bar{\alpha}_{t-1}+\beta_t}\mathbf{I}_D\\ &=\frac{\beta_t(1-\bar{\alpha}_{t-1})}{1-\bar{\alpha}_{t-1}+\beta_t\bar{\alpha}_{t-1}}\mathbf{I}_D.\end{aligned} \tag{A37}$$

Recall Eq. (4):

$$\beta_t=1-\frac{\bar{\alpha}_t}{\bar{\alpha}_{t-1}}. \tag{A38}$$

namely,

$$\beta_t\bar{\alpha}_{t-1}=\bar{\alpha}_{t-1}-\bar{\alpha}_t. \tag{A39}$$

We can obtain $\widetilde{\mathbf{\Sigma}}_t$ after plugging Eq. (A39) into Eq. (A37):

$$\widetilde{\mathbf{\Sigma}}_t=\frac{1-\bar{\alpha}_{t-1}}{1-\bar{\alpha}_t}\beta_t\mathbf{I}_D. \tag{A40}$$

Further, we can think of $\frac{1-\bar{\alpha}_{t-1}}{1-\bar{\alpha}_t}\beta_t$ as a product of $\beta_t$, so we let $\tilde{\beta}_t=\frac{1-\bar{\alpha}_{t-1}}{1-\bar{\alpha}_t}\beta_t$. Thus, $\widetilde{\mathbf{\Sigma}}_t=\tilde{\beta}_t\mathbf{I}_D$.

Next, let us derive $\tilde{\boldsymbol{\mu}}_t$. Firstly, simplify the second equation of Eq. (A35):

$$\tilde{\boldsymbol{\mu}}_t=\widetilde{\mathbf{\Sigma}}_t\left(\frac{\sqrt{1-\beta_t}\mathbf{x}_t}{\beta_t}+\frac{\sqrt{\bar{\alpha}_{t-1}}\mathbf{x}_0}{1-\bar{\alpha}_{t-1}}\right). \tag{A41}$$

Substitute $\widetilde{\mathbf{\Sigma}}_t$ with Eq. (A40):

$$\begin{aligned}\tilde{\boldsymbol{\mu}}_t&=\frac{1-\bar{\alpha}_{t-1}}{1-\bar{\alpha}_t}\beta_t\mathbf{I}_D\left(\frac{\sqrt{1-\beta_t}\mathbf{x}_t}{\beta_t}+\frac{\sqrt{\bar{\alpha}_{t-1}}\mathbf{x}_0}{1-\bar{\alpha}_{t-1}}\right)\\ &=\frac{\sqrt{1-\beta_t}\mathbf{x}_t}{\beta_t}\frac{1-\bar{\alpha}_{t-1}}{1-\bar{\alpha}_t}\beta_t+\frac{\sqrt{\bar{\alpha}_{t-1}}\mathbf{x}_0}{1-\bar{\alpha}_{t-1}}\frac{1-\bar{\alpha}_{t-1}}{1-\bar{\alpha}_t}\beta_t\end{aligned}$$

$$= \frac{\sqrt{1-\beta_t}(1-\bar{\alpha}_{t-1})}{1-\bar{\alpha}_t}\mathbf{x}_t + \frac{\sqrt{\bar{\alpha}_{t-1}}\beta_t}{1-\bar{\alpha}_t}\mathbf{x}_0. \tag{A42}$$

So far, the derivations of Eqs. (7), (8), and (9) have been completed.

### A.3 Derivation of Eq. (16)

First, change the form of Eq. (6):

$$\mathbf{x}_0 = \frac{1}{\sqrt{\bar{\alpha}_t}}(\mathbf{x}_t - \sqrt{1-\bar{\alpha}_t}\boldsymbol{\epsilon}_0^t). \tag{A43}$$

Plug the above equation into Eq. (8):

$$\begin{aligned}
\tilde{\boldsymbol{\mu}}_t(\mathbf{x}_t, \mathbf{x}_0) &= \frac{\sqrt{1-\beta_t}(1-\bar{\alpha}_{t-1})}{1-\bar{\alpha}_t}\mathbf{x}_t + \frac{\sqrt{\bar{\alpha}_{t-1}}\beta_t}{1-\bar{\alpha}_t}\mathbf{x}_0 \\
&= \frac{\sqrt{1-\beta_t}(1-\bar{\alpha}_{t-1})}{1-\bar{\alpha}_t}\mathbf{x}_t + \frac{\sqrt{\bar{\alpha}_{t-1}}\beta_t}{1-\bar{\alpha}_t} \times \frac{\mathbf{x}_t - \sqrt{1-\bar{\alpha}_t}\boldsymbol{\epsilon}_0^t}{\sqrt{\bar{\alpha}_t}} \\
&= \left(\frac{\sqrt{\alpha_t}(1-\bar{\alpha}_{t-1})}{1-\bar{\alpha}_t} + \frac{\beta_t}{(1-\bar{\alpha}_t)\sqrt{\alpha_t}}\right)\mathbf{x}_t - \frac{\beta_t}{\sqrt{1-\bar{\alpha}_t}} \times \frac{\boldsymbol{\epsilon}_0^t}{\sqrt{\alpha_t}} \\
&= \left(\frac{\alpha_t(1-\bar{\alpha}_{t-1}) + 1 - \alpha_t}{(1-\bar{\alpha}_t)\sqrt{\alpha_t}}\right)\mathbf{x}_t - \frac{1}{\sqrt{\alpha_t}}\frac{\beta_t}{\sqrt{1-\bar{\alpha}_t}}\boldsymbol{\epsilon}_0^t \\
&= \frac{1-\alpha_t\bar{\alpha}_{t-1}}{(1-\bar{\alpha}_t)\sqrt{\alpha_t}}\mathbf{x}_t - \frac{1}{\sqrt{\alpha_t}}\frac{\beta_t}{\sqrt{1-\bar{\alpha}_t}}\boldsymbol{\epsilon}_0^t \\
&= \frac{1}{\sqrt{\alpha_t}}\left(\mathbf{x}_t - \frac{1-\alpha_t}{\sqrt{1-\bar{\alpha}_t}}\boldsymbol{\epsilon}_0^t\right).
\end{aligned} \tag{A44}$$

Further, $\alpha_t$ can be replaced by $1-\beta_t$:

$$\tilde{\boldsymbol{\mu}}_t = \frac{1}{\sqrt{1-\beta_t}}\left(\mathbf{x}_t - \frac{\beta_t}{\sqrt{1-\bar{\alpha}_t}}\boldsymbol{\epsilon}_0^t\right). \tag{A45}$$

Thus, we finish the derivation of Eq. (16).

## Appendix B Proof of the optimization process

In Subsection 5.3, we get $q(\mathbf{x}_{t-1}|\mathbf{x}_t, \mathbf{x}_0) = \mathcal{N}\left(\mathbf{x}_{t-1}; \tilde{\boldsymbol{\mu}}_t(\mathbf{x}_t, \mathbf{x}_0), \widetilde{\boldsymbol{\Sigma}}_t\right)$. In Subsection 5.4, we know that $p_{\boldsymbol{\theta}}(\mathbf{x}_{t-1}|\mathbf{x}_t, \mathbf{y}, d) = \mathcal{N}\left(\mathbf{x}_{t-1}; \boldsymbol{\mu}_{\boldsymbol{\theta}}(\mathbf{x}_t, t, \mathbf{y}, d), \boldsymbol{\Sigma}_{\boldsymbol{\theta}}(\mathbf{x}_t, t, \mathbf{y}, d)\right)$, where $\boldsymbol{\Sigma}_{\boldsymbol{\theta}}(\mathbf{x}_t, t, \mathbf{y}, d)$ is set to $\widetilde{\boldsymbol{\Sigma}}_t$. In this section, we first want to prove the relationship between minimizing the loss function in Eq. (12), denoted as $\mathcal{L}$, and optimizing $p_{\boldsymbol{\theta}}(\mathbf{x}_{t-1}|\mathbf{x}_t, \mathbf{y}, d)$ to make it close to $q(\mathbf{x}_{t-1}|\mathbf{x}_t, \mathbf{x}_0)$.

The first thing is to use an appropriate mathematical tool to represent the proximity between two distributions, namely, quantify the distance. In machine learning, Kullback-Leibler divergence or KL divergence is the most widely used metric to measure the distance between two distributions. Specifically, the smaller the KL divergence, the ***"closer"*** or ***"more similar"*** the two distributions are. The KL divergence can be defined by the following equation:

$$D_{\mathbb{KL}}(P \parallel Q) \triangleq \int_{\mathcal{X}} P(x)\log\frac{P(x)}{Q(x)}\mathrm{d}x, \tag{B1}$$

where $D_{\mathbb{KL}}$ is the KL divergence, $P$ and $Q$ are two arbitrary continuous distributions, and $\mathcal{X}$ denotes the sample space. The KL divergence is also known as the information gain or the relative entropy [92].

Because the distributions $p_{\boldsymbol{\theta}}$ and $q$ researched here are both multivariate Gaussian distributions, we give the next simple proposition without proof to show the KL divergence between two multivariate Gaussian distributions:

**Proposition 1.** The KL divergence between two multivariate Gaussian distributions is

$$D_{\mathbb{KL}}\big(\mathcal{N}(\mathbf{x};\boldsymbol{\mu}_1,\boldsymbol{\Sigma}_1)\parallel\mathcal{N}(\mathbf{y};\boldsymbol{\mu}_2,\boldsymbol{\Sigma}_2)\big)=\frac{1}{2}\left[\mathrm{Tr}(\boldsymbol{\Sigma}_2^{-1}\boldsymbol{\Sigma}_1)+(\boldsymbol{\mu}_2-\boldsymbol{\mu}_1)^\top\boldsymbol{\Sigma}_2^{-1}(\boldsymbol{\mu}_2-\boldsymbol{\mu}_1)-D+\log\frac{|\boldsymbol{\Sigma}_2|}{|\boldsymbol{\Sigma}_1|}\right], \quad \text{(B2)}$$

where $(\boldsymbol{\mu}_1,\boldsymbol{\Sigma}_1)$ and $(\boldsymbol{\mu}_2,\boldsymbol{\Sigma}_2)$ represent the mean vectors and covariance matrices of the two distributions, respectively. $D$ means the dimension of the two random vectors, $\mathbf{X}$ and $\mathbf{Y}$.

Then, a key theorem is given and proven.

**Theorem 2.** Minimizing the loss function $\mathcal{L}$ is equivalent to making $p_{\boldsymbol{\theta}}(\mathbf{x}_{t-1}|\mathbf{x}_t,\mathbf{y},d)$ close to $q(\mathbf{x}_{t-1}|\mathbf{x}_t,\mathbf{x}_0)$.

*Proof.* Since we have $p_{\boldsymbol{\theta}}(\mathbf{x}_{t-1}|\mathbf{x}_t,\mathbf{y},d)=\mathcal{N}\left(\mathbf{x}_{t-1};\boldsymbol{\mu}_{\boldsymbol{\theta}}(\mathbf{x}_t,t,\mathbf{y},d),\widetilde{\boldsymbol{\Sigma}}_t\right)$ and $q(\mathbf{x}_{t-1}|\mathbf{x}_t,\mathbf{x}_0)=\mathcal{N}\left(\mathbf{x}_{t-1};\tilde{\boldsymbol{\mu}}_t(\mathbf{x}_t,\mathbf{x}_0),\widetilde{\boldsymbol{\Sigma}}_t\right)$, plug them into Eq. (B2) in Proposition 1:

$$\begin{aligned}
&D_{\mathbb{KL}}\big(p_{\boldsymbol{\theta}}(\mathbf{x}_{t-1}|\mathbf{x}_t,\mathbf{y},d)\parallel q(\mathbf{x}_{t-1}|\mathbf{x}_t,\mathbf{x}_0)\big)\\
&=D_{\mathbb{KL}}\left(\mathcal{N}\left(\mathbf{x}_{t-1};\boldsymbol{\mu}_{\boldsymbol{\theta}}(\mathbf{x}_t,t,\mathbf{y},d),\widetilde{\boldsymbol{\Sigma}}_t\right)\parallel\mathcal{N}\left(\mathbf{x}_{t-1};\tilde{\boldsymbol{\mu}}_t(\mathbf{x}_t,\mathbf{x}_0),\widetilde{\boldsymbol{\Sigma}}_t\right)\right)\\
&=\frac{1}{2}\left[\mathrm{Tr}\left(\widetilde{\boldsymbol{\Sigma}}_t^{-1}\widetilde{\boldsymbol{\Sigma}}_t\right)+(\tilde{\boldsymbol{\mu}}_t-\boldsymbol{\mu}_{\boldsymbol{\theta}})^\top\widetilde{\boldsymbol{\Sigma}}_t^{-1}(\tilde{\boldsymbol{\mu}}_t-\boldsymbol{\mu}_{\boldsymbol{\theta}})-D+\log\frac{\left|\widetilde{\boldsymbol{\Sigma}}_t\right|}{\left|\widetilde{\boldsymbol{\Sigma}}_t\right|}\right].
\end{aligned} \quad \text{(B3)}$$

Further, we have known $\widetilde{\boldsymbol{\Sigma}}_t=\tilde{\beta}_t\mathbf{I}_D$, so $\widetilde{\boldsymbol{\Sigma}}_t^{-1}=\tilde{\beta}_t^{-1}\mathbf{I}_D$. It is worth noting that $p_{\boldsymbol{\theta}}$ and $q$ are named isotropic Gaussian distributions due to the property. Then, plug the two equations into Eq. (B3):

$$\begin{aligned}
&D_{\mathbb{KL}}\big(p_{\boldsymbol{\theta}}(\mathbf{x}_{t-1}|\mathbf{x}_t,\mathbf{y},d)\parallel q(\mathbf{x}_{t-1}|\mathbf{x}_t,\mathbf{x}_0)\big)\\
&=\frac{1}{2}\left[\mathrm{Tr}\left(\tilde{\beta}_t^{-1}\mathbf{I}_D\tilde{\beta}_t\mathbf{I}_D\right)+(\tilde{\boldsymbol{\mu}}_t-\boldsymbol{\mu}_{\boldsymbol{\theta}})^\top\tilde{\beta}_t^{-1}\mathbf{I}_D(\tilde{\boldsymbol{\mu}}_t-\boldsymbol{\mu}_{\boldsymbol{\theta}})-D+\log 1\right]\\
&=\frac{1}{2}\left[\mathrm{Tr}(\mathbf{I}_D)+\tilde{\beta}_t^{-1}(\tilde{\boldsymbol{\mu}}_t-\boldsymbol{\mu}_{\boldsymbol{\theta}})^\top(\tilde{\boldsymbol{\mu}}_t-\boldsymbol{\mu}_{\boldsymbol{\theta}})-D\right]\\
&=\frac{1}{2}\left[D+\tilde{\beta}_t^{-1}\|\tilde{\boldsymbol{\mu}}_t-\boldsymbol{\mu}_{\boldsymbol{\theta}}\|^2-D\right]\\
&=\frac{1}{2\tilde{\beta}_t}[\|\tilde{\boldsymbol{\mu}}_t-\boldsymbol{\mu}_{\boldsymbol{\theta}}\|^2].
\end{aligned} \quad \text{(B4)}$$

Recall Eq. (12):

$$\mathcal{L}=\mathbb{E}_{t\sim\mathrm{Uniform}[1,T],\mathbf{x}_0\sim q_0,\boldsymbol{\epsilon}\sim\mathcal{N}(0,\mathbf{I}_D)}\|\tilde{\boldsymbol{\mu}}_t-\boldsymbol{\mu}_{\boldsymbol{\theta}}(\mathbf{x}_t,t,\mathbf{y},d)\|_2^2, \quad \text{(B5)}$$

and $\tilde{\beta}_t=\frac{1-\bar{\alpha}_{t-1}}{1-\bar{\alpha}_t}\beta_t>0$. Therefore,

$$\min_{\boldsymbol{\theta}}\mathcal{L}\Longleftrightarrow\min_{\boldsymbol{\theta}}\|\tilde{\boldsymbol{\mu}}_t-\boldsymbol{\mu}_{\boldsymbol{\theta}}(\mathbf{x}_t,t,\mathbf{y},d)\|_2^2\Longleftrightarrow\min_{\boldsymbol{\theta}}\frac{1}{2\tilde{\beta}_t}[\|\tilde{\boldsymbol{\mu}}_t-\boldsymbol{\mu}_{\boldsymbol{\theta}}\|^2]\Longleftrightarrow\min_{\boldsymbol{\theta}}D_{\mathbb{KL}}\big(p_{\boldsymbol{\theta}}(\mathbf{x}_{t-1}|\mathbf{x}_t,\mathbf{y},d)\parallel q(\mathbf{x}_{t-1}|\mathbf{x}_t,\mathbf{x}_0)\big). \quad \text{(B6)}$$

In other words, minimizing the loss function $\mathcal{L}$ is equivalent to making $p_{\boldsymbol{\theta}}(\mathbf{x}_{t-1}|\mathbf{x}_t,\mathbf{y},d)$ close to $q(\mathbf{x}_{t-1}|\mathbf{x}_t,\mathbf{x}_0)$, which completes the proof of the Theorem 2. □

Further, our final goal is to maximize the likelihood function $p_{\boldsymbol{\theta}}(\mathbf{x}_0|\mathbf{y},d)$ or the log-likelihood $\log p_{\boldsymbol{\theta}}(\mathbf{x}_0|\mathbf{y},d)$ based on knowledge of variational autoencoders (VAEs). Then, we present another theorem and the corresponding proof below to show the equivalent relation between reducing the distance of $p_{\boldsymbol{\theta}}(\mathbf{x}_{t-1}|\mathbf{x}_t,\mathbf{y},d)$ and $q(\mathbf{x}_{t-1}|\mathbf{x}_t,\mathbf{x}_0)$, and maximizing $\log p_{\boldsymbol{\theta}}(\mathbf{x}_0|\mathbf{y},d)$:

**Theorem 3.** Making $p_{\boldsymbol{\theta}}(\mathbf{x}_{t-1}|\mathbf{x}_t,\mathbf{y},d)$ close to $q(\mathbf{x}_{t-1}|\mathbf{x}_t,\mathbf{x}_0)$ is equivalent to maximizing the log-likelihood

function $\log p_{\boldsymbol{\theta}}(\mathbf{x}_0|\mathbf{y},d)$.

*Proof.* The proof of this theorem mainly refers to the variational inference in VAEs and [66].

$$\begin{aligned}\max_{\boldsymbol{\theta}} \log p_{\boldsymbol{\theta}}(\mathbf{x}_0|\mathbf{y},d) &= \max_{\boldsymbol{\theta}} \log \int_{\mathcal{X}} p_{\boldsymbol{\theta}}(\mathbf{x}_{0:T}|\mathbf{y},d)\mathrm{d}\mathbf{x}_{1:T} \\ &= \max_{\boldsymbol{\theta}} \log \int_{\mathcal{X}} \frac{p_{\boldsymbol{\theta}}(\mathbf{x}_{0:T}|\mathbf{y},d)q(\mathbf{x}_{1:T}|\mathbf{x}_0)}{q(\mathbf{x}_{1:T}|\mathbf{x}_0)}\mathrm{d}\mathbf{x}_{1:T} \\ &= \max_{\boldsymbol{\theta}} \log \mathbb{E}_{q(\mathbf{x}_{1:T}|\mathbf{x}_0)}\left[\frac{p_{\boldsymbol{\theta}}(\mathbf{x}_{0:T}|\mathbf{y},d)}{q(\mathbf{x}_{1:T}|\mathbf{x}_0)}\right], \end{aligned} \tag{B7}$$

where, $p_{\boldsymbol{\theta}}(\mathbf{x}_{0:T}|\mathbf{y},d)$ means the joint distribution of $p_{\boldsymbol{\theta}}(\mathbf{x}_0|\mathbf{y},d), p_{\boldsymbol{\theta}}(\mathbf{x}_1|\mathbf{y},d), \dots, p_{\boldsymbol{\theta}}(\mathbf{x}_T|\mathbf{y},d)$ . Similarly, $q(\mathbf{x}_{1:T}|\mathbf{x}_0)$ is also a joint distribution. Then, based on Eq. (B7) and Jensen's inequality, we have

$$\begin{aligned}
&\max_{\boldsymbol{\theta}} \log p_{\boldsymbol{\theta}}(\mathbf{x}_0|\mathbf{y},d) \\
&\geq \max_{\boldsymbol{\theta}} \mathbb{E}_{q(\mathbf{x}_{1:T}|\mathbf{x}_0)}\left[\log \frac{p_{\boldsymbol{\theta}}(\mathbf{x}_{0:T}|\mathbf{y},d)}{q(\mathbf{x}_{1:T}|\mathbf{x}_0)}\right] \\
&= \max_{\boldsymbol{\theta}} \mathbb{E}_{q(\mathbf{x}_{1:T}|\mathbf{x}_0)}\left[\log \frac{p(\mathbf{x}_T|\mathbf{y},d)\prod_{t=1}^{T} p_{\boldsymbol{\theta}}(\mathbf{x}_{t-1}|\mathbf{x}_t,\mathbf{y},d)}{\prod_{t=1}^{T} q(\mathbf{x}_t|\mathbf{x}_{t-1})}\right] \\
&= \max_{\boldsymbol{\theta}} \mathbb{E}_{q(\mathbf{x}_{1:T}|\mathbf{x}_0)}\left[\log \frac{p(\mathbf{x}_T|\mathbf{y},d)p_{\boldsymbol{\theta}}(\mathbf{x}_0|\mathbf{x}_1,\mathbf{y},d)}{q(\mathbf{x}_1|\mathbf{x}_0)} + \log \frac{\prod_{t=2}^{T} p_{\boldsymbol{\theta}}(\mathbf{x}_{t-1}|\mathbf{x}_t,\mathbf{y},d)}{\prod_{t=2}^{T} q(\mathbf{x}_t|\mathbf{x}_{t-1},\mathbf{x}_0)}\right] \\
&= \max_{\boldsymbol{\theta}} \mathbb{E}_{q(\mathbf{x}_{1:T}|\mathbf{x}_0)}\left[\log \frac{p(\mathbf{x}_T|\mathbf{y},d)p_{\boldsymbol{\theta}}(\mathbf{x}_0|\mathbf{x}_1,\mathbf{y},d)}{q(\mathbf{x}_T|\mathbf{x}_0)} + \sum_{t=2}^{T} \log \frac{p_{\boldsymbol{\theta}}(\mathbf{x}_{t-1}|\mathbf{x}_t,\mathbf{y},d)}{q(\mathbf{x}_{t-1}|\mathbf{x}_t,\mathbf{x}_0)}\right] \\
&= \max_{\boldsymbol{\theta}} \mathbb{E}_{q(\mathbf{x}_{1:T}|\mathbf{x}_0)}\left[\log p_{\boldsymbol{\theta}}(\mathbf{x}_0|\mathbf{x}_1,\mathbf{y},d) + \log \frac{p(\mathbf{x}_T|\mathbf{y},d)}{q(\mathbf{x}_T|\mathbf{x}_0)} + \sum_{t=2}^{T} \log \frac{p_{\boldsymbol{\theta}}(\mathbf{x}_{t-1}|\mathbf{x}_t,\mathbf{y},d)}{q(\mathbf{x}_{t-1}|\mathbf{x}_t,\mathbf{x}_0)}\right] \\
&= \max_{\boldsymbol{\theta}} \mathbb{E}_{q(\mathbf{x}_1|\mathbf{x}_0)}[\log p_{\boldsymbol{\theta}}(\mathbf{x}_0|\mathbf{x}_1,\mathbf{y},d)] - \mathbb{E}_{q(\mathbf{x}_T|\mathbf{x}_0)}\left[\log \frac{q(\mathbf{x}_T|\mathbf{x}_0)}{p(\mathbf{x}_T|\mathbf{y},d)}\right] - \sum_{t=2}^{T} \mathbb{E}_{q(\mathbf{x}_{t-1},\mathbf{x}_t|\mathbf{x}_0)}\left[\log \frac{q(\mathbf{x}_{t-1}|\mathbf{x}_t,\mathbf{x}_0)}{p_{\boldsymbol{\theta}}(\mathbf{x}_{t-1}|\mathbf{x}_t,\mathbf{y},d)}\right] \\
&= \max_{\boldsymbol{\theta}} \left\{\mathbb{E}_{q(\mathbf{x}_1|\mathbf{x}_0)}[\log p_{\boldsymbol{\theta}}(\mathbf{x}_0|\mathbf{x}_1,\mathbf{y},d)] - D_{\mathbb{KL}}\big(q(\mathbf{x}_T|\mathbf{x}_0) \parallel p(\mathbf{x}_T|\mathbf{y},d)\big) - \sum_{t=2}^{T} \mathbb{E}_{q(\mathbf{x}_t|\mathbf{x}_0)} D_{\mathbb{KL}}\big(q(\mathbf{x}_{t-1}|\mathbf{x}_t,\mathbf{x}_0) \parallel p_{\boldsymbol{\theta}}(\mathbf{x}_{t-1}|\mathbf{x}_t,\mathbf{y},d)\big)\right\},
\end{aligned} \tag{B8}$$

where $\mathbb{E}_{q(\mathbf{x}_1|\mathbf{x}_0)}[\log p_{\boldsymbol{\theta}}(\mathbf{x}_0|\mathbf{x}_1,\mathbf{y},d)]$ is interpreted as a reconstruction term following [66], and DDPM finds that omitting this term is beneficial and does not affect the generations. $D_{\mathbb{KL}}\big(q(\mathbf{x}_T|\mathbf{x}_0) \parallel p(\mathbf{x}_T|\mathbf{y},d)\big)$ is a prior matching term. Because $q(\mathbf{x}_T|\mathbf{x}_0) = \mathcal{N}(\mathbf{x}_T; \sqrt{\bar{\alpha}_T}\mathbf{x}_0, (1-\bar{\alpha}_T)\mathbf{I}_D)$ (recall Eq. (5) ) and when $T$ closes to infinity, $q(\mathbf{x}_T|\mathbf{x}_0) \approx \mathcal{N}(\mathbf{x}_T; 0, \mathbf{I}_D)$, and $p(\mathbf{x}_T|\mathbf{y},d) = p(\mathbf{x}_T) = \mathcal{N}(0, \mathbf{I}_D)$ (recall Subsection 5.3), there are no learnable parameters in this term, and it is approximately equal to 0. Therefore, this matching term can also be omitted. So,

$\max_{\boldsymbol{\theta}} \log p_{\boldsymbol{\theta}}(\mathbf{x}_0|\mathbf{y},d) \Longleftrightarrow \max_{\boldsymbol{\theta}} -\sum_{t=2}^{T} \mathbb{E}_{q(\mathbf{x}_t|\mathbf{x}_0)} D_{\mathbb{KL}}\big(q(\mathbf{x}_{t-1}|\mathbf{x}_t,\mathbf{x}_0) \parallel p_{\boldsymbol{\theta}}(\mathbf{x}_{t-1}|\mathbf{x}_t,\mathbf{y},d)\big)$

$\Longleftrightarrow \min_{\boldsymbol{\theta}} \sum_{t=2}^{T} \mathbb{E}_{q(\mathbf{x}_t|\mathbf{x}_0)} D_{\mathbb{KL}}\big(q(\mathbf{x}_{t-1}|\mathbf{x}_t,\mathbf{x}_0) \parallel p_{\boldsymbol{\theta}}(\mathbf{x}_{t-1}|\mathbf{x}_t,\mathbf{y},d)\big)$. It is important to note that Eq. (B8) is usually known as the variational lower bound (VLB) or the evidence lower bound (ELBO) of $\log p_{\boldsymbol{\theta}}(\mathbf{x}_0|\mathbf{y},d)$. By the above derivation, it is clear that maximizing $\log p_{\boldsymbol{\theta}}(\mathbf{x}_0|\mathbf{y},d)$ is approximately equivalent to making $p_{\boldsymbol{\theta}}(\mathbf{x}_{t-1}|\mathbf{x}_t,\mathbf{y},d)$ close to $q(\mathbf{x}_{t-1}|\mathbf{x}_t,\mathbf{x}_0)$. The proof is completed. □

**Remark 2.** There are two points we should be aware of: (1) Combining Theorem 2 and Theorem 3, we see that minimizing the loss function $\mathcal{L}_1$ is equivalent to maximizing the log-likelihood $\log p_{\boldsymbol{\theta}}(\mathbf{x}_0|\mathbf{y},d)$. Therefore, we prove the effectiveness of the optimization process. (2) In the parameterization of $p_{\boldsymbol{\theta}}(\mathbf{x}_{t-1}|\mathbf{x}_t,\mathbf{y},d)$, we assume that the covariance matrix $\boldsymbol{\Sigma}_{\boldsymbol{\theta}}(\mathbf{x}_t,t,\mathbf{y},d)$ is equal to $\widetilde{\boldsymbol{\Sigma}}_t$, that is the covariance matrix of $q(\mathbf{x}_{t-1}|\mathbf{x}_t,\mathbf{x}_0)$ for simplicity. In fact, $\boldsymbol{\Sigma}_{\boldsymbol{\theta}}(\mathbf{x}_t,t,\mathbf{y},d)$ can also be trainable just like $\boldsymbol{\mu}_{\boldsymbol{\theta}}(\mathbf{x}_t,t,\mathbf{y},d)$. We leave that to future work.

## Appendix C Calculation process of the attention block (AB)

This appendix details the calculation process of AB described in Subsection 5.5.3, which consists of a self-attention block (SAB) and a parallel cross-attention block (PCAB).

**SAB.** It is known that the input tensor is $\mathbf{M}_s$. SAB firstly generates Query ($\mathbf{Q}$), Key ($\mathbf{K}$), and Value ($\mathbf{V}$) matrices by a IN layer, three 3 × 3 convolutions, and reshape operations:

$$\mathbf{Q} = \mathrm{R}[\mathbf{W}_Q \circledast \mathrm{IN}(\mathbf{M}_s)], \tag{C1a}$$

$$\mathbf{K} = \mathrm{R}[\mathbf{W}_K \circledast \mathrm{IN}(\mathbf{M}_s)], \tag{C1b}$$

$$\mathbf{V} = \mathrm{R}[\mathbf{W}_V \circledast \mathrm{IN}(\mathbf{M}_s)], \tag{C1c}$$

where $\mathbf{Q}, \mathbf{K}, \mathbf{V} \in \mathbb{R}^{C\times HW}$. $\mathbf{W}_Q, \mathbf{W}_K$, and $\mathbf{W}_V$ represent the Query, Key, and Value weight matrices learned by the three 3 × 3 convolutions, respectively. $\circledast$ denotes convolution operation. $\mathrm{R}$ means the reshape operation. Then self-attention can be computed by the following formula:

$$\mathrm{Attention}(\mathbf{Q}, \mathbf{K}, \mathbf{V}) \triangleq \mathbf{V}\left[\mathrm{SoftMax}\left(\frac{\mathbf{Q}^\top \mathbf{K}}{\sqrt{S}}\right)\right]^\top \in \mathbb{R}^{C\times HW}, \tag{C2}$$

where, $S = C$ represents the scaling parameter to control the magnitude of the product. Finally, the output can be obtained after a skip connection:

$$\mathbf{O}_s = \mathrm{PW\ Conv}[\mathrm{R}(\mathrm{Attention}(\mathbf{Q}, \mathbf{K}, \mathbf{V}))] + \mathbf{M}_s, \tag{C3}$$

where $\mathrm{PW\ Conv}$ is a point-wise convolution. The skip connection is adopted to fuse the original feature.

**PCAB.** Two cross-attention blocks (CABs) form the core part of PCAB. Because the remaining parts of PCAB have been introduced in Subsection 5.5.3, here we focus on CAB. Since the 2 CABs have the same structures, for simplicity, we just take $\mathrm{CAB}(\mathbf{O}_s, \mathbf{E}_t)$ as an example to explain the calculation process of CAB.

Similarly with SAB, CAB also firstly generates Query, Key, and Value matrices:

$$\mathbf{Q} = \mathrm{R}[\mathbf{W}_Q \circledast \mathrm{IN}(\mathbf{M}_s)], \tag{C4a}$$

$$\mathbf{K}_t = \mathrm{R}[\mathbf{W}_K \times \mathbf{E}_t], \tag{C4b}$$

$$\mathbf{V}_t = \mathrm{R}[\mathbf{W}_V \times \mathbf{E}_t], \tag{C4c}$$

where $\mathbf{Q} \in \mathbb{R}^{C\times HW}$ and $\mathbf{W}_Q$ represents the Query projection matrix learned by a 3 × 3 convolution, same as the SAB. $\mathbf{K}_t, \mathbf{V}_t \in \mathbb{R}^{C\times 1}$. $\mathbf{E}_t \in \mathbb{R}^{D_t\times 1\times 1}$ is an expanded matrix from the vanilla embedding to align with $\mathbf{M}_s$ on dimensions (the same as the operation in Res-Block). $\mathbf{W}_K$ and $\mathbf{W}_V$ represent the Key, and Value projection matrices learned by two point-wise convolutions. It is worth noting that this allocation strategy that $\mathbf{Q}$ comes from $\mathbf{M}_s$, and $\mathbf{K}_t$ and $\mathbf{V}_t$ come from $\mathbf{E}_t$ is a common trick when we need to compute cross-attention on data with two kinds of domains, just like Stable Diffusion and Grounding DINO [93]. Then, plugging $\mathbf{Q}$, $\mathbf{K}_t$, and $\mathbf{V}_t$ into Eq. (C2), we obtain the $\mathrm{Attention}(\mathbf{Q}, \mathbf{K}_t, \mathbf{V}_t) \in \mathbb{R}^{C\times HW}$. Next, the output of $\mathrm{CAB}(\mathbf{O}_s, \mathbf{E}_t)$, $\mathbf{O}_C^t$, can be finally obtained by executing the following formula:

$$\mathbf{O}_C^t = \mathrm{PW\ Conv}[\mathrm{R}(\mathrm{Attention}(\mathbf{Q}, \mathbf{K}_t, \mathbf{V}_t))]. \tag{C5}$$